\documentclass[10pt,twocolumn,letterpaper]{article}

\usepackage[pagenumbers]{cvpr} 

\usepackage[dvipsnames]{xcolor}

\definecolor{cvprblue}{rgb}{0.21,0.49,0.74}
\usepackage[pagebackref,breaklinks,colorlinks,citecolor=cvprblue]{hyperref}
\usepackage{bm}
\usepackage{multirow}
\usepackage{graphicx}
\usepackage{bbding}
\usepackage{colortbl}
\usepackage{amsmath,amssymb}
\usepackage{threeparttable}
\usepackage{booktabs}
\usepackage[capitalize]{cleveref}
\usepackage[sectionbib]{chapterbib}
\usepackage{chapterbib}

\definecolor{bestresult}{RGB}{255,0,0}    
\definecolor{secbestresult}{RGB}{0,0,255} 

\def\eg{\emph{e.g}\onedot} 
\def\ie{\emph{i.e}\onedot} 
 
\def\etc{\emph{etc}\onedot} 
 
\def\etal{\emph{et al}\onedot}

\makeatother

\def\paperID{1234} 
\def\confName{CVPR}
\def\confYear{2024}

\title{Technique Report of CVPR 2024 PBDL Challenges}

\author{Ying Fu$^*$ \and Yu Li$^*$ \and Shaodi You$^*$ \and Boxin Shi$^*$ \and
Linwei Chen \and Yunhao Zou \and Zichun Wang \and Yichen Li \and Yuze Han \and Yingkai Zhang \and Jianan Wang
 \and Qinglin Liu \and Wei Yu \and Xiaoqian Lv \and Jianing Li \and Shengping Zhang \and Xiangyang Ji \and Yuanpei Chen \and Yuhan Zhang \and Weihang Peng \and Liwen Zhang \and Zhe Xu \and Dingyong Gou \and Cong Li \and Senyan Xu \and Yunkang Zhang \and Siyuan Jiang \and Xiaoqiang Lu \and Licheng Jiao \and Fang Liu \and Xu Liu \and Lingling Li \and Wenping Ma \and Shuyuan Yang \and Haiyang Xie \and Jian Zhao \and Shihua Huang \and Peng Cheng \and Xi Shen \and Zheng Wang \and Shuai An \and Caizhi Zhu \and Xuelong Li \and Tao Zhang \and Liang Li \and Yu Liu \and Chenggang Yan \and Gengchen Zhang \and Linyan Jiang \and Bingyi Song \and Zhuoyu An \and Haibo Lei \and Qing Luo \and Jie Song \and Yuan Liu \and Qihang Li \and Haoyuan Zhang \and Lingfeng Wang \and Wei Chen \and Aling Luo \and Cheng Li \and Jun Cao \and Shu Chen \and Zifei Dou \and Xinyu Liu \and Jing Zhang \and Kexin Zhang \and Yuting Yang \and Xuejian Gou \and Qinliang Wang \and Yang Liu \and Shizhan Zhao \and Yanzhao Zhang \and Libo Yan \and Yuwei Guo \and Guoxin Li \and Qiong Gao \and Chenyue Che \and Long Sun \and Xiang Chen \and Hao Li \and Jinshan Pan \and Chuanlong Xie \and Hongming Chen \and Mingrui Li \and Tianchen Deng \and Jingwei Huang \and Yufeng Li \and Fei Wan \and Bingxin Xu \and Jian Cheng \and Hongzhe Liu \and Cheng Xu \and Yuxiang Zou \and Weiguo Pan \and Songyin Dai \and Sen Jia \and Junpei Zhang \and Puhua Chen
}

\begin{document}
\maketitle
\let\thefootnote\relax\footnotetext{$^*$ Ying Fu, Yu Li, Shaodi You and Boxin Shi are the challenge organizers. Ying Fu is with Beijing Institute of Technology, Yu Li is with International Digital Economy Academy, Shaodi You is with University of Amsterdam, Boxin Shi is with Peking University.}

\begin{abstract}


The intersection of physics-based vision and deep learning presents an exciting frontier for advancing computer vision technologies. By leveraging the principles of physics to inform and enhance deep learning models, we can develop more robust and accurate vision systems. Physics-based vision aims to invert the processes to recover scene properties such as shape, reflectance, light distribution, and medium properties from images. In recent years, deep learning has shown promising improvements for various vision tasks, and when combined with physics-based vision, these approaches can enhance the robustness and accuracy of vision systems. This technical report summarizes the outcomes of the Physics-Based Vision Meets Deep Learning (PBDL) 2024 challenge, held in CVPR 2024 workshop. The challenge consisted of eight tracks, focusing on Low-Light Enhancement and Detection as well as High Dynamic Range (HDR) Imaging. This report details the objectives, methodologies, and results of each track, highlighting the top-performing solutions and their innovative approaches.


\end{abstract}
\section{Introduction}
\label{sec:intro}

The integration of physics-based vision with deep learning offers a powerful paradigm for addressing complex computer vision problems. Physics-based vision seeks to model and invert physical processes to recover scene properties such as shape~\cite{wu2010fusing,sarlin2021back}, reflectance~\cite{fu2020illumination,nguyen2014training}, and light distribution~\cite{broxton2020immersive} from images. Deep learning, on the other hand, excels at learning representations and patterns from large datasets. Combining these approaches allows for the development of models that are not only data-driven but also grounded in physical principles, leading to enhanced performance in various vision tasks such as object recognition~\cite{Hong2021Crafting}, scene understanding~\cite{chen2023instance}, and image restoration~\cite{pharr2023physically,wei2020physics}.

To explore the potential of this integrated approach, we organized a comprehensive challenge at CVPR 2024, held in conjunction with the Physics-Based Vision Meets Deep Learning (PBDL) workshop. The challenge comprised eight tracks, divided into two main categories: Low-Light Enhancement and Detection, and High Dynamic Range (HDR) Imaging. Each track was designed to address specific challenges in the field and to stimulate innovation in both theoretical and practical aspects. For instance, low-light enhancement aims to improve image visibility in poorly lit environments, which is crucial for applications like autonomous driving and surveillance~\cite{wang2022sfnet}. HDR imaging, on the other hand, focuses on capturing a wider range of luminance levels to produce more realistic and detailed images, which is essential for photography and cinematography~\cite{debevec2008recovering}. This report details the objectives, methodologies, and results of each track, highlighting the top-performing solutions and their innovative approaches. In the following, we present an overview of the individual tracks. 

\subsection{Low-Light Enhancement and Detection Challenge}

\begin{enumerate}
	\item \textbf{Low-light Object Detection and Instance Segmentation}: This track aimed to improve the robustness of object detection and instance segmentation algorithms in low-light conditions. Participants developed methods to handle noise, color distortion, and detail loss, common issues in low-light environments.
	
\item \textbf{Low-light Raw Video Denoising with Realistic Motion}: Focusing on enhancing video quality in low-light conditions, this track involved denoising raw video sequences with realistic motion. The goal was to reduce noise while preserving motion integrity.

\item \textbf{Low-light SRGB Image Enhancement}: This track targeted the enhancement of SRGB images captured in low-light conditions. Participants worked on methods to recover normal-light images from very dim environments, addressing noise, color bias, and over-exposure issues.

\item \textbf{Extreme Low-Light Image Denoising}: Participants in this track aimed to develop algorithms capable of denoising images captured under extremely low-light conditions, pushing the boundaries of what is achievable in terms of noise reduction and detail preservation.

\item \textbf{Low-light Raw Image Enhancement}: This track focused on enhancing raw images captured in low-light scenarios. By leveraging the higher bit-depth of raw data, participants aimed to improve the overall image quality significantly.

\end{enumerate}

\subsection{High Dynamic Range Imaging Challenge}

\begin{enumerate}
\item \textbf{HDR Reconstruction from a Single Raw Image}: This track aimed at reconstructing high dynamic range images from single raw images. The challenge was to avoid potential misalignments common in multi-image fusion techniques while capturing a broad spectrum of intensity levels.

\item \textbf{Highspeed HDR Video Reconstruction from Events}: Participants developed methods to reconstruct HDR videos from event-based camera data. The goal was to combine the high temporal resolution of event cameras with HDR imaging techniques.

\item \textbf{Raw Image Based Over-Exposure Correction}: This track focused on correcting over-exposed regions in raw images. Participants aimed to develop techniques to recover details in both over- and under-exposed areas, resulting in visually pleasing and information-rich images.

\end{enumerate}

\subsection{Summary of Challenge Outcomes}
The challenge attracted numerous teams from around the world, each bringing innovative approaches to tackle these complex problems. This report provides a comprehensive review of the methodologies and results for each track, highlighting the top-performing solutions. The participating teams demonstrated significant advancements in low-light enhancement and HDR imaging, showcasing the potential of combining physics-based vision with deep learning. The top three methods for each track are detailed, offering insights into the state-of-the-art techniques and their practical applications.

Through this challenge, we have not only advanced the field of computer vision but also demonstrated the mutual benefits of integrating physics-based models with deep learning. The results of this challenge pave the way for future research and development in this exciting interdisciplinary area. The following sections will delve into each track individually, presenting the objectives, methodologies, and outcomes in detail.

\section{Low-light Object Detection and Instance Segmentation}
Performing object detection and instance segmentation~\cite{fang2021instances} under low-light conditions poses several challenges. \eg, images captured in low-light environments often suffer from poor quality, leading to loss of detail, color distortion, and prominent noise. These factors significantly hinder the performance of downstream vision tasks, particularly object detection and instance segmentation.

To address this challenge, the CVPR 2024 PBDL Challenge on Low-light Object Detection and Instance Segmentation aims to assess and enhance the robustness of object detection and instance segmentation algorithms on images captured in low-light environmental conditions.

In the low-light object detection track (Table \ref{tab:track1_lowlightdet_results}), the top three teams demonstrated exceptional performance. Both GroundTruth and Xocean secured the 1st rank, achieving an average precision (AP) score of 0.76. They displayed remarkable accuracy in detecting objects under low-light conditions, with AP scores of 0.89 and 0.81 at IoU thresholds of 0.50 and 0.75, respectively. UnoWhoiam secured the 3rd rank with an AP score of 0.75, showcasing their strong performance in this challenging task.

For low-light instance segmentation (Table \ref{tab:track1_lowlightseg_results}), the competition was equally intense. GroundTruth achieved the 1st rank with a mask AP score of 0.62, demonstrating their excellent ability to segment instances accurately in low-light images. 
UnoWhoiam secured the 2nd rank with an mask AP score of 0.59, while Xocean secured the 3rd rank with an mask AP score of 0.58. Both teams exhibited impressive performance in low-light instance segmentation, further emphasizing the significance of their contributions.

These results highlight the remarkable advancements made by the participating teams in addressing the challenges of low-light object detection and instance segmentation. The top-ranking teams have showcased their expertise and innovation in developing robust algorithms that excel in low-light conditions, paving the way for future advancements in computer vision research.


\begin{table}[t]
	\centering
	\setlength{\tabcolsep}{8pt}
	\caption{Leaderboard of the low-light object detection.}
	\begin{threeparttable}
		\begin{tabular}{ccccc}
			\toprule
			\textbf{Rank} & \textbf{Team} & \textbf{AP$^{box}$} & \textbf{AP$^{box}_{50}$} & \textbf{AP$^{box}_{75}$} \\ \hline
			1 & GroundTruth & 0.76 & 0.89 & 0.81 \\ 
			1 & Xocean & 0.76 & 0.89 & 0.81 \\ 
			3 & UnoWhoiam & 0.75 & 0.94 & 0.86 \\ \bottomrule
		\end{tabular}
	\end{threeparttable}
	\label{tab:track1_lowlightdet_results}
\end{table}



\begin{table}[t]
	\centering
	\caption{Leaderboard of the low-light instance segmentation.}
	\begin{threeparttable}
		\begin{tabular}{ccccc}
			\toprule
			\textbf{Rank} & \textbf{Team} & \textbf{AP$^{mask}$} & \textbf{AP$^{mask}_{50}$} & \textbf{AP$^{mask}_{75}$} \\ \hline
			1 & GroundTruth & 0.62 & 0.82 & 0.65 \\ 
			2 & UnoWhoiam & 0.59 & 0.87 & 0.61 \\ 
			3 & Xocean & 0.58 & 0.79 & 0.61 \\ \bottomrule
		\end{tabular}
	\end{threeparttable}
	\label{tab:track1_lowlightseg_results}
\end{table}

\begin{figure*}[t!]
\centering
\includegraphics[width=0.95\linewidth]{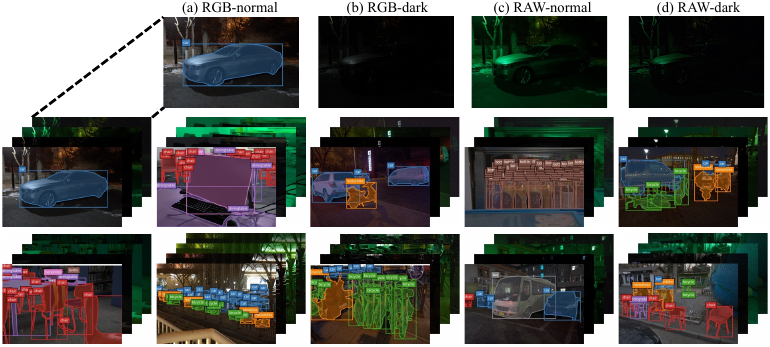}
\caption{\small Example scenes in LIS dataset.  
Four image types (long-exposure normal-light and short-exposure low-light images in both RAW and sRGB formats) are captured for each scene.}
\label{fig:track1_dataset}
\end{figure*}


\subsection{Low-light Instance Segmentation Dataset}

To systematically investigate the effectiveness of the proposed method in real-world conditions, a real low-light image dataset for instance segmentation is necessary and foundamental.
The challenge utilizes the Low-light Instance Segmentation (LIS) dataset, introduced by \cite{2021Crafting, chen2023instance}. 

It is collected using a Canon EOS 5D Mark IV camera. Figure~\ref{fig:track1_dataset} shows examples of annotated images from LIS dataset. The LIS dataset exhibits the following characteristics:

\begin{itemize}
\item \textbf{Paired samples.} The LIS dataset includes images in both sRGB-JPEG (typical camera output) and RAW formats. Each format consists of paired short-exposure low-light and corresponding long-exposure normal-light images. We term these four types of images \textit{sRGB-dark, sRGB-normal, RAW-dark, and RAW-normal}. To ensure pixel-wise alignment, we mounted the camera on a sturdy tripod and used remote control via a mobile app to avoid vibrations.
\item \textbf{Diverse scenes.} The LIS dataset consists of 2230 image pairs collected in various indoor and outdoor scenes. To increase the diversity of low-light conditions, we used a series of ISO levels (\eg, 800, 1600, 3200, 6400) to capture long-exposure reference images and deliberately decreased the exposure time by various low-light factors (\eg, 10, 20, 30, 40, 50, 100) to capture short-exposure images, simulating very low-light conditions.
\item \textbf{Instance-level pixel-wise labels.} For each image pair, we provide precise instance-level pixel-wise labels annotated by professional annotators. This results in 10,504 labeled instances across eight common object classes: bicycle, car, motorcycle, bus, bottle, chair, dining table, and TV.
\end{itemize}

The LIS dataset includes images captured in different scenes (indoor and outdoor) and under varying illumination conditions. As shown in Figure~\ref{fig:track1_dataset}, object occlusion and densely distributed objects add to the challenges presented by the low-light conditions.


\subsection{GroundTruth Team's Method}
\subsubsection{Network Architecture}

\begin{figure}[t]
\centering
\includegraphics[scale=0.24]{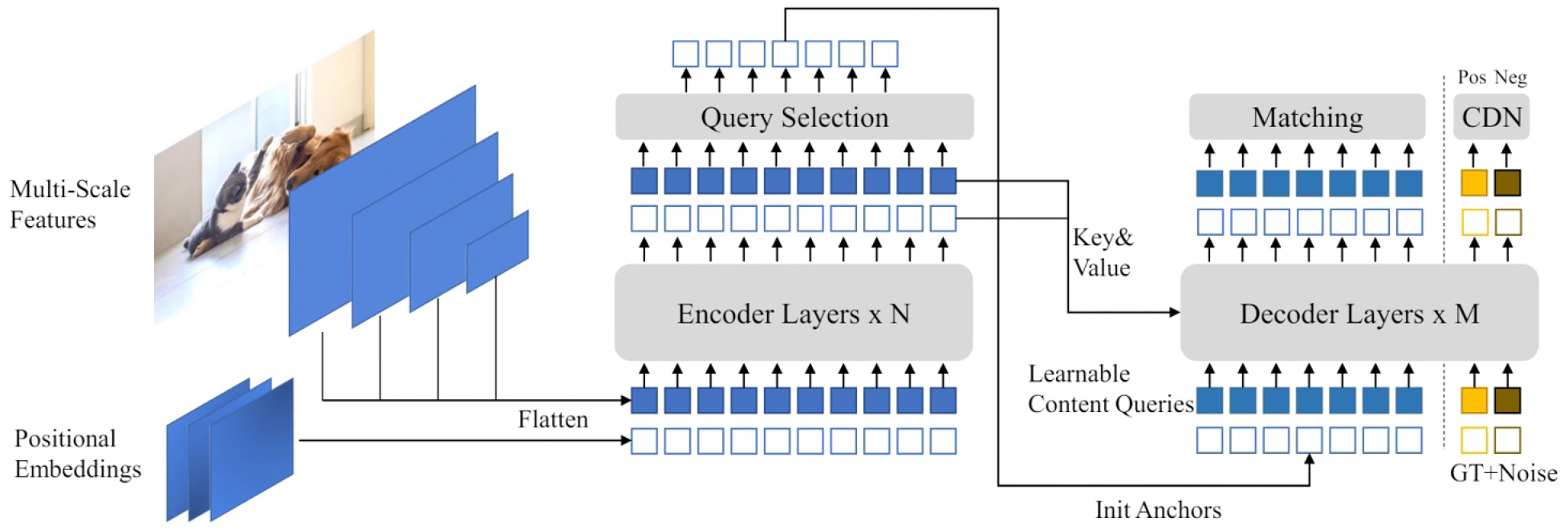}
\caption{Framework of DINO~\cite{zhang2022dino}.}
\label{fig:track1_dino}
\end{figure}

\begin{figure}[t]
\centering
\includegraphics[scale=0.32]{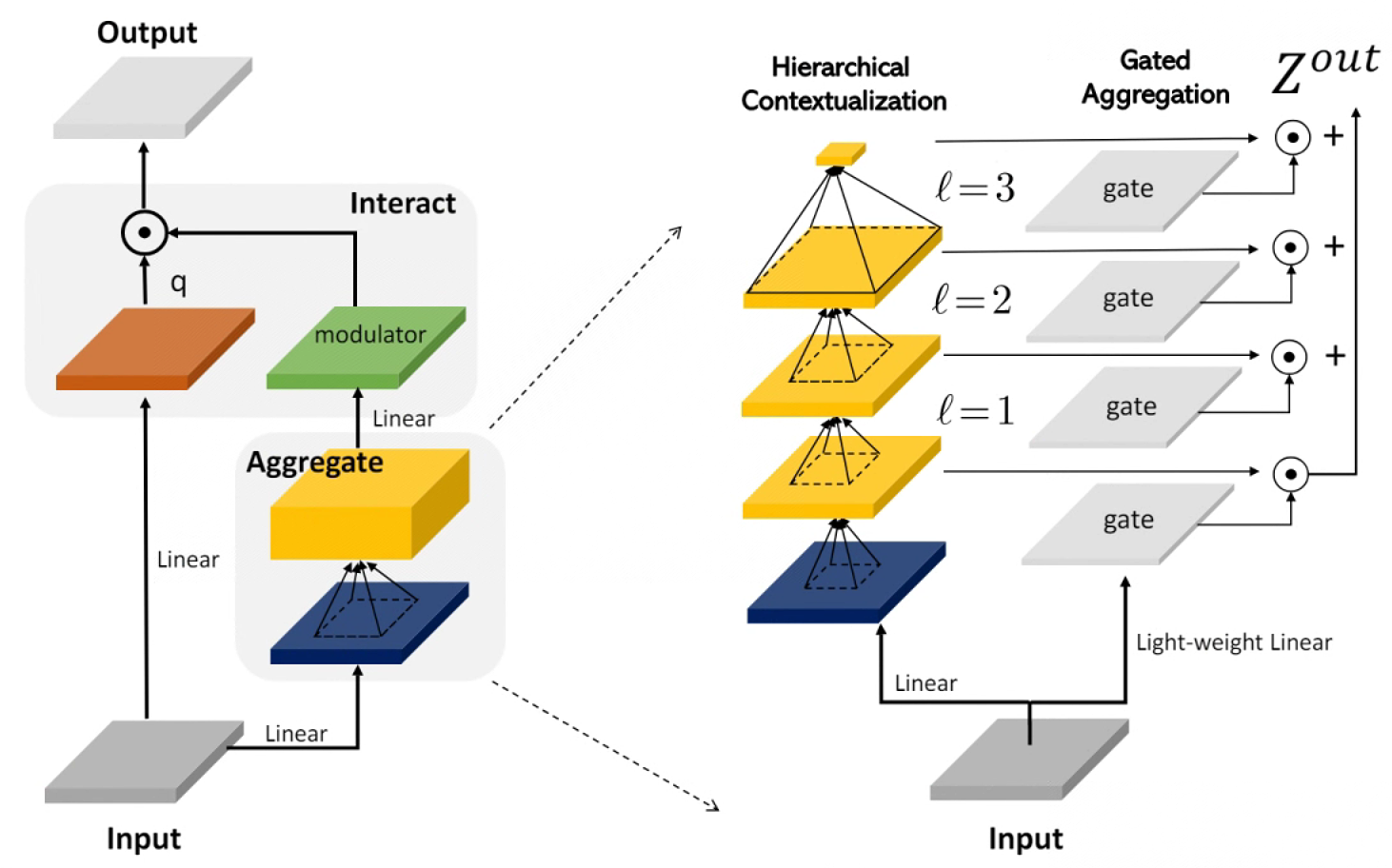}
\caption{Framework of FocalNet~\cite{focalnet}.}
\label{fig:track1_focalnet}
\end{figure}

\noindent{\bf Object detection.}
DINO~\cite{zhang2022dino} is adopted as the detector which uses a contrastive way for denoising training, a mixed query selection method for anchor initialization, and a look forward twice scheme for box prediction in an end-to-end manner, as shown in Figure~\ref{fig:track1_dino}. The most advanced and robust backbone FocalNet-Large~\cite{focalnet} is utilized to extrack informative features, which introduce focal attention to additionally aggregate summarized visual tokens far
away to capture coarse-grained and long-range visual dependencies, as shown in Figure~\ref{fig:track1_focalnet}. 
In order to increase the receptive field of each roi feature, we exploit the roi pooling on the feature map of the corresponding level to get the global context feature, which is used to enhance the roi feature of the corresponding level by adding them. We also add SyncBN to each box head to make the training process more stable.

\begin{figure}[t]
\centering
\includegraphics[scale=0.32]{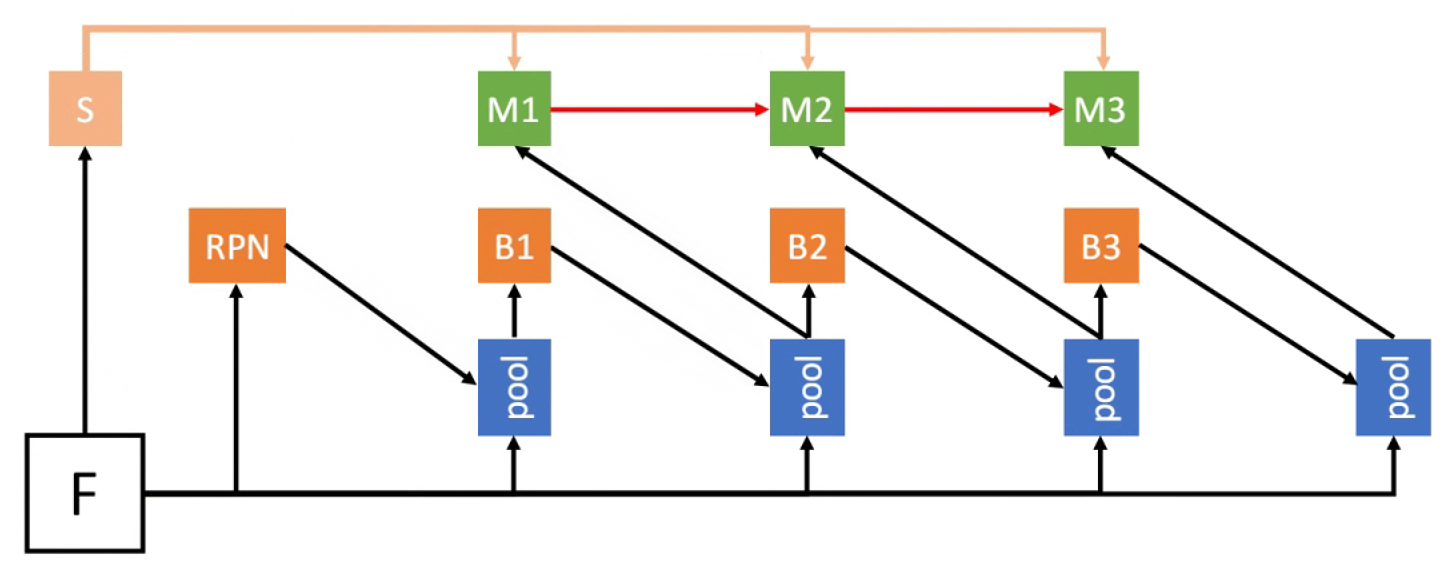}
\caption{Framework of HTC~\cite{HTC}.}
\label{fig:track1_htc}
\end{figure}

\begin{figure}[t]
\centering
\includegraphics[scale=0.23]{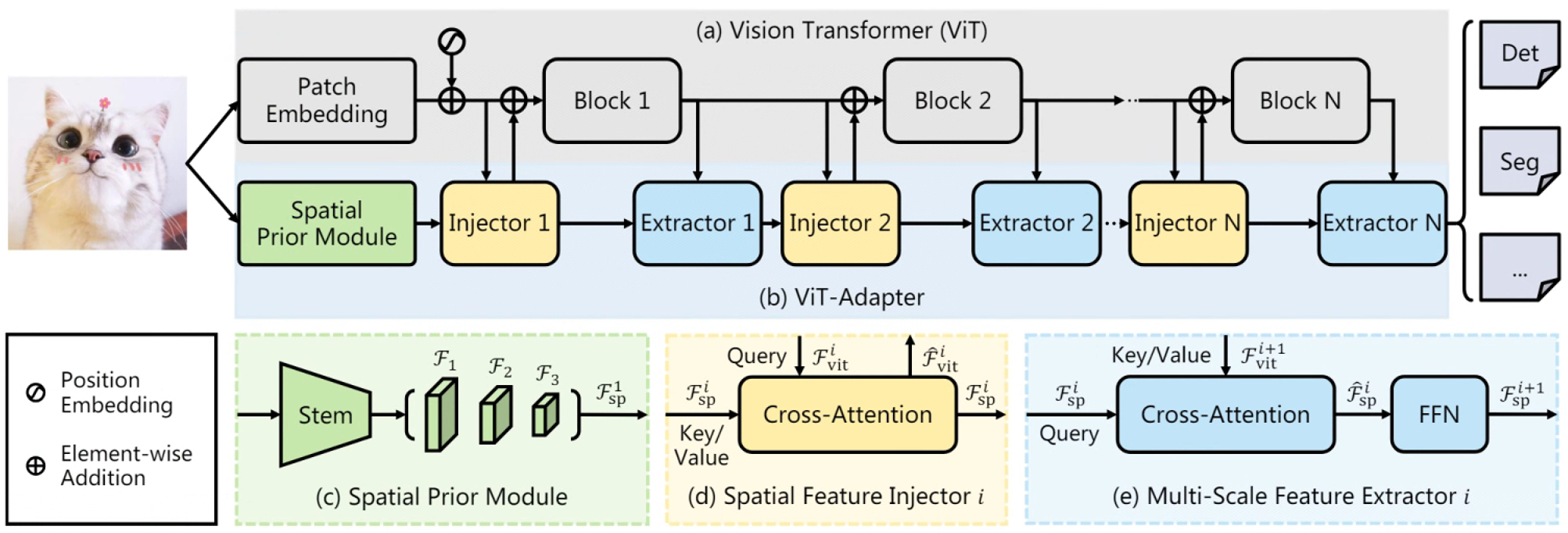}
\caption{Framework of ViT-adapter~\cite{vitadapter}.}
\label{fig:track1_vitadapter}
\end{figure}

\vspace{+0.918mm}
\noindent{\bf Instance segmentation.}
HTC~\cite{HTC} is adopted as our detector which can learn more discriminative features progressively while integrating complementary features together in each stage, as shown in Figure~\ref{fig:track1_htc}. To simplify its use, we directly employ the original masks of objects as semantic maps. The most advanced and robust backbone ViT-adapter~\cite{vitadapter} is utilized to introduce the image-related inductive bias to a plain ViT~\cite{vit}, which allows plain ViT to achieve comparable performance to vision-specific transformers, as shown in Figure~\ref{fig:track1_vitadapter}. 
In order to increase the receptive field of each roi feature, we exploit the roi pooling on the feature map of the corresponding level to get the global context feature, which is used to enhance the roi feature of the corresponding level by adding them. We also add SyncBN to each box head to make the training process more stable.

\begin{figure*}
\centering
\includegraphics[scale=0.55]{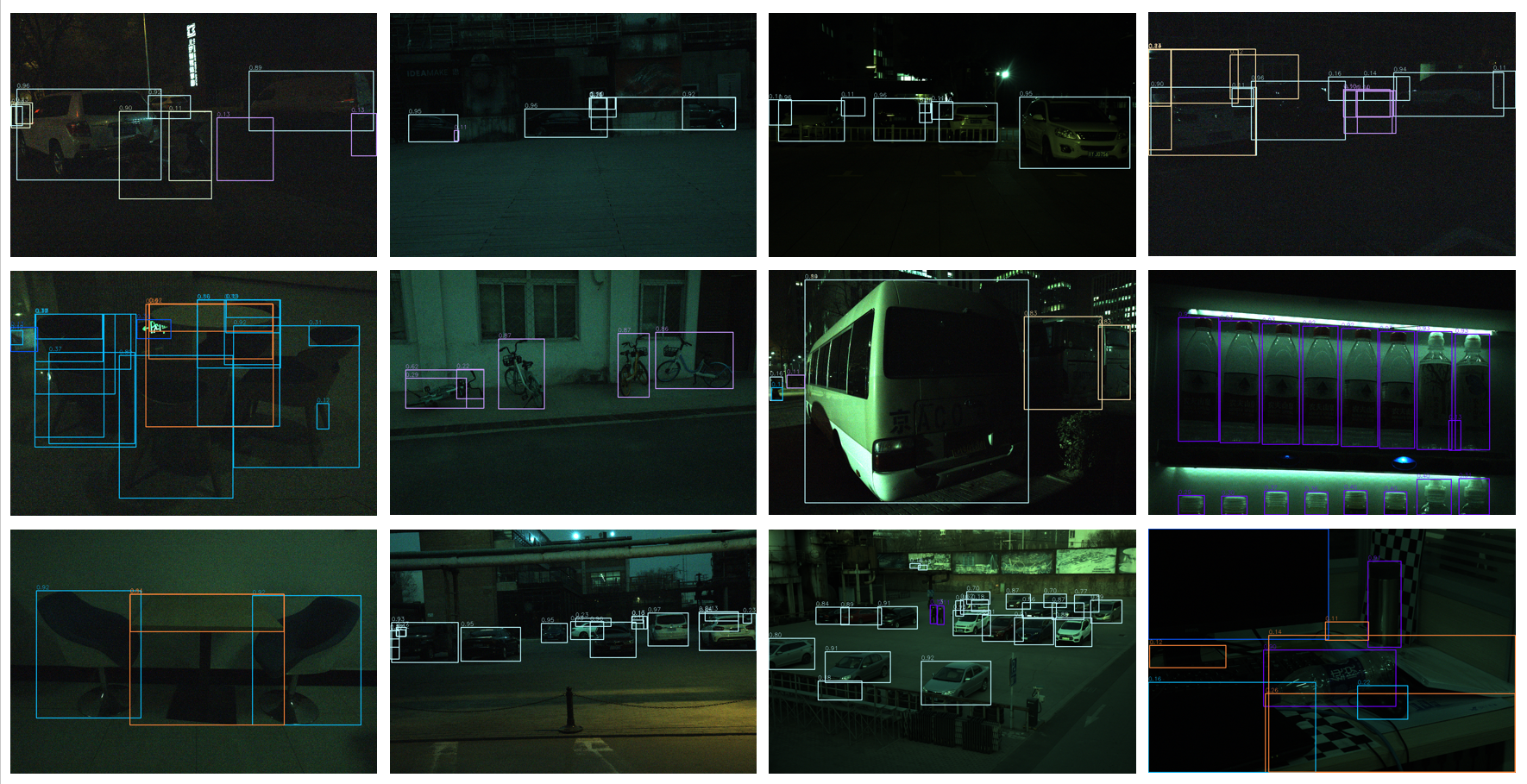}
\caption{Visual results of our method on the testing set.}
\label{fig:track1_od}
\end{figure*}

\subsubsection{Implementation Details}

\noindent{\bf Dataset usage.}
The challenge uses the Low-light Instance Segmentation (LIS) dataset, introduced by \cite{chen2023instance}, which contains 892 labeled images as the training set and 669 images as the testing set. The LIS dataset comprises paired images collected across various scenes, encompassing both indoor and outdoor environments. We utilize all labeled data for training and do not perform online evaluations during training. After training, we directly use the last checkpoint to predict the testing data.

\vspace{+0.918mm}
\noindent{\bf Training details.}
During training, we take the model pre-trained on the Object365 dataset and finetuned on the COCO dataset as the pre-trained model. Specifically, our model is trained on 8 NVIDIA Tesla V100-32G with a total batch size of 8, numbers of queries of 900, and numbers of proposals of 100. Since the training set is small, we train the detector using the AdamW optimizer with an initial learning rate of 0.0001 and weight decay of 0.0001, to alleviate overfitting. We employ the standard 1$\times$ schedule to train the model, and random horizontal flipping with a probability of 0.5 and random resize-crop-resize are introduced as weak augmentation.

\vspace{+0.918mm}
\noindent{\bf Testing details.}
During testing, simple test-time augmentation like horizontal flipping and multi-scale testing are exploited, in which the scales include $\times$1.0, $\times$1.125, $\times$1.25, $\times$1.375, and $\times$1.5. The NMS is not adopted and the detector directly outputs 100 box predictions end to end. Specifically, the initial test image size is 1333x800, and horizontal flipping is adopted to boost model performance. After obtaining ten predictions with different scale augmentation, we further use weighted boxed fusion (WBF)~\cite{wbf} to ensemble them as our final submission, which achieves an AP of 0.76 in the test phase.

In addition, we attempt to introduce some advanced low-light image enhancement methods, such as CIDNet~\cite{CIDNet}, GlobalDiff~\cite{GlobalDiff}, and Retinexformer~\cite{Retinexformer}, to enhance the challenge data, and perform detection algorithm on the enhanced images. Unfortunately, the performance has not been improved or even decreased. We argue that since the challenge dataset does not have pairs of low-light and normal scene images, this leads us to use these image enhancement methods for cross-domain inference, which corrupts the distributional information in the data itself, and ultimately leads to a degradation of detection performance.

Some visual results of our method on the testing set are shown in Figure~\ref{fig:track1_od}.


\subsection{Xocean Team's Method}
\subsubsection{Network Architecture}
\noindent{\bf Object detection.}
Several prior studies \cite{2023lis,Hong2021Crafting,fu2022gan,chen2022hybrid} have endeavored to enhance image cognition performance in extreme conditions. Despite demonstrating superior efficacy compared to their respective baselines, we have observed that employing conventional methodologies on the dataset in this challenge yields comparable effectiveness while being straightforward to implement. Consequently, we adopt a simplified approach by treating the low-light images from the challenge dataset as conventional RGB images.

\begin{figure}
    \centering
    \includegraphics[width=0.5\textwidth]{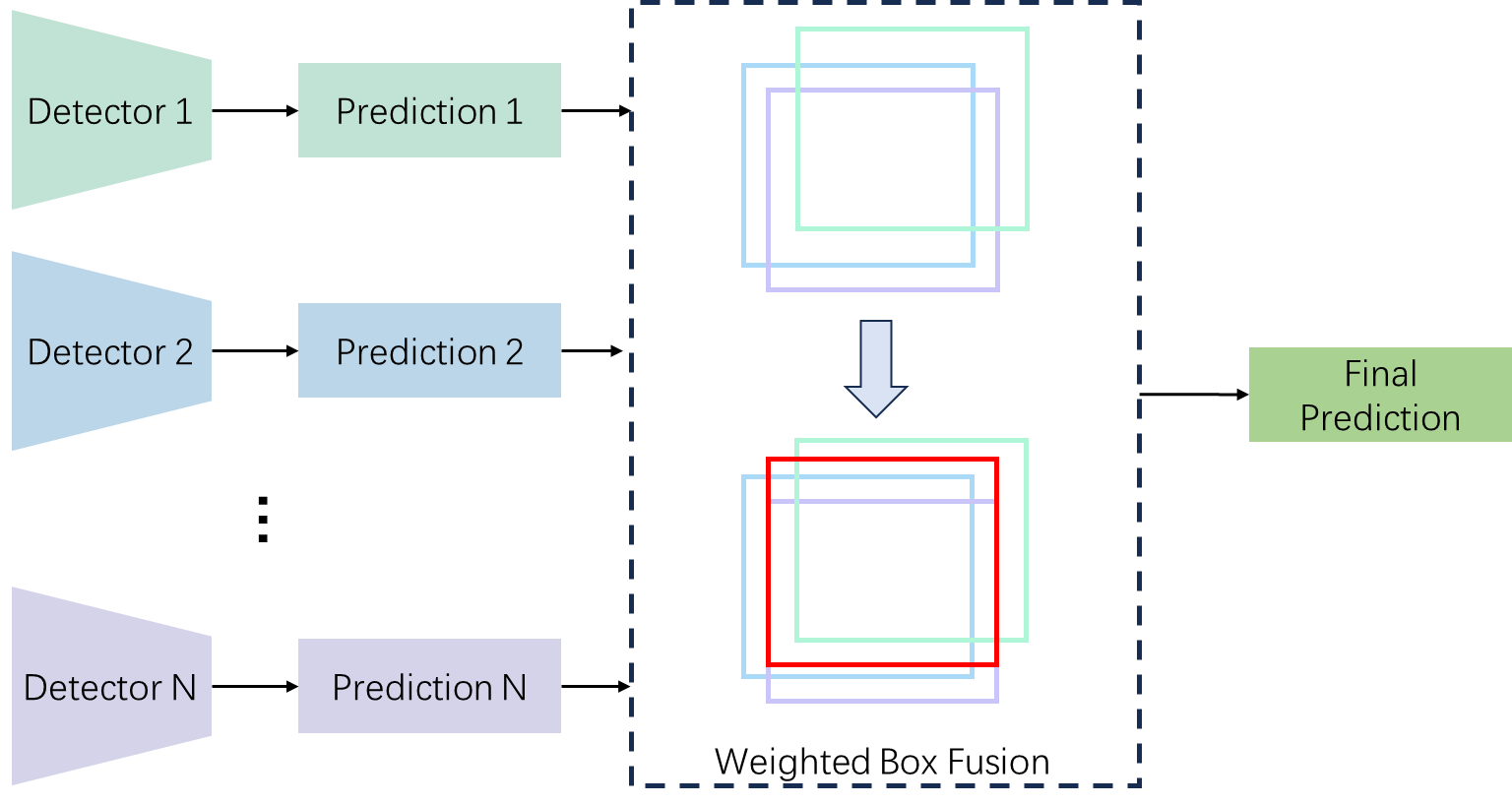}
    \caption{The overview of the proposed object detection framework.we trained several detectors, including RTMDet \cite{lyu2022rtmdet}, YOLOX \cite{ge2021yolox}, Dino \cite{zhang2022dino} and Co-DETR \cite{zong2023detrs} on the challenge datasets, and then ensemble the predictions from those models to achieve better results. We employed Weighted Box Fusion \cite{solovyev2021weighted} as our ensemble method.}
    \label{fig:track1_framework}
    \vspace{-3mm}
\end{figure}

As shown in Figure~\ref{fig:track1_framework}, we trained several detectors, including RTMDet \cite{lyu2022rtmdet}, YOLOX \cite{ge2021yolox}, Dino \cite{zhang2022dino} and Co-DETR \cite{zong2023detrs} on the challenge datasets, and then ensemble the predictions from those models to achieve better results. We employed Weighted Box Fusion \cite{solovyev2021weighted} as our ensemble method.

RTMDet \cite{lyu2022rtmdet} is an efficient real-time object detector that surpasses the YOLO series. Apart from adjusting the number of output classes, we made no modifications to RTMDet. RTMDet-x and RTMDet-l models were chosen due to their high mAP on the COCO dataset.


YOLOX \cite{ge2021yolox} is a highly advanced detector that represents a significant improvement upon the YOLO series. Apart from adjusting the number of output classes, we made no modifications to YOLOX. Taking into account both performance and training costs, we opted for YOLOX-l.


DINO \cite{zhang2022dino} is an advanced end-to-end object detector. Apart from adjusting the number of output classes, we made no modifications to DINO. DINO-Swin-L model was chosen due to its high mAP on the COCO dataset.


Co-DETR \cite{zong2023detrs} is a novel training scheme aimed at improving the efficiency and effectiveness of DETR-based detectors. Apart from adjusting the number of output classes, we made no modifications


\vspace{+0.918mm}
\noindent{\bf Instance segmentation.}
We trained a single RTMDet \cite{lyu2022rtmdet} model for instance segmentation without employing any ensemble methods.

\subsubsection{Implement Details}
\label{dataset}
\noindent{\bf Dataset usage.}
We solely utilized the challenge dataset for training. Additionally, we attempted to augment our training data by incorporating the COCO dataset which was unprocessed according to \cite{brooks2019unprocessing}, preserving annotations with common classes. However, this augmentation did not yield improved results. It is necessary to point out that we still utilized the pretrained weights on the unprocessed \cite{brooks2019unprocessing} COCO dataset to initialize some of the models, aiming to enhance the diversity of our model zoo, which proves advantageous for ensemble methods.

During the initial phase of the challenge, only annotations for the training set were available. Initially, we randomly divided the training set into a proxy training set and a validation set using an 8:2 ratio. Subsequently, we trained the models and optimized the training settings to enhance performance. These settings were then uniformly applied for training on the original complete training dataset, ensuring full utilization of the available data. 

\vspace{+0.918mm}
\noindent{\bf Training.}
To achieve higher performance, we initialized the model weights using pretrained weights from the COCO \cite{lin2014microsoft} Dataset. However, Co-DETR \cite{zong2023detrs} was an exception, as we found that the pretrained weights obtained by training first on Object365 \cite{shao2019objects365} and then on COCO \cite{lin2014microsoft} performed better than those from COCO \cite{lin2014microsoft}.

During the validation and test phases, we retained the weights from the last epoch for evaluation on the official validation and test sets.

We utilized the MMDetection framework \cite{mmdetection} to conduct all experiments on 4 machines, each equipped with 8 NVIDIA RTX 3090/4090 GPUs.

Due to the extensive nature of our training process, which involved training over 18 models for ensemble, providing detailed training configurations in this paper may not be feasible. We recommend referring to the config files in our code repository for more comprehensive information.

\vspace{+0.918mm}
\noindent{\bf Ensemble.}
Table~\ref{tab:track1_wbf} briefly describes the type of model and any specific strategies employed. For example, "Dino-Swin-L" signifies the use of the Dino model with the Swin-L Backbone, while "Dino-Swin-L with TTA" indicates the same model enhanced by test-time augmentation (TTA). Additionally, the descriptions encompass different versions of the RTMDet and Co-DETR models, which may incorporate varying parameters like dropout rates or random seeds during the training phase. "obj2coco" indicates that we use pretrained weights obtained by training first on Object365 \cite{shao2019objects365} and then on COCO \cite{lin2014microsoft} to initialize the parameters of the model. 

These predictions were then utilized in the weighted box fusion to ensemble predictions. The weight of each prediction was determined using a grid search algorithm on the proxy validation set described in~\ref{dataset}.

For more details about the ensemble process, please refer to the configuration files in our code project.

\begin{table}[t]
  \centering
  \setlength{\tabcolsep}{10pt}
  \caption{Ensemble Strategy. "Dino-Swin-L" signifies the use of the Dino model with the Swin-L Backbone, while "Dino-Swin-L with TTA" indicates the same model enhanced by test-time augmentation (TTA). Additionally, the descriptions encompass different versions of the RTMDet and Co-DETR models, which may incorporate varying parameters like dropout rates or random seeds during the training phase. "obj2coco" indicates that we use pretrained weights obtained by training first on Object365 \cite{shao2019objects365} and then on COCO \cite{lin2014microsoft} to initialize the parameters of the model.}
    \begin{tabular}{ccc}
    \toprule
    ID    & Weight & Description \\
    \midrule
    1     & 1     & Dino-Swin-L \\
    2     & 1     & Dino-Swin-L with TTA \\
    3     & 1     & RTMDet-l \\
    4     & 1     & RTMDet-l \\
    5     & 1     & RTMDet-l \\
    6     & 1     & RTMDet-l \\
    7     & 1     & RTMDet-l \\
    8     & 1     & RTMDet-l \\
    9     & 1     & RTMDet-x \\
    10    & 1     & RTMDet-l \\
    11    & 1     & RTMDet-l \\
    12    & 1     & RTMDet-l \\
    13    & 1     & RTMDet-l \\
    14    & 1     & YOLOX-l with TTA \\
    15    & 8     & Co-DETR \\
    16    & 8     & Co-DETR \\
    17    & 10    & Co-DETR-dropout0.6-obj2coco \\
    18    & 10    & Co-DETR-dropout0.3 \\
    \bottomrule
    \end{tabular}%
  \label{tab:track1_wbf}%
\end{table}%

As shown in Table~\ref{tab:track1_result}, our ensemble method for the Object Detection track attained a mean Average Precision (mAP) of 0.76. Additionally, our RTMDet model for the Instance Segmentation track achieved an mAP of 0.58.
\begin{table}[htbp]
  \centering
  \setlength{\tabcolsep}{12pt}
  \caption{Results of our methods}
    \begin{tabular}{ccc}
    \toprule
          & mAP   & mAP50 \\
    \midrule
    Object Detection & 0.76  & 0.89 \\
    Instance Segmentation & 0.58  & 0.79 \\
    \bottomrule
    \end{tabular}%
  \label{tab:track1_result}%
\end{table}%


\subsection{UnoWhoiam Team's Method}
\subsubsection{Network Architecture}
\noindent{\bf Object detection.}
We utilized DINO as our foundational network.
As shown  in Figure~\ref{fig:track1_dino}, DINO is an advanced end-to-end Transformer detector that employs several innovative techniques, including contrastive denoising training, look forward twice, and mixed query selection. These techniques significantly enhance both training efficiency and detection performance.
We chose DINO for our competition due to its demonstrated efficiency and robustness in handling complex detection tasks. Its high performance on benchmark datasets make it an ideal choice for achieving competitive results in the specific task of ``Low-light Object Detection and Instance Segmentation" competition.

\begin{figure}
\centering
\includegraphics[width=0.9918\linewidth]{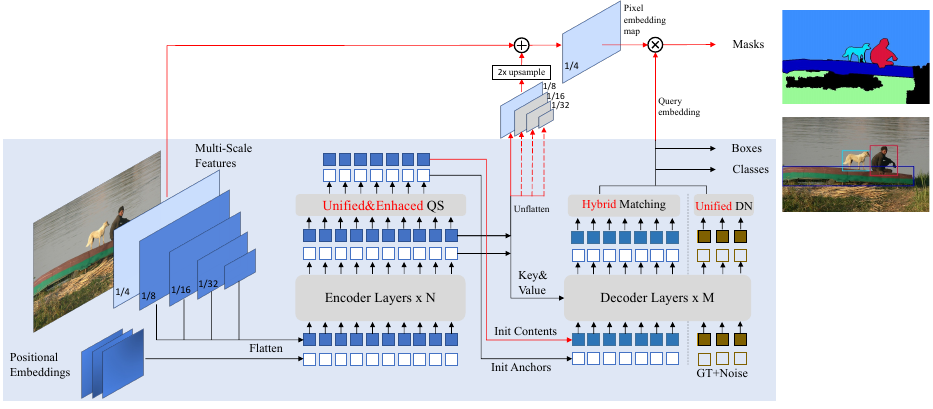}
\caption{Framework of Mask DINO~\cite{2023maskdino}.}
\label{fig:track1_maskdino}
\end{figure}

\vspace{+0.918mm}
\noindent{\bf Instance segmentation.}
We utilized Mask DINO~\cite{2023maskdino} as our foundational network. 
As shown in Figure~\ref{fig:track1_maskdino},
Mask DINO is a unified Transformer-based framework designed for both object detection and image segmentation. This network is an extension of DINO, which was originally developed for detection, and adapts it to handle segmentation tasks with minimal modifications to key components. Mask DINO stands out due to its superior performance, outperforming previous specialized models and achieving the best results in instance, panoptic, and semantic segmentation tasks among models with fewer than one billion parameters.

One of the critical advantages of Mask DINO is its ability to enable task cooperation, demonstrating that detection and segmentation can mutually enhance each other within query-based models. Additionally, Mask DINO leverages better visual representations pre-trained on large-scale detection datasets to improve semantic and panoptic segmentation. This synergistic approach not only enhances the performance but also provides a robust and versatile framework capable of handling multiple vision tasks effectively. By employing Mask DINO, we aim to leverage these strengths to achieve superior results in the ``Low-light Object Detection and Instance Segmentation" competition.

\vspace{+0.918mm}
\noindent{\bf Feature alignment.}
We integrated the Feature-aligned Pyramid Network (FaPN)~\cite{2021fapn} to enhance our network for both object detection and instance segmentation. FaPN is a simple yet effective top-down pyramidal architecture designed to generate multi-scale features for dense image prediction. 
FaPN comprises two key modules: a feature alignment module and a feature selection module. The feature alignment module learns transformation offsets of pixels to contextually align upsampled higher-level features, while the feature selection module emphasizes lower-level features rich in spatial details. Empirical results show that FaPN consistently and substantially improves performance over the original FPN across four dense prediction tasks and three datasets.

We chose FaPN for our competition due to its demonstrated ability to improve multi-scale feature generation. Its integration into our network aims to leverage these strengths, thereby enhancing our model's accuracy in the competition.



\begin{figure}
\centering
\includegraphics[width=0.9918\linewidth]{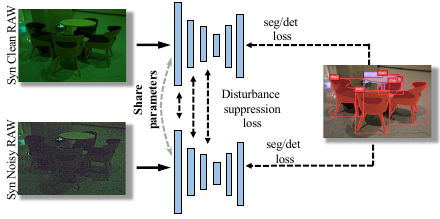}
\caption{Framework of Disturbance Suppression Learning~\cite{chen2023instance}.}
\label{focalnet}
\end{figure}

\subsubsection{Training and Testing Details}

%

\noindent{\bf Training details.}
During training, we use a model pre-trained on the Object365 dataset and fine-tuned on the COCO dataset as our base. Our training setup includes 8 RTX 3090 GPUs, with a total batch size of 8. All other settings are kept the same as in the original paper.
We follow the standard 1$\times$ training schedule and apply weak data augmentation techniques, including random horizontal flipping with a probability of 0.5 and random resize-crop-resize.

\iftrue
\vspace{+0.918mm}
\noindent{\bf Disturbance suppression learning.}
When fine-tuned on COCO, we utilize the low-light RAW synthetic pipeline from~\cite{chen2023instance}, which consists of two steps, namely, unprocessing and noise injection, to obtain synthetic low-light clean/noisy RAW images.
We adopt disturbance suppression learning from previous work~\cite{chen2023instance}.
Ideally, a robust network should extract similar features whether the input image is corrupted by noise or not. To achieve this, we introduce disturbance suppression learning, which encourages the network to learn disturbance-invariant features during training. This approach is independent of architectural considerations.

The total loss for learning is defined as:
\begin{equation}
\begin{aligned}
\color{black}
	L(\theta) = L_{\text{IS}}(x;\theta) 
	+ \alpha L_{\text{IS}}(x';\theta) + \beta L_{\text{DS}}(x, x';\theta),
\end{aligned}
\end{equation}
where \( x \) is the clean synthetic RAW image, \( x' \) is its noisy version, and \( \alpha \) and \( \beta \) are the weights of the respective losses. We empirically set \( \alpha = 1 \) and \( \beta = 0.01 \).

The loss \( L_{\text{IS}} \) is the task loss, \eg, instance segmentation loss, which consists of classification loss, bounding box regression loss, and segmentation (per-pixel classification) loss. The specific formula for \( L_{\text{IS}} \) is related to the model, we employ the same loss as the origianl model.
This loss is applied to both the clean image \( x \) and the noisy image \( x' \) to ensure the model performs consistently regardless of noise.

The loss \( L_{\text{DS}} \) is the feature disturbance suppression loss, defined as:
\begin{equation}
\begin{aligned}
\vspace{-2mm} 
\color{black}
	L_{\text{DS}}(x, x';\theta) = \sum_{i=1}^{n} 
	\lVert f^{(i)}(x;\theta) - f^{(i)}(x';\theta)\rVert^2_2,
\vspace{-2mm}
\end{aligned}
\end{equation}
where \( f^{(i)}(x;\theta) \) represents the \( i \)-th stage of feature maps of the model. By minimizing the Euclidean distance between the clean features \( f^{(i)}(x;\theta) \) and the noisy features \( f^{(i)}(x';\theta) \), the disturbance suppression loss encourages the model to learn disturbance-invariant features. This reduces feature disturbance caused by image noise and improves the model's robustness to corrupted low-light images.

Unlike perceptual loss~\citep{2020qis}, our approach does not require pretraining a teacher model, making our training process simpler and faster. With \( L_{\text{IS}}(x;\theta) \) and \( L_{\text{IS}}(x';\theta) \), our model can learn discriminative features from both clean and noisy images, maintaining stable accuracy regardless of noise. In contrast, the 	``student" model in perceptual loss~\citep{2020qis} only sees noisy images, which can degrade performance on clean images and limit robustness. Additionally, the domain gap between the feature distributions of the teacher and student models can harm the learning process. By minimizing the distance between clean and noisy features predicted by the same model, we avoid this problem.

\fi

\vspace{+0.918mm}
\noindent{\bf Testing details.}
During testing, we employ simple test-time augmentation techniques such as horizontal flipping and multi-scale testing. The multi-scale testing involves resizing the shorter side of the image to various sizes: 400, 500, 600, 700, 800, 900, 1000, 1100, and 1200 pixels. 
Horizontal flipping is also used to enhance model performance. 
For detection, after obtaining ten predictions with different scale augmentations, we use Weighted Box Fusion (WBF)~\cite{wbf} to ensemble them for our final submission. 



\subsection{Teams and Affiliations}
\noindent\textbf{GroundTruth}

\noindent\textbf{Title:} Technique Report of Team GroundTruth for CVPR 2024 PBDL Challenge Low-light Object Detection and Instance Segmentation

\noindent\textbf{Members:} 
Xiaoqiang Lu (\href{mailto:xqlu@stu.xidian.edu.cn}
{xqlu@stu.xidian.edu.cn}), Licheng Jiao, Fang Liu, Xu Liu, Lingling Li, Wenping Ma, Shuyuan Yang


\noindent\textbf{Affiliations:} 
School of Artificial Intelligence, Xidian University

\vspace{1em}

\noindent\textbf{Xocean}

\noindent\textbf{Title:}
Tech Report of Low-light Object Detection and Instance Segmentation Challenge

\noindent\textbf{Members:} 
Haiyang Xie\textsuperscript{1,7} (\href{mailto:whuocean@whu.edu.cn}
{whuocean@whu.edu.cn}), 
Jian Zhao\textsuperscript{6,7}, Shihua Huang\textsuperscript{2}, Peng Cheng\textsuperscript{3}, 
Xi Shen\textsuperscript{2}, Zheng Wang\textsuperscript{1}, Shuai An\textsuperscript{5}, Caizhi Zhu\textsuperscript{2}, Xuelong Li\textsuperscript{4}

\noindent\textbf{Affiliations:}  
\textsuperscript{1}School of Computer Science, Wuhan University, 
\textsuperscript{2}Intellindust, 
\textsuperscript{3}Beijing Forestry University, \textsuperscript{4}Institute of AI (TeleAI), China Telecom, 
\textsuperscript{5}Harbin Institute of Technology, \textsuperscript{6}School of Artificial Intelligence, Optics and Electronics (iOPEN), Northwestern Polytechnical University, 
\textsuperscript{7}EVOL Lab, Institute of AI (TeleAI), China Telecom

\vspace{1em}
\noindent\textbf{UnoWhoiam}

\noindent\textbf{Title:}
Technique Report of Team UnoWhoiam for CVPR 2024 PBDL Challenge Low-light Object Detection and Instance Segmentation

\noindent\textbf{Members:} 
Linwei Chen\textsuperscript{1} (\href{mailto:chenlinwei@bit.edu.cn}
{chenlinwei@bit.edu.cn}), Ying Fu\textsuperscript{1}, Tao Zhang\textsuperscript{2}, Liang Li\textsuperscript{2}, Yu Liu\textsuperscript{3}, Chenggang Yan\textsuperscript{2}

\noindent\textbf{Affiliations:} 
\textsuperscript{1}Beijing Institute of Technology,
\textsuperscript{2}Lishui Institute of Hangzhou Dianzi University,
\textsuperscript{3}Tsinghua University
%

\section{Low-light raw video denoising with realistic motion}

Supervised deep-learning methods have shown their effectiveness on raw video denoising in low-light. However, existing training datasets have speciﬁc drawbacks, e.g., inaccurate noise modeling in synthetic datasets, simple motion created by hand or ﬁxed motion, and limited-quality ground truth caused by the beam splitter in real captured datasets. These defects signiﬁcantly decline the performance of network when tackling real low-light video sequences, where noise distribution and motion patterns are extremely complex.
To address this challenge, the CVPR 2024 PBDL Challenge on low-light raw video denoising with realistic motion aims to improve the recovery quality of realistic videos with complex motion.

\begin{table}[t]
	\centering
	\setlength{\tabcolsep}{12pt}
	\caption{Leaderboard of low-light raw video denoising with realistic motion.}
	\label{tab:track_2_result_all}
	\begin{threeparttable}
		\begin{tabular}{cccc}
			\toprule
			\textbf{Rank} & \textbf{Team} & \textbf{PSNR} & \textbf{SSIM} \\ \hline
			1 & ZichunWang & 45.47 & 0.99 \\ 
			2 & wql & 39.06 & 0.96 \\ 
			3 & mmmmmm & 33.64 & 0.88 \\ \bottomrule
		\end{tabular}
	\end{threeparttable}
\end{table}

As shown in the Table \ref{tab:track_2_result_all}, in this TRACK, all the teams achieved great denoising performance. 
The first place team is ZichunWang, with PSNR and SSIM metrics of 45.47 and 0.99. 
The second place team is wql, with PSNR and SSIM metrics of 39.06 and 0.96. 
The third place team is mmmmmm, with PSNR and SSIM metrics of 33.64 and 0.88. 
These results show excellent denoising capabilities for real-world videos, and also demonstrate that the participants excellent ability in designing algorithms for the denoising task, making an important contribution to the future development of video denoising.

\subsection{Low-light Raw Video Denoising Dataset}

\begin{figure*}
\begin{center}
\includegraphics{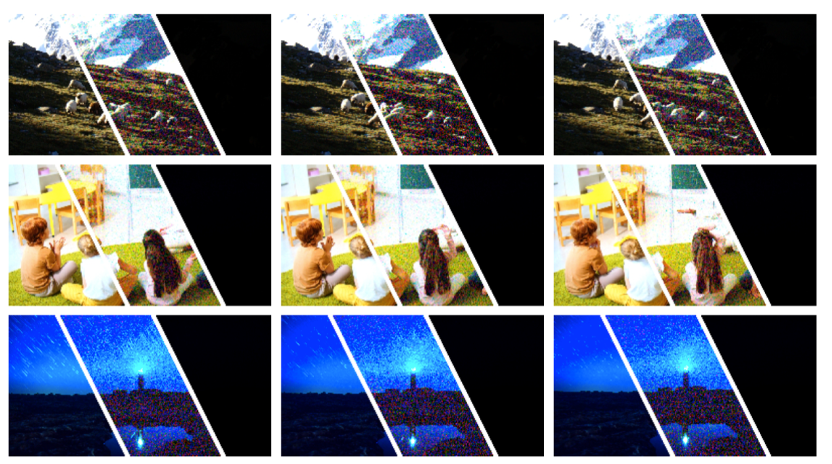}
\end{center}
   \caption{Several representative examples for low/normal-light images in the LLRVD dataset..}
\label{fig:track2_data_show}
\end{figure*}

In this competition, we ﬁrst collect ~70 high-quality 4k videos from the internet, then play them on the DELL U2720QM monitor. We use a Sony Alpha 7R IV full-frame mirrorless camera. The size of the Bayer image is 9504×6336. The scenes of the video clips contain indoor and outdoor, ranging from natural landscapes to extreme sports. This relatively large range of scenes also has an advantage compared to previous datasets. Examples of our data are in Fig.~\ref{fig:track2_data_show}.

\begin{itemize}
	\item \textbf{Realistic scene motion.}
We collect paired low-light raw videos with realistic motion, showing great generalization to the complex scenarios in the real world.

	\item \textbf{210 clips.}
It contains ~210 video pairs, each scene contains three noise levels.

	\item \textbf{High-quality ground truth.}
Previous datasets are all collected in degraded conditions, which may significantly decline the performance of the network trained on them when tackling real scenes. We directly obtain realistic motion in our raw low-light video denoising dataset, featuring high-quality data.

	\item \textbf{No extra equipment.}
Our dataset collecting pipeline requires no extra equipment used in previous datasets.

\end{itemize}

\subsection{ZichunWang Team's Method}

In this section, we show the overall architecture of our proposed method, and describe the basic 3D spatial-temporal self-attention block with convolution.

\subsubsection{Network Architecture}

\vspace{+0.918mm}
\noindent{\bf Overall Pipeline.}
Encoder-decoder is a classic architecture for low-level image tasks, exemplified by U-net \cite{ronneberger2015u}. The main issue for adopting the design of U-net to video denoising is how to efficiently use the redundant temporal information. To align temporal features, existing methods often use an auxiliary module for alignment, including convolution only \cite{claus2019videnn,tassano2020fastdvdnet}, deformable convolution \cite{yue2020supervised}, optical flow \cite{xue2019video}. However, sub-optimal alignment operation may harm its performance.

Besides, most existing methods use convolution for multi-frame features fusion, where the lacking of long-range modeling ability may decline their recovery result. Some methods utilize spatial self-similarity, \textit{e.g.}  \cite{yue2020supervised}, while the abundant temporal-spatial self-similarity in the extra temporal dimension is not fully exploited.

Self-attention is suited for aggregating self-similarity since it can dynamically allocate weight for each pixel. 
To this end, we combine 3D temporal-spatial attention with the hierarchical design of U-net. 
Nonetheless, Transformer may suffer from the deficiency of local feature extraction, which is indispensable for recovering image details. Thus, we combine the locality of convolution with the long-range interaction of self-attention in each Transformer block.

\begin{figure*}
    \centering
\includegraphics[width=0.87\linewidth]{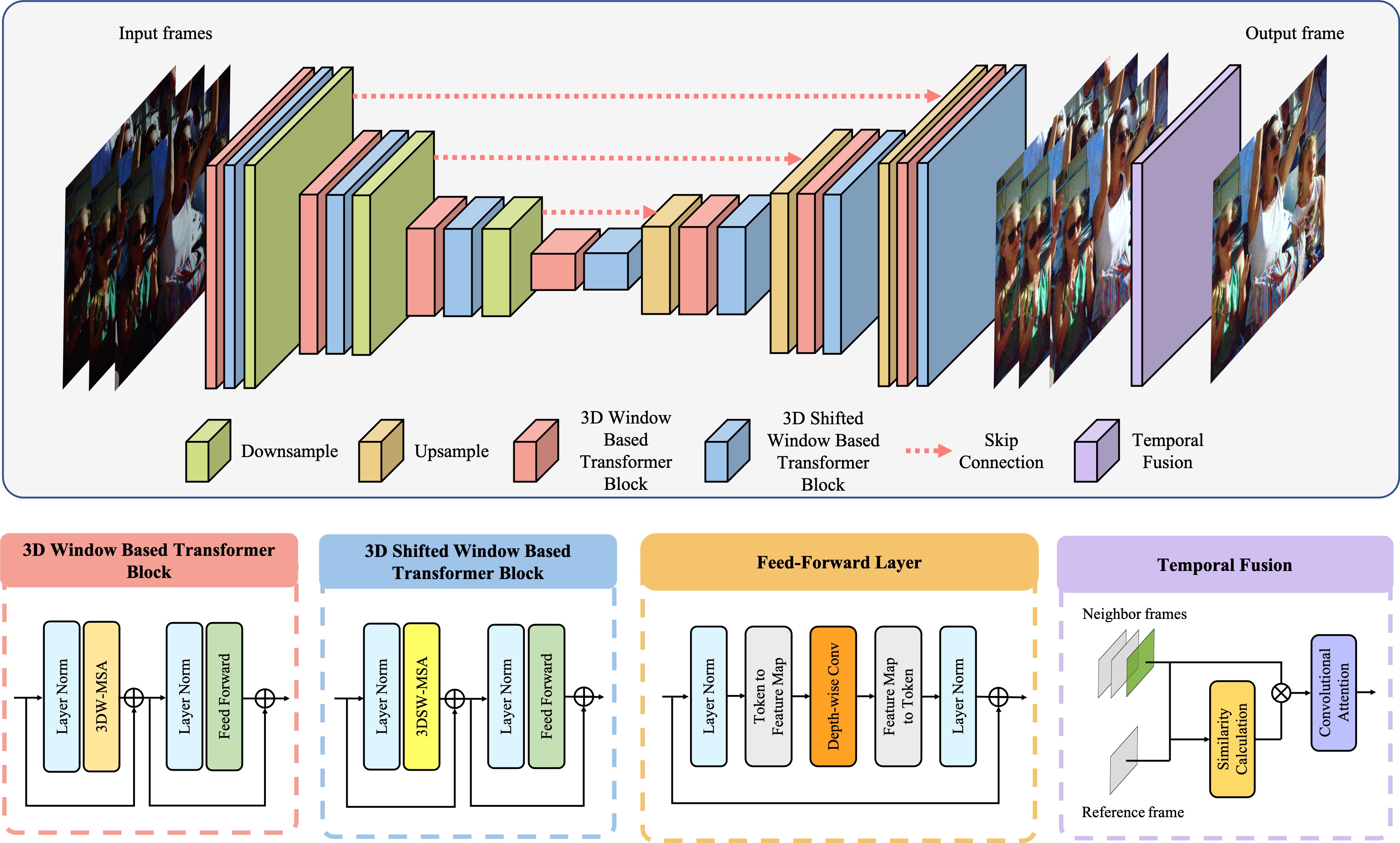}
     \caption{Overview of network architecture. Swin Transformer denotes Shifted Window-based Transformer. 3D (S)W-MSA denotes 3D (Shifted)Window-based Multi-head Self-attention. LN denotes Layer Normalization. Convolutional Attention denotes our final fusion block.
}
     \label{fig:track_2_wzc_network}
  \end{figure*}

The overview architecture of our network is shown in Fig.~\ref{fig:track_2_wzc_network}. We focus on the Raw2Raw video denoising task, where the input and output are all in the raw domain. The input is of size $T \times H \times W \times 4$. $T$ represents the number of input frames, with each frame containing $H \times W \times 4$ pixels in the Bayer pattern. The output frame is of size $H \times W \times 4$. To embed the pixels in images as tokens, we first apply a $3 \times 3$ convolution.
 After embedding, all the tokens pass through $K$ encoders and patch merging layers. Each encoder contains $M$ Shifted Window Transformer blocks. For downsampling, we use the $4\times4$ convolution and double the dimensions. Symmetrically, the decoder includes $K$ Transformer blocks and patch expanding layers. The output of decoder layers is then projected back to image patches. Finally, the extracted multi-frame features are temporally fused to handle the misalignment.

\vspace{+0.918mm}
\noindent{\bf 3D Swin Transformer Block.}
Since vanilla self-attention \cite{dosovitskiy2020image} is computationally consuming, directly adopting it to video denoising is not affordable due to the extra temporal dimension. Besides, Transformer \cite{dosovitskiy2020image} has strong long-range modeling ability but neglects local features, which is vital for recovering details. To extract locality with less computational effort, we apply 3D shifted window-based multi-head self-attention (3DSW-MSA) and 3D window-based multi-head self-attention (3DW-MSA) \cite{liu2021swin}, together with depth-wise convolution in the feed-forward layer. In this way, we can effectively extract the local features by convolution, at the same time fully taking advantage of intrinsic temporal-spatial self-similarity by the long-range modeling ability of the Transformer.

Two consecutive 3D shifted window-based Transformer blocks are computed as:

\begin{equation}
\begin{aligned}
&\hat{\mathbf{z}}^{l}=3 \mathrm{DW}-\mathrm{MSA}\left(\mathrm{LN}\left(\mathbf{z}^{l-1}\right)\right)+\mathbf{z}^{l-1}, \\
&\mathbf{z}^{l}=\mathrm{FFN}\left(\mathrm{LN}\left(\hat{\mathbf{z}}^{l}\right)\right)+\hat{\mathbf{z}}^{l}, \\
&\hat{\mathbf{z}}^{l+1}=3 \mathrm{DSW}-\mathrm{MSA}\left(\mathrm{LN}\left(\mathbf{z}^{l}\right)\right)+\mathbf{z}^{l}, \\
&\mathbf{z}^{l+1}=\mathrm{FFN}\left(\mathrm{LN}\left(\hat{\mathbf{z}}^{l+1}\right)\right)+\hat{\mathbf{z}}^{l+1},
\end{aligned}
\end{equation}
where $\hat{\mathbf{z}}^{l}$ and $\mathbf{z}^{l}$ represent the output features of the 3DW-MSA and 3DSW-MSA for $l^{th}$ block. A LayerNorm (LN) is added before MSA and after the FeedForward layer (FFN). Following the previous studies, we add the relative position encoding $B \in \mathbb{R}^{T^{2} \times M^{2} \times M^{2}}$ to the 3D attention block. The self-attention is computed as:
\begin{equation}
	\begin{aligned}
\text { Attention }(Q, K, V)=\operatorname{SoftMax}\left(Q K^{T} / \sqrt{d}+B\right) V,
	\end{aligned}
\end{equation}
where $Q, K, V \in \mathbb{R}^{T M^{2} \times d}$ are the query, key and value matrices. $d$ is the dimension of the query and key features. $T M^{2}$ is the number of tokens per window. And, the values of $B$ are taken from the 3D bias matrix $\hat{B} \in \mathbb{R}^{(2 T-1) \times(2 M-1) \times(2 M-1)}$, corresponding to the temporal range of $[-T+1, T-1]$ and the spatial range of $[-M+1, M-1]$.

\vspace{+0.918mm}
\noindent{\bf Temporal Fusion.}
After the exploitation of spatial-temporal self-similarity, features in neighbor frames are
fused for the recovery of the reference frame. However, it is not appropriate to simply combine these frames, since the complex motion in real videos makes each neighbor frame contribute variously to the central reference frame. Intuitively, the closer between the features in the neighbor frame and reference frame, the more information a neighbor frame can provide for recovery. Therefore, we first extract the features by embedding, then compute the similarity between the features of each neighbor and the reference features in an embedded space: 
\begin{equation}
	\begin{aligned}
		S\left(F_{t+i}, F_{t}\right)=\operatorname{Sim}\left(\theta\left(F_{t+i}\right)^{T}, \phi\left(F_{t}\right)\right),
	\end{aligned}
\end{equation}
where $\theta$ and $\phi$ are embedding functions. $\operatorname{Sim}$ denotes the similarity calculation function.  
Here we also adopt the dot product following previous work \cite{wang2019edvr} for similarity calculation.
$F_t$ refers to the reference frame and $F_{i+t}$ refers to the neighbor frames where $i \in [-T+1, T-1]$. After getting the similarity matrix, we adaptively re-weight the features in the temporal dimension,
\begin{equation}
	\begin{aligned}
		\tilde{F}_{t+i}={F}_{t+i} \odot S\left({F}_{t+i}, {F}_{t}\right),
	\end{aligned}
\end{equation}
\begin{equation}
	\begin{aligned}
		F_{\text {fusion}}=\operatorname{Conv}([\tilde{F}_{t-T}, \cdots, \tilde{F}_{t}, \cdots, \tilde{F}_{t+T}]),
	\end{aligned}
\end{equation}
where $\odot$ and $[\cdot, \cdot, \cdot]$ denote the element-wise multiplication and concatenation respectively. We then concatenate all the features and gather them together for the reconstructed frame by convolution layer. Finally, a convolutional attention module \cite{woo2018cbam} is used to spatially enhance the feature representation.

\subsection{Wql Team's Method}

Our team, with the username wql on Codalab, achieved a final score of 38.82 on the leaderboard, ranking second. In this report, we will present all the technical details for solving this task. The task of this competition is to denoise and restore low-light raw videos. Considering the low-light characteristics of the data, we divide the task into two subtasks: low-light restoration and denoising. The key to video restoration lies in fully utilizing inter-frame information. After extensive experiments, we determined to use the Shift-Net model for low-light restoration. To avoid compromising the performance of the model, we converted the original data to the RGB format for training. Given that the restored videos still contain a large amount of noise, we applied the RVRT model for denoising again, resulting in high-quality output. Experimental results demonstrate that our strategy is effective, achieving an outstanding score of 38.82 on the LLRVD dataset.

\begin{figure*}[h]
    \centering
    \includegraphics[width=1\linewidth]{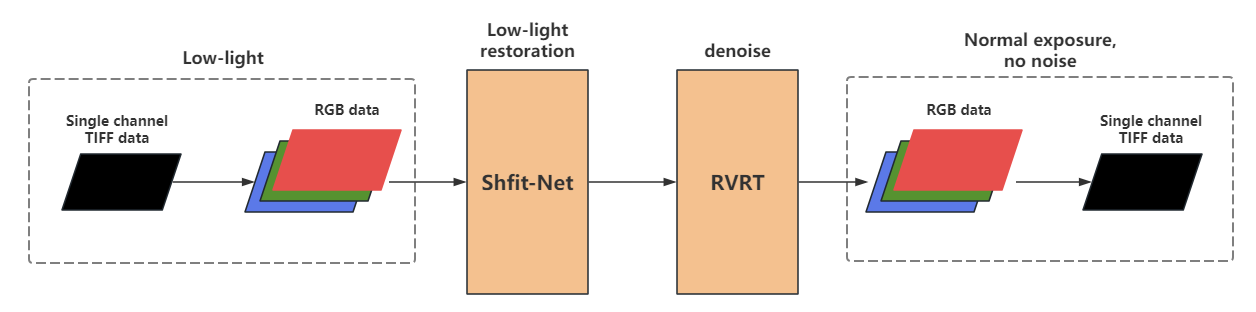}
    \caption{The overall experimental architecture diagram.}
    \label{fig:track_2_overview}
\end{figure*}

Our entire task solution is illustrated in Fig.~\ref{fig:track_2_overview}. For the original low-light video, it is first converted into an RGB format video. Then, it undergoes restoration to normal lighting conditions through the Shift-Net \cite{li2023simple} model. At this point, the video's lighting level is normal, but there still exists a significant amount of noise. Subsequently, it undergoes denoising through RVRT \cite{liang2022recurrent}, resulting in the final video restoration and denoising outcome.

The key to this track relies on the utilization of inter-frame information. Existing deep learning methods often depend on complex network architectures such as optical flow estimation, deformable convolutions, and cross-frame self-attention layers, leading to high computational costs. After extensive literature review, our team ultimately chose Shift-Net as the main model. This model proposes a simple yet effective video restoration and denoising framework, surpassing existing state-of-the-art methods not only in accuracy but also with its parameter count of only two versions, 4.1M and 12.3M, much smaller than existing advanced models. The model is based on grouped spatiotemporal displacements, a lightweight and direct technique that implicitly captures inter-frame correspondences through multi-frame aggregation. By introducing grouped spatial displacements, a broad effective receptive field is obtained, and combined with basic 2D convolutions, this simple framework can effectively aggregate inter-frame information.
Despite the restoration of low-light images using Shift-Net, the images still contain a significant amount of noise. Therefore, our team chose the RVRT model to denoise the restored images, aiming for high-quality video restoration.

\subsubsection{Network Architecture}

\vspace{+0.918mm}
\noindent{\bf Data preprocessing.}
Our team has chosen the Shift-Net model for low-light restoration, which defaults to RGB input format. To maintain the model's performance, we decided to convert the raw images in the dataset to RGB format for training purposes.
The storage format of raw images is RGBG, and since the human eye is more sensitive to green, for better visualization, we extract the RGB three-channel data and multiply R and B by 2, while keeping G unchanged. The processed images are saved in PNG format for model training. Due to the low pixel values in low-light sequences, there may be information loss when saving as PNG, so the conversion process is performed online to mitigate this.

\vspace{+0.918mm}
\noindent{\bf Shift-Net.}
Most previous video restoration methods have employed complex architectures such as optical flow, deformable convolutions, and self-attention layers. This team proposes a simple yet effective grouped spatiotemporal displacement block to implicitly establish temporal correspondence.

\begin{figure}[t]
\centering
\includegraphics[width=1\linewidth]{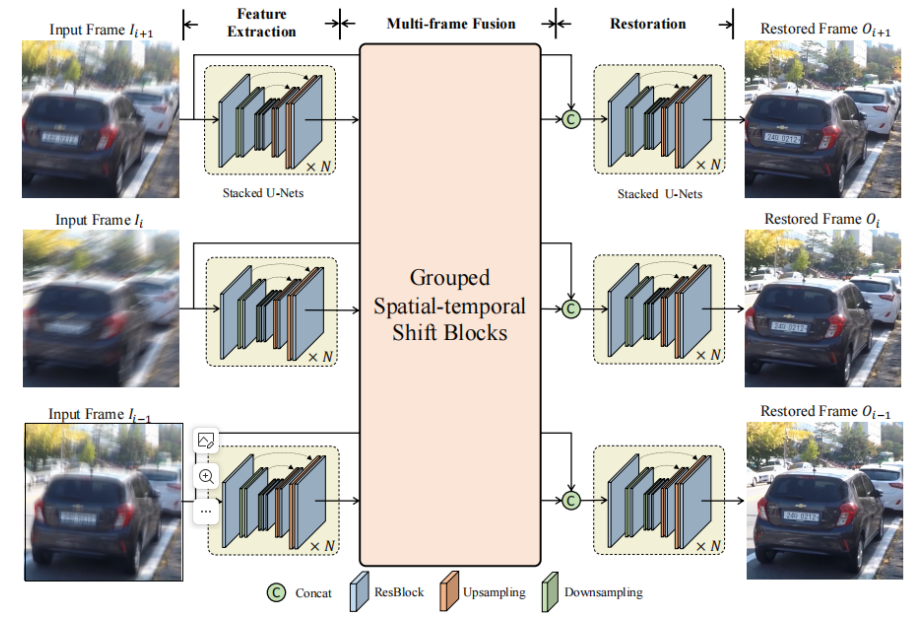}
\caption{Overview of the Group Shift-Net. It adopts a three-stage design: feature extraction, multi-frame fusion, and final restoration. Grouped spatial-temporal shift blocks are proposed to achieve multi-frame aggregation. }
\label{fig:track_2_frog}
\end{figure}

As shown in Fig.~\ref{fig:track_2_frog}, the framework of this model adopts a three-stage design: 1) feature extraction, 2) multi-frame feature fusion with grouped spatiotemporal offsets, 3) final restoration.

\vspace{+0.918mm}
\noindent{\bf Feature extraction.}
Each frame Ii typically suffers from different types of degradation (such as noise or blur), which affects temporal correspondence modeling. A two-dimensional U-Net-like structure is adopted to mitigate the negative impact of degradation and extract frame-level features.

\vspace{+0.918mm}
\noindent{\bf Multi-frame feature fusion.}
At this stage, a grouped spatiotemporal displacement block is proposed to move different features from adjacent frames to the reference frame, implicitly establishing temporal correspondence. Keyframe features are fully aggregated with features from neighboring frames to obtain corresponding aggregate features. By employing spatiotemporal displacements in different directions and distances, multiple candidate displacements are provided for frame matching. By stacking multiple grouped spatiotemporal displacement blocks, our framework achieves long-term aggregation.

\vspace{+0.918mm}
\noindent{\bf Final restoration.}
Finally, similar to the U-Net structure, taking low-quality input frames and corresponding aggregate features as input, the model generates the final result for each frame.

\begin{figure}[t]
    \centering
    \includegraphics[width=1\linewidth]{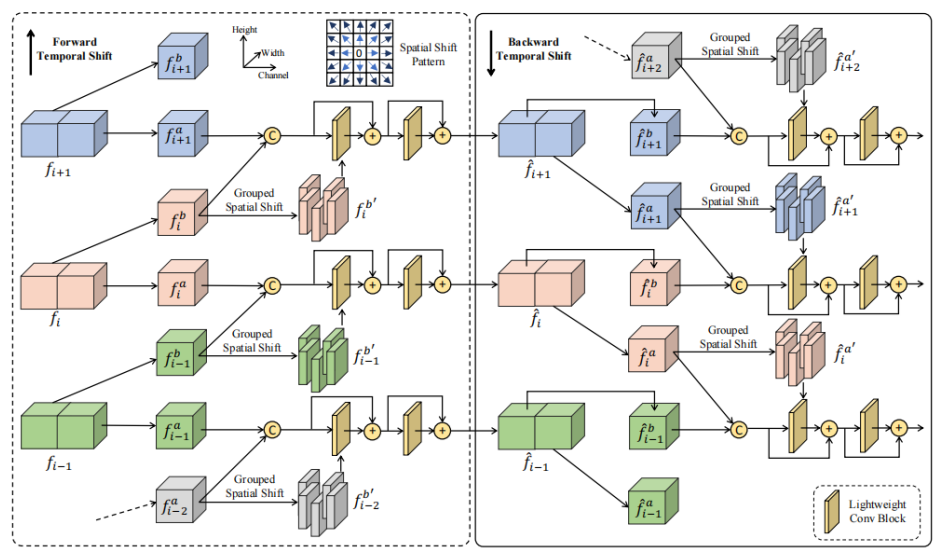}
    \caption{The operations of Grouped Spatial-temporal Shift (GSTS). We stack the forward temporal shift (FTS) blocks (\textit{Left}) and backward temporal shift (BTS) blocks (\textit{Right}) alternatively to achieve bi-directional propagation. Grouped spatial shift provides multiple candidate displacements within large spatial fields and establish temporal correspondences implicitly.}
    
    \label{fig:track_2_GSTS}
\end{figure}

In multi-frame fusion, frame features are aggregated with adjacent features to obtain temporally fused features. We adopt a two-dimensional U-Net structure for multi-frame fusion, maintaining skip connections within the U-Net. Instead of multiple 2D convolutional blocks, we replace them with stacked Grouped Spatiotemporal Shift (GSTS) blocks to effectively establish temporal correspondence and perform multi-frame fusion. GSTS blocks are not applied at the finest scale to save computational costs. The GSTS block consists of three parts: 1) temporal displacement, 2) spatial shift, 3) lightweight fusion layer, as illustrated in Fig.~\ref{fig:track_2_GSTS}.

\begin{figure}[h]
    \centering
    \includegraphics[width=1\linewidth]{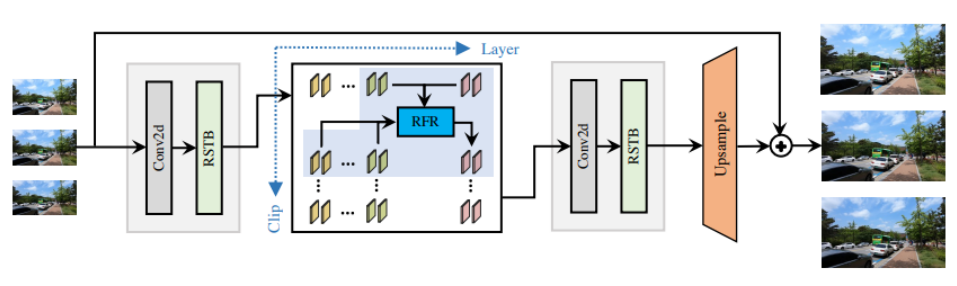}
    \caption{Overall Framework of RVRT}
    \label{fig:track_2_rvrt}
\end{figure}

\vspace{+0.918mm}
\noindent{\bf RVRT.}
RVRT demonstrates excellent performance in the field of video denoising, as shown in Fig.~\ref{fig:track_2_rvrt}. The framework consists of three parts: shallow feature extraction, recurrent feature refinement, and frame reconstruction. Shallow feature extraction utilizes convolutional layers and multiple RSTB blocks from SwinIR to extract features from low-quality videos (LQ). Subsequently, the recurrent feature refinement module performs temporal modeling, and guided deformable attention is employed for video alignment. Finally, multiple RSTB blocks are fed to generate the final features, followed by HQ reconstruction using pixelShuffle.

\subsubsection{Implementation Detail}


\vspace{+0.918mm}
\noindent \textbf{Shift-Net}. We opted for the standard version of the Shift-Net model for training. Since each video in the training set randomly contains three levels of noise, the data reading strategy during training also involves randomly selecting one noise level. The training parameters include a batch size of 4, a learning rate of 4e-4, and 120,000 iterations. The training was conducted using a single NVIDIA RTX 3090 GPU, lasting for 48 hours, without loading pretrained weights.

\vspace{+0.918mm}
\noindent \textbf{RVRT.}
For denoising training with RVRT, only the labels are retrieved during data loading, with a certain amount of noise added. The training parameters include a batch size of 4, a learning rate of 1e-5, and 40,000 iterations. Training was conducted using a single NVIDIA RTX 3090 GPU, lasting for 7 hours, without loading pretrained weights.
%


\subsection{Mmmmmm Team's Method}
In this competition \cite{10003653}, we first attempted video denoising using raw images, employing models such as RViDeNet \cite{yue2020supervised} and EMVD \cite{maggioni2021efficient}. However, due to unsatisfactory results, we later switched to the RGB image-oriented model MIRNetv2 \cite{Zamir2022MIRNetv2}. With a specific training strategy, we achieved our final score.

\subsubsection{Network Architecture}

We employed a multi-scale approach that preserves the original high-resolution features through the network hierarchy, thereby minimizing the loss of precise spatial details. Simultaneously, it encodes multi-scale context by using parallel convolution streams to process features at lower spatial resolutions. The multi-resolution parallel branches operate complementarily to the main high-resolution branch, providing more accurate and contextually enriched feature representations.

One major distinction between MIRNetv2 and other multi-scale image processing methods lies in how contextual information is aggregated. While other methods focus on processing each scale separately, MIRNetv2 progressively exchanges and fuses information from coarse to fine resolution levels. Additionally, unlike methods that use simple concatenation or averaging of features from multi-resolution branches, MIRNetv2 introduces a new selective kernel fusion approach. This approach dynamically selects the useful set of kernels from each branch representation using a self-attention mechanism. Moreover, the proposed fusion block combines features with varying receptive fields while preserving their distinctive complementary characteristics.

The MIRNetv2 network is divided into four modules: Dual-Pixel Defocus Deblurring, Image Denoising, Image Super-Resolution, and Image Enhancement. Among them, I employed the Dual-Pixel Defocus Deblurring module for my task.

\vspace{+0.918mm}
\noindent{\bf Dual-Pixel Defocus Deblurring.}
Images captured with a wide aperture have a shallow depth of field, meaning that regions outside the depth of field become out of focus. Given an image with defocus blur, the goal of defocus deblurring is to generate a globally sharp image. Existing defocus deblurring methods either directly deblur images or first estimate the defocus disparity map and then use it to guide the deblurring process. Modern cameras are equipped with dual-pixel sensors, where each pixel location has two photodiodes, thereby generating two sub-aperture views. The phase difference between these views is useful for measuring the amount of defocus blur at each scene point. Recently, Abuolaim et al. introduced a dual-pixel deblurring dataset (DPDD) and a new method based on encoder-decoder design. In this paper, our focus is also on directly using dual-pixel data to deblur images. Previous defocus deblurring works have employed encoder-decoder architectures that repeatedly use downsampling operations, resulting in significant loss of important details. In contrast, the architectural design of our method enables the preservation of texture details required for the restored image.

\begin{figure}[t]
\centering
\includegraphics[width=1\linewidth]{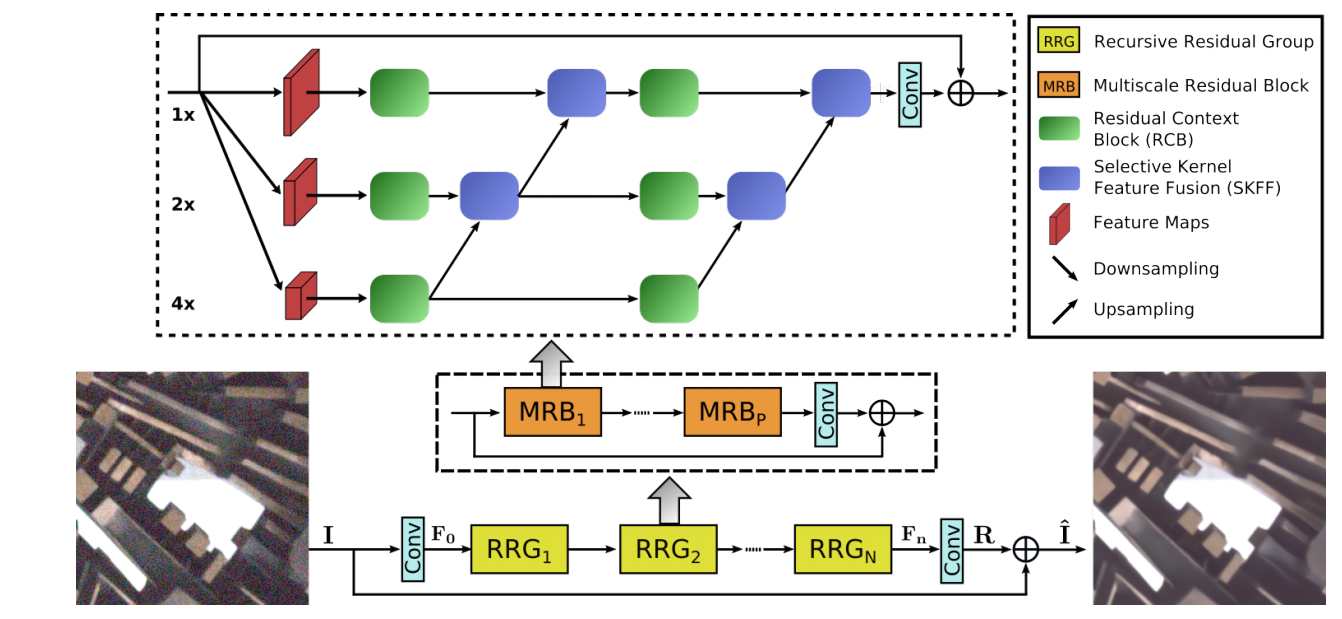}
\caption{\label{fig:track_2_yxy_network}
    The proposed MIRNet-v2 framework is aimed at learning enriched feature representations for image restoration and enhancement. Based on a recursive residual design, MIRNet-v2 comprises multiple-scale residual blocks (MRBs), with its main branch dedicated to maintaining spatially precise high-resolution representations throughout the entire network, while complementary parallel branches provide better contextual features.}
\end{figure}

\vspace{+0.918mm}
\noindent{\bf Visualization.}
According to the ISP process provided on the official competition website, we processed the TIFF images in RGGB order, performed grayscale balancing correction, and added an additional step of normalization before outputting PNG images to make the RGB images appear clearer and brighter. The specific process is illustrated in the following Fig.~\ref{fig:track_2_yxy_pipeline}.

\begin{figure*}[t]
\centering
\includegraphics[width=1\linewidth]{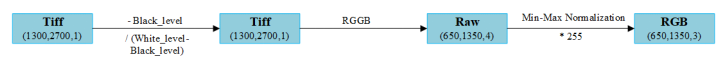}
\caption{\label{fig:track_2_yxy_pipeline}Data Processing Pipeline}
\end{figure*}

\begin{figure}[t]
\centering
\includegraphics[width=0.75\linewidth]{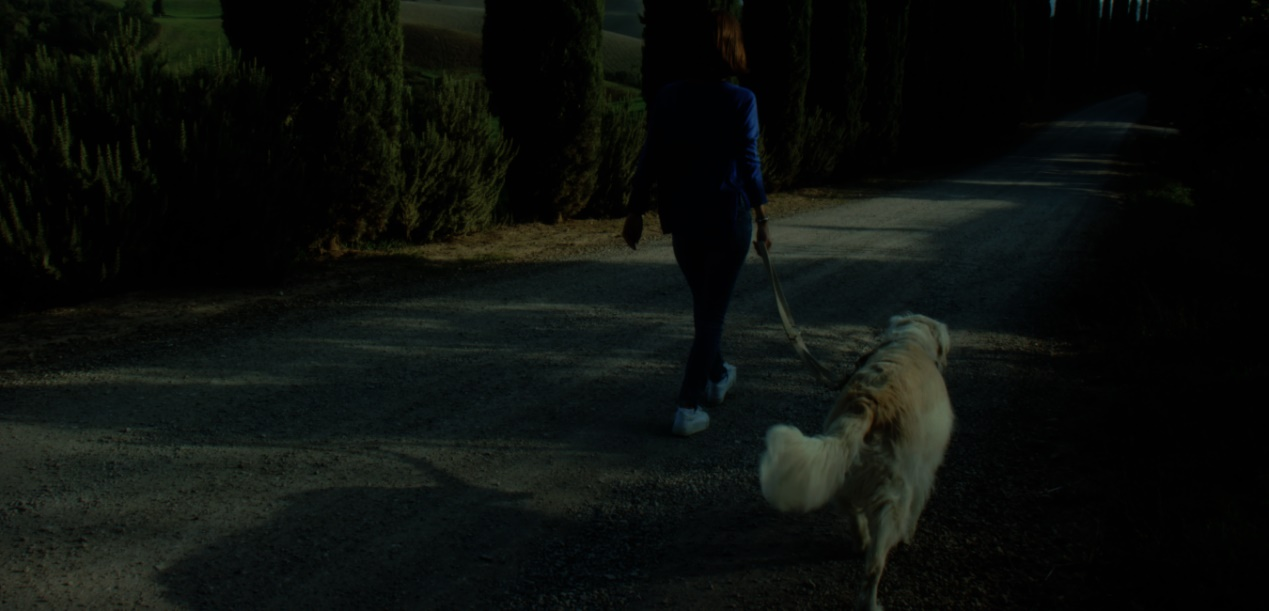}
\caption{\label{fig:track_2_yxy_dog1}GT Direct Visualization}
\end{figure}

\begin{figure}[t]
\centering
\includegraphics[width=0.75\linewidth]{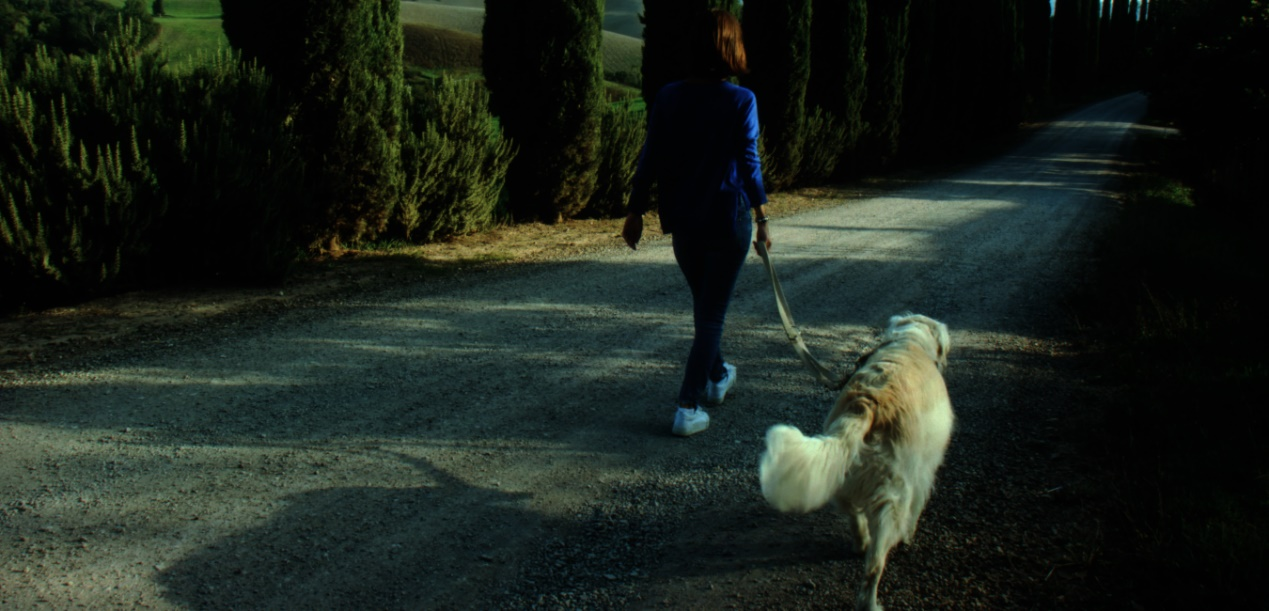}
\caption{\label{fig:track_2_yxy_dog2}Normalized Images}
\end{figure}

\subsubsection{Implementation Details}

Training dataset contains a total of 72 scenes. The noisy dataset for each scene contains 10 consecutive images with three different noise levels. The validation set consists of scenes 3 and 4, with images containing noise levels of 125, 160, and 200. The test set includes scenes 16, 22, 36, and 66, each with ten images at noise levels of 100, 125, 160, 200, 250, and 320. Image size is (1300, 2700, 1), formatted as TIFF images arranged in RGGB order. Black level is 512, and white level is 15360. Specific parameter settings for the training process:

\vspace{+0.918mm}
\noindent{\bf Dataset setup.} The dataset includes training and validation sets for model training and evaluation.
    Double-pixel depth images are used for model training with geometric augmentation.
    Different training batch sizes (8, 5, 4, 2, 1, 1) and iteration numbers (92000, 64000, 48000, 36000, 36000, 24000) are set to gradually improve training effectiveness.
    Progressive training strategy is employed, starting from smaller image cropping sizes and gradually increasing the crop size (128, 160, 192, 256, 320, 384).

\vspace{+0.918mm}
\noindent{\bf Network architecture setup.}
  MIRNet\_v2 network architecture is used for the task of deblurring double-pixel depth images.
   The network has 6 input channels and 3 output channels.

\vspace{+0.918mm}
\noindent{\bf Training setup.}
   Total iteration number is set to 300,000, and cosine annealing restarts learning rate scheduler is used.
   Adam optimizer is employed with a learning rate of 2e-4.

\vspace{+0.918mm}
\noindent{\bf Validation setup.}:
   Validation is performed every 2000 iterations, and the PSNR validation metric is calculated.
   Validation images are not saved.

With these training strategies, I trained on the training set for 300,000 iterations and achieved good performance.showing in Table \ref{tab:track_2_yxy_table1}:

\begin{table}[t]
\centering
\setlength{\tabcolsep}{14pt}
\caption{Comparison between the two models}
\label{tab:track_2_yxy_table1}
\begin{tabular}{ccc}
\toprule
 Model & Input & Score \\
\midrule
EMVD & Raw & 29.44 \\
MIRNetv2 & RGB & 27.15 \\
MIRNetv2* & Normalied RGB & 32.94 \\
\bottomrule
\end{tabular}
\end{table}

  From the perspective of input image types, directly training with raw images yields lower scores, as shown in Fig.~\ref{fig:track_2_yxy_network}. This could be due to the resulting test images having blurry details, unclear textures, and a greenish hue. However, when input images are RGB images, the model MIRNetv2 is used for deblurring tasks. When our input images are not normalized, the test results tend to be darker with heavier colors. Then, after normalization, the color changes in the resulting test images are smaller, and the details and textures become clearer and more visible.

\begin{figure}[t]
\centering
\includegraphics[width=0.75\linewidth]{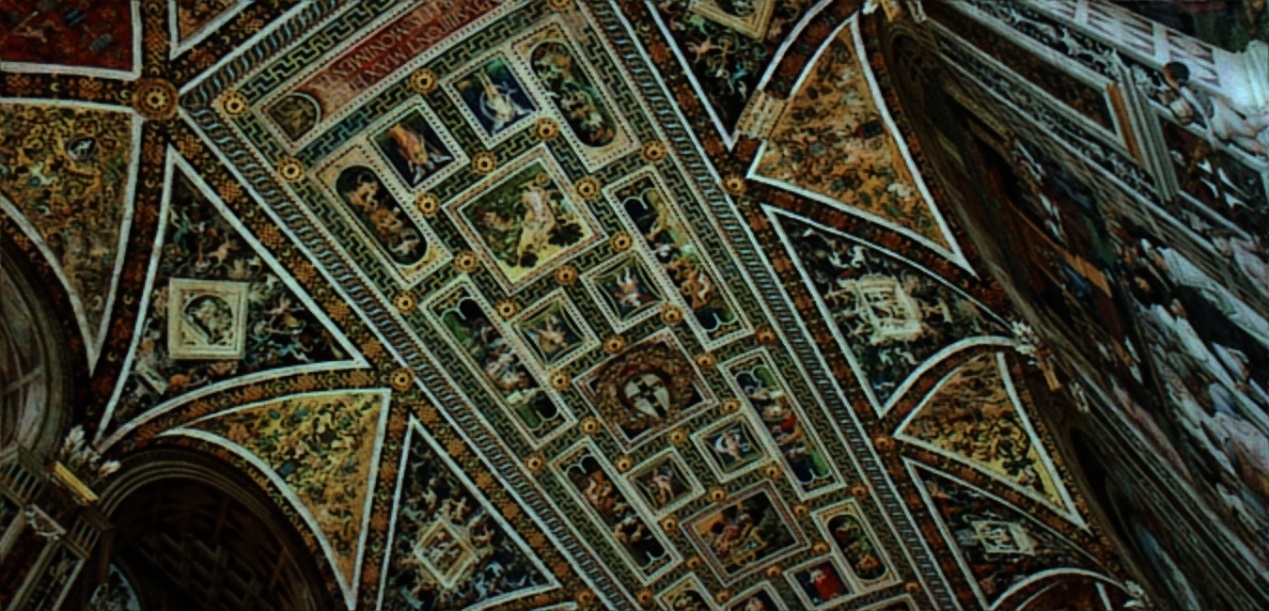}
\caption{\label{fig:track_2_yxy_floor1}
     EMVD}
\end{figure}

\begin{figure}[t]
\centering
\includegraphics[width=0.75\linewidth]{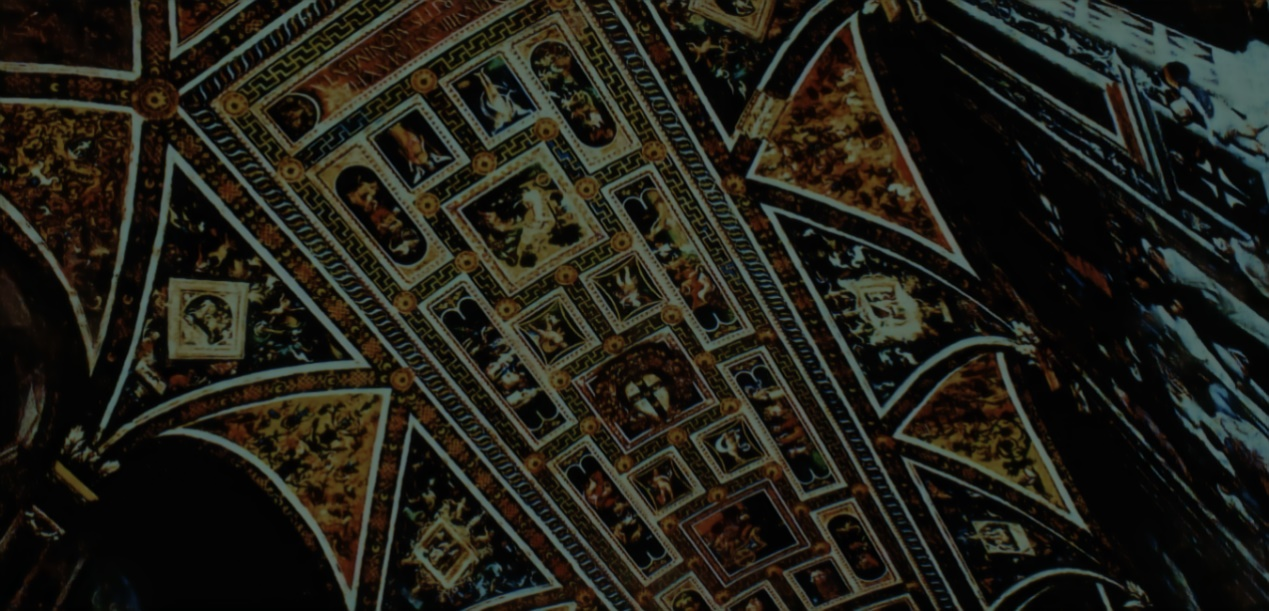}
\caption{\label{fig:track_2_yxy_floor2}
     Unnormalized MIRNetv2}
\end{figure}

\begin{figure}[t]
\centering
\includegraphics[width=0.75\linewidth]{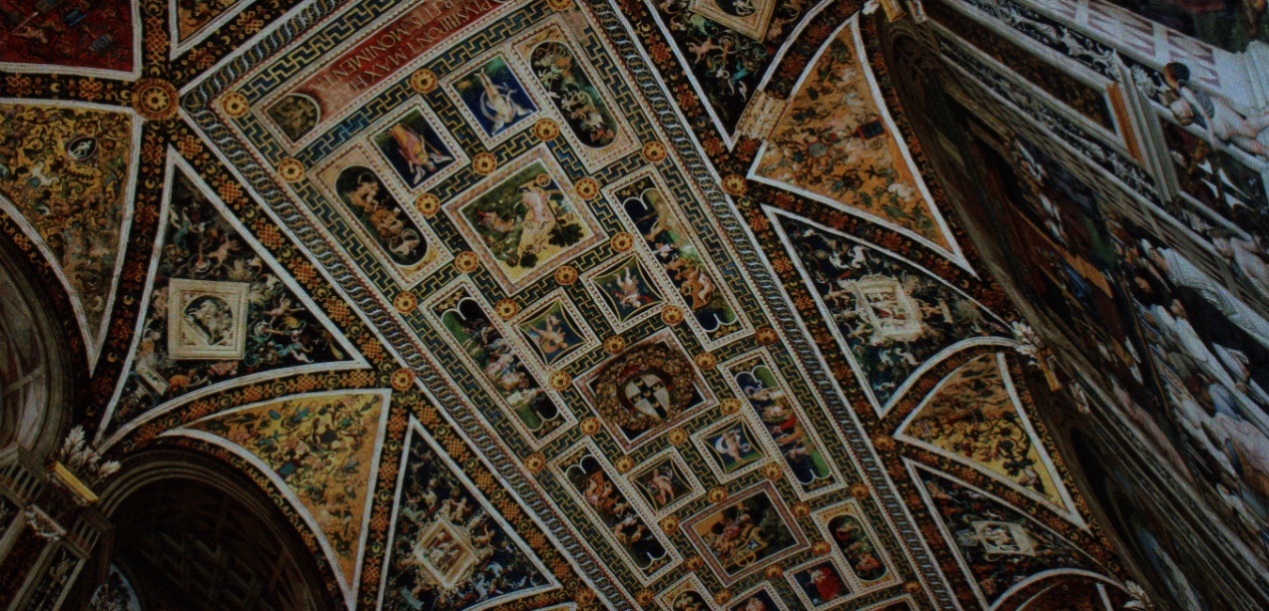}
\caption{\label{fig:track_2_yxy_floor3}
     Normalized MIRNetv2}
\end{figure}

\begin{figure}[t]
\centering
\includegraphics[width=0.75\linewidth]{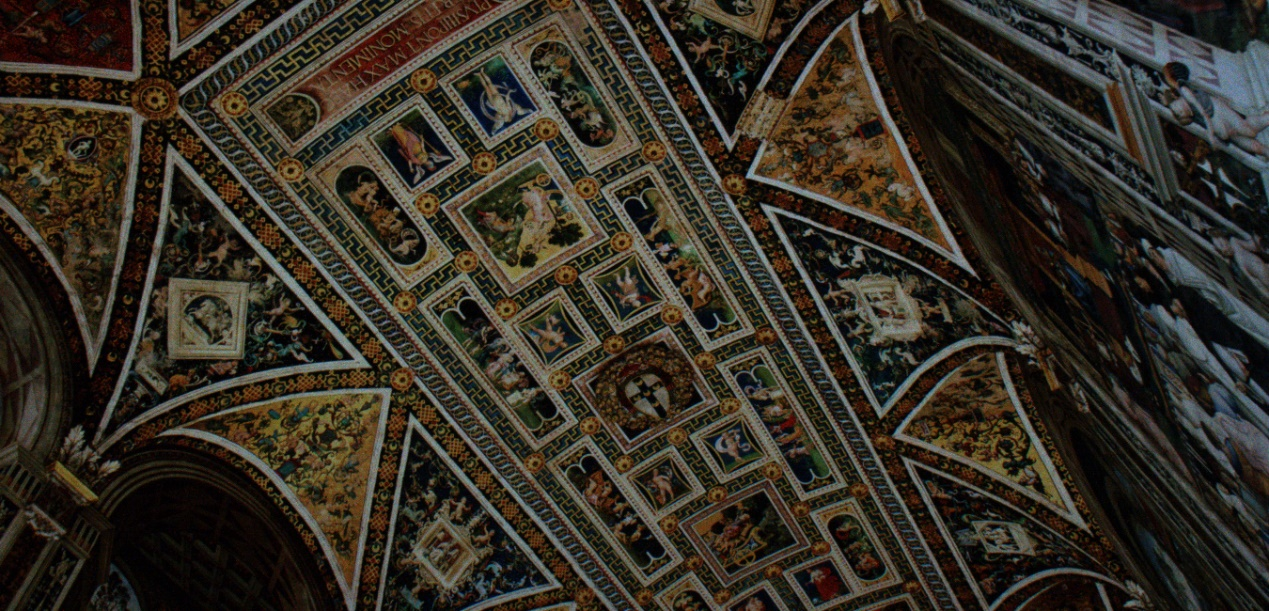}
\caption{\label{fig:track_2_yxy_floor4}
     Original images in the test set}
\end{figure}

  During the early stages of training, we also utilized the RViDeNet model, undergoing training in three phases: predenoising, pretraining, and finetuning. However, as training required four different noise levels while each scene in our training set only had three noise levels, our strategy was to directly duplicate the highest noise level from the training set. This led to poor generalization of our model during training, resulting in unsatisfactory performance during the validation phase. Therefore, we did not continue using this model during the testing phase. Nevertheless, we still believe that with a sufficient dataset and training time, this model has the potential to achieve better results.

\subsubsection{Teams and Affiliations}

\noindent \textbf{ZichunWang}

\noindent \textbf{Title:} Zichun Wang's Team Technique Report of CVPR 2024 PBDL Challenge Low-light raw video denoising with realistic motion

\noindent \textbf{Members:} Zichun Wang (\href{mailto:xiaoding310@gmail.com}{xiaoding310@gmail.com}), Ying Fu

\noindent \textbf{Affiliations:} Beijing Institute of Technology

\vspace{1em}
\noindent \textbf{wql}

\noindent \textbf{Title:} Low-light raw video denoising with realistic motion

\noindent \textbf{Members:}
Qinliang Wang (\href{mailto:23171214446@stu.xidian.edu.cn}{23171214446@stu.xidian.\linebreak edu.cn}), Xuejian Gou, Yang Liu, Lingling Li, Fang Liu, Wenping Ma

\noindent \textbf{Affiliations:} School of Artificial Intelligence, Xidian University

\vspace{1em}

\noindent \textbf{mmmmmm}

\noindent \textbf{Title:} Low-light raw video denoising with realistic motion

\noindent \textbf{Members:} Xinyue Yu (\href{mailto:}{23171214697@stu.xidian.edu.cn}), Sen Jia, Junpei Zhang, Licheng Jiao, Xu Liu, Puhua Chen

\noindent \textbf{Affiliations:} Intelligent Perception and Image Understanding Lab, Xidian University


\section{Low-light SRGB Image Enhancement}
\label{sec:intro}
\noindent
Compared with normal-light images, quality 
degradation of low-light images captured under terrible 
lighting conditions is serious due to inevitable environmental or 
technical constraints, leading to unpleasant visual perception including 
details degradation, color distortion, and severe noise. These phenomena have 
a significant impact on the performance of advanced downstream visual tasks, such 
as image classification, object detection, semantic segmentation \cite{ding2021dual,perez2020object,ren2015faster,yuan2019multi,zhang2023learning}, \etc. To mitigate 
the degradation of image quality, low-light image enhancement~\cite{lv2021attention} has become an important topic in the 
low-level image processing community to effectively improve visual quality and restore image details.
\\
\indent
To address this challenge, the CVPR 2024 PBDL 
Low-light sRGB Image Enhancement Challenge aims to evaluate 
and improve the visual quality of image enhancement algorithms 
in the field of low-light image enhancement.
\\
\indent
In the low-light sRGB image enhancement track (Table \ref{tab:track3_tab1}), 
the top three teams demonstrated exceptional performance. 
The IMAGCX team secured the first place, achieving a PSNR 
score of 22.70 and an SSIM score of 0.82. The chm team 
came in second, with a PSNR score of 22.62 and an SSIM 
score of 0.82. The WanFly team achieved the third place
 with a PSNR score of 21.82 and an SSIM score of 0.81. 
 Their enhanced images achieved excellent visual quality,
  showcasing their strong performance in this challenging task.
\\
\indent
These results highlight the significant progress made 
by the participating teams in addressing the challenges 
of low-light sRGB image enhancement. The top-ranking teams 
demonstrated their expertise and innovative capabilities in 
developing image enhancement algorithms that excel in low-light 
conditions, paving the way for future advancements in computer 
vision research.
\begin{table}[t]
	\centering
	\setlength{\tabcolsep}{12pt}
	\caption{Leaderboard of the low-light SRGB Image Enhancement}
	\label{tab:track3_tab1}
	\begin{threeparttable}
		\begin{tabular}{cccc}
			\toprule
			\textbf{Rank} & \textbf{Team} & \textbf{PSNR} & \textbf{SSIM} \\ \hline
			1 & IMAGCX & 22.70 & 0.82 \\ 
			2 & chm & 22.62 & 0.82 \\ 
			3 & WanFly & 21.82 & 0.81 \\ \bottomrule
		\end{tabular}
	\end{threeparttable}
\end{table}
\subsection{Low-Light SRGB Image Enhacement Dataset}
\noindent
To systematically investigate the effectiveness of 
the proposed method in real-world conditions, 
a real low-light image dataset for 
image enhancement is necessary 
and fundamental. The challenge utilizes 
the Paired Normal/Low-light Images (PNLI) 
dataset, introduced by \cite{fu2022gan}. 
\\
\indent
It is collected 
using a Canon EOS 5D Mark IV camera. Fig. \ref{fig:track3_Dataset} 
shows examples of paired normal/low-light images from the PNLI 
dataset. The PNLI dataset exhibits the following 
characteristics:
\begin{figure*}[t!]
  \centering
  \includegraphics[width=0.95\linewidth]{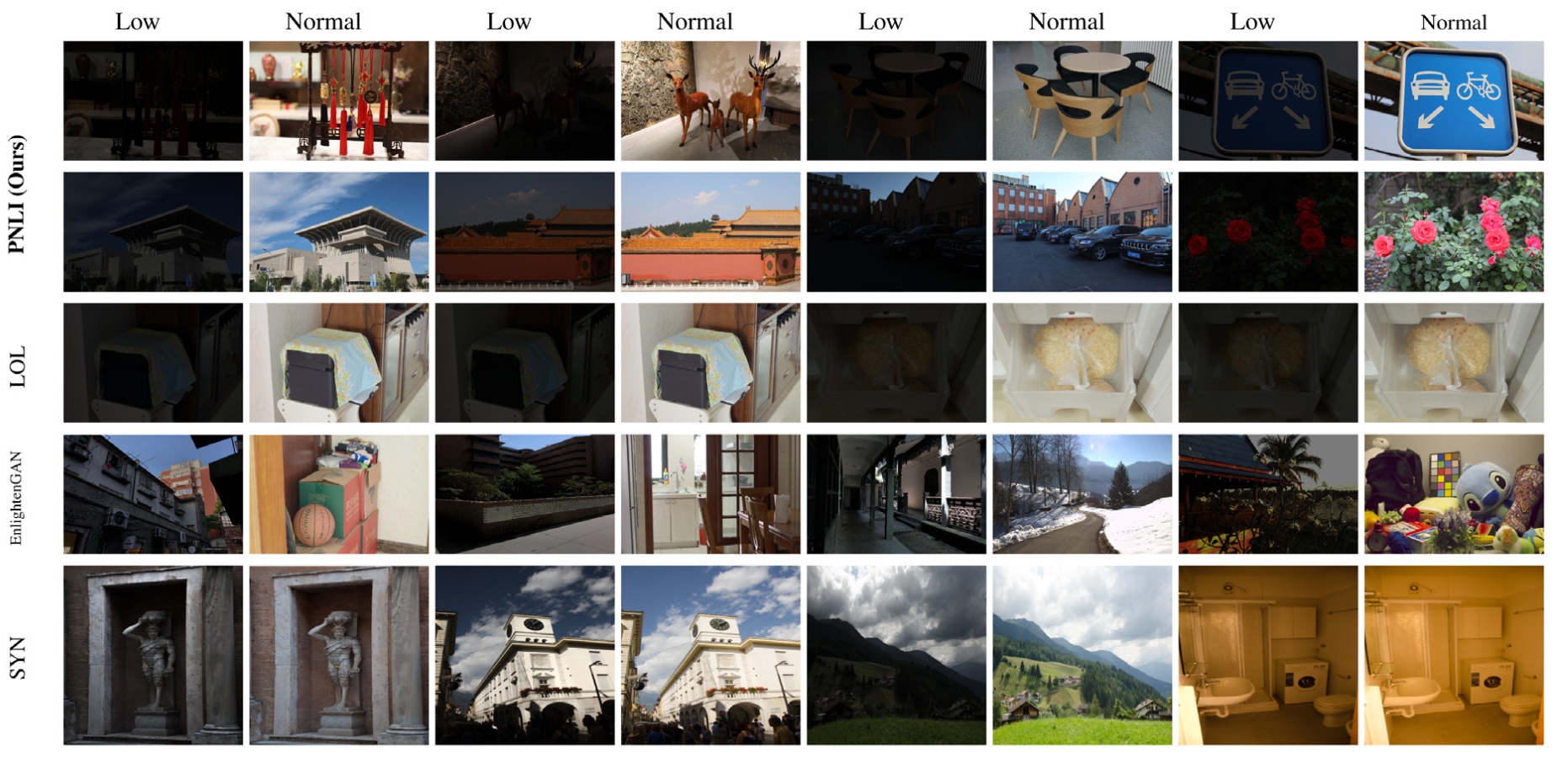}
   \caption{Several representative examples for low/normal-light images in PNLI dataset, LOL dataset, SYN dataset and EnlightenGAN dataset. Objects and scenes captured in our PNLI dataset are more diverse, abundant and superior.}
   \label{fig:track3_Dataset}
\end{figure*}
\\
\begin{itemize}
  \setlength{\itemsep}{0pt}
  \setlength{\parsep}{0pt}
  \setlength{\parskip}{0pt}
  \item It contains 2,000 image pairs, which is four times the size of the LOL dataset.
  \item Different from the existing real scenes dataset, i.e., LOL, there are no repeated scenes in our PNLI dataset, which is more abundant and superior than LOL. (There are many very similar scenes with little difference in the LOL dataset, as shown in Fig. \ref{fig:track3_Dataset})
  \item All images in PNLI are collected from considerably more real scenes, which contain both indoor and outdoor scenes. In addition, the object categories in images are rich and common.
  \item Excellent visual quality and clarity, which might help in learning pixel-level contextual information.
  \item The darkness levels of low-light images in PNLI are rich, and it can truly restore various situations where the actual image brightness is missing due to insufficient ambient light or human operation mistakes. Therefore, it can effectively verify the stability and robustness of our proposed method.
\end{itemize}

\subsection{IMAGCX Team's Method}

\subsubsection{Network Architecture}
\begin{figure*}[t!]
  \centering
  \includegraphics[width=0.95\linewidth]{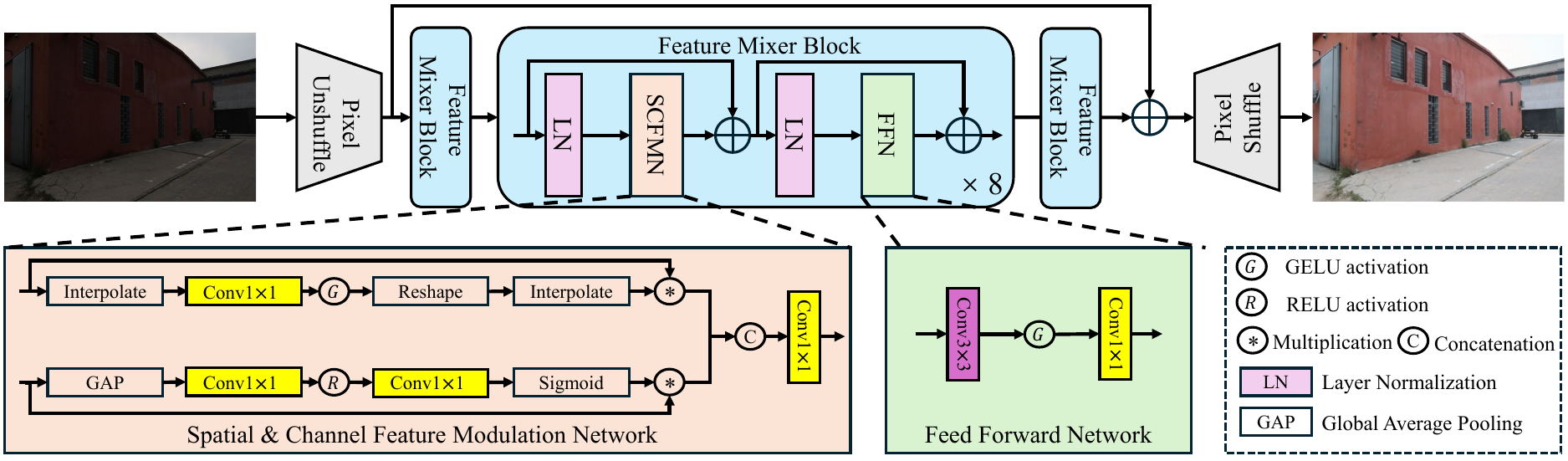}
   \caption{Several representative examples for low/normal-light images in PNLI dataset, LOL dataset, SYN dataset and EnlightenGAN dataset. Objects and scenes captured in our PNLI dataset are more diverse, abundant and superior.}
   \label{fig:track3_IMAGCX}
\end{figure*}

To solve UHD 
low-light image enhancement, several recent state-of-the-art 
methods have been proposed, for example LLFormer \cite{wang2023ultra}, UHDFour\cite{li2023embedding}, 
and MixNet \cite{wu2024mixnet}. We first conduct cross-domain generalization 
analysis on these methods, and we find that MixNet can better 
generalize to unseen real images. Thus, MixNet \cite{wu2024mixnet} is employed 
as the network backbone for low-light image enhancement.
\\
\indent
Fig.~\ref{fig:track3_IMAGCX} shows the overview of the network architecture. It aims to map an UHD low-light input image $x \in \mathbb{R}^{H \times W \times C}$ to its corresponding normal-clear version $y \in \mathbb{R}^{H \times W \times C}$, where $H$, $W$, and $C$ represent height, width, and channel, respectively. To reduce computational complexity, it downsample the input to 1/4 of the original resolution by PixelUnshuffle. Subsequently, the shallow features go through multiple deep feature mixer blocks. Each feature mixer block mainly consists of a feature modulation network and a feed forward network. To better capture long-range pixel dependencies in UHD images, feature modulation network combines spatial and channel dimensions for joint feature modeling. Finally, we use PixelShuffle upsampling to reconstruct the final image. 
\\

\subsubsection{Implementation Details}
We conduct model training in PyTorch framework on 8 NVIDIA GeForce RTX 4090 GPUs. Furthermore, we incorporate other public UHD low-light image enhancement datasets (UHD-LL~\cite{li2023embedding} and UHD-LOL~\cite{wang2023ultra}) into the network training. Similar to~\cite{liu2024ntire}, patches at the size of $2000 \times 2000$ are randomly cropped from the image pairs as training samples. The training data is augmented with random rotation and flipping. To optimize the network, we adopt L1 loss as the optimization objective, and we employ the Adam optimizer with a learning rate $2 \times 10^{-4}$. In total, we perform 600k iterations. During the testing phase, we perform full-resolution inference using one NVIDIA GeForce RTX 4090 GPU. Note that we employ a self-ensemble strategy to further improve performance.
The code and model are released at \href{https://drive.google.com/file/d/11Yn6Q4gCjLxWllwApaWRJW5K5DRyylF4/view?usp=sharing}{code}.

\subsection{chm Team's Method}

\subsubsection{Network Architecture}
\begin{figure}[!t]
  \centering
  \includegraphics[width=1.0\linewidth]{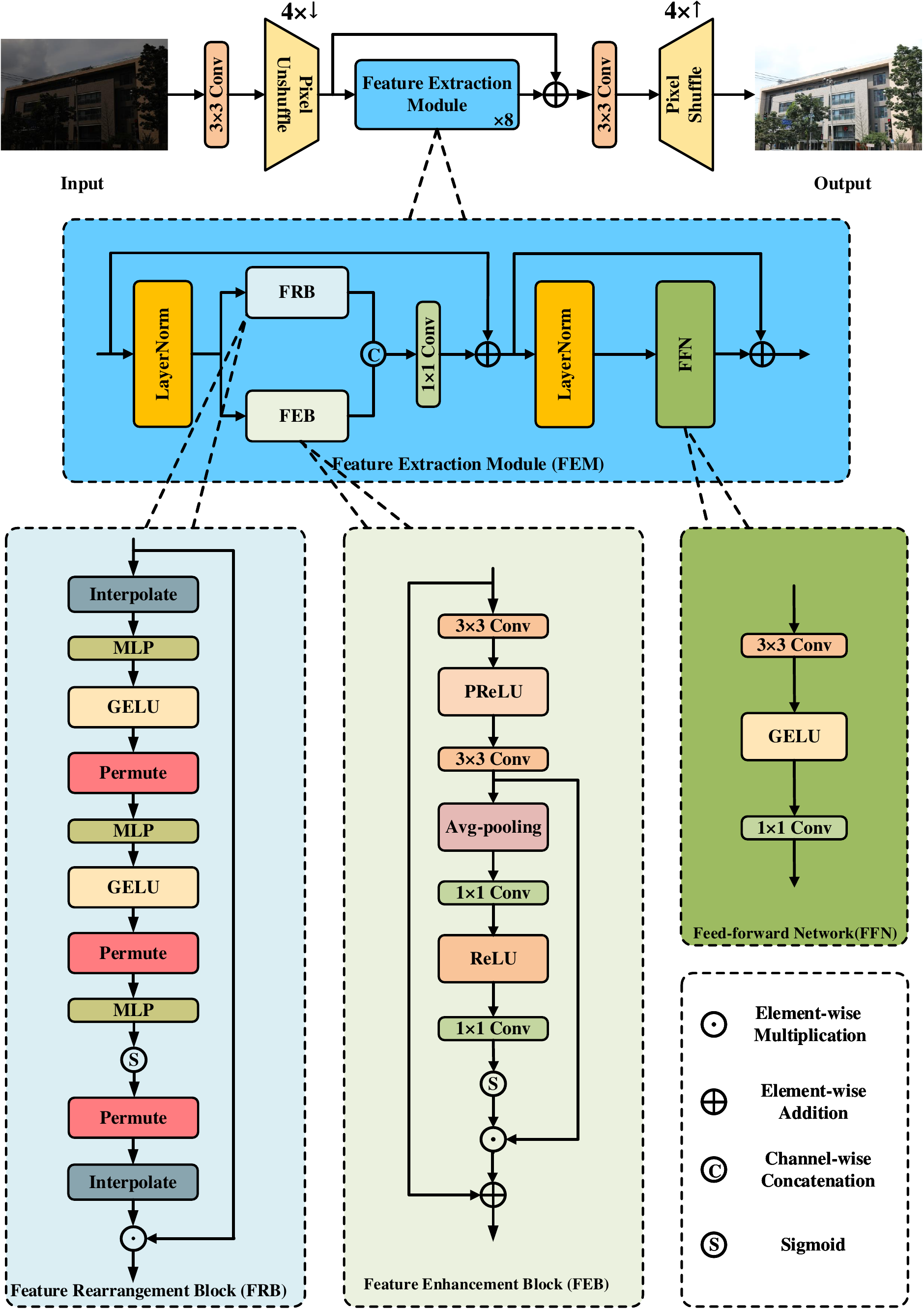}
  \caption{The network architecture of team SuperGo.}
  \label{fig:track3_chm}
  \end{figure}

Fig.~\ref{fig:track3_chm} illustrates the overall architecture of our method. Specifically, the input $x$ is
first reshaped to feature tensor via 
PixelUnshuffle ($4 \times \downarrow$) to preserve 
original information, and then fed to 8 feature 
extraction modules. Finally, the output feature $y$ 
is reshaped to the original height and width of input $x$ 
via Pixelshuffle ($4 \times \uparrow$). The feature extraction 
module mainly contains a feature rearrangement block (FRB), 
a feature enhancement block (FEB), and a feed-forward network 
(FFN). Here, FRB adopts MLP-based tensor dimensional 
transformations \cite{zhuang2023dimensional}, while FEB employs CNN-based local operators \cite{zamir2021multi}. The overall process can be represented as follows:
\begin{equation}
\begin{split}
&F_{1}=\operatorname{Conv}\left[\operatorname{FRB}\left(\operatorname{LN}\left(F_{0}\right)\right) ; \operatorname{FEB}\left(\operatorname{LN}\left(F_{0}\right)\right)\right]+F_{0}, \\
&F_{2}=\operatorname{FFN}\left(\operatorname{LN}\left(F_{1}\right)\right)+F_{1},
\end{split}
\end{equation}
where $F_{0}$ denote the input features, $F_{1}$ denote the intermediate features and $F_{2}$ denote the output features. LN refers to the layer normalization.

\subsubsection{Implementation Details}
To supervise the training process, we employ the L1 loss as the objective function. We conduct model training on 4 NVIDIA TESLA V100s with 32GB memory. In total, we perform 500 epochs of training. During the training, we adopt the Adam optimizer with a learning rate of $2 \times 10^{-4}$. The patch size is set to be $768 \times 768$ pixels and the batch size is set to be 16. To augment the training data, we apply random horizontal and vertical flips. For testing images, we use one NVIDIA GeForce RTX 4090 GPU with 24GB memory. The source code and pre-trained model are available at \href{https://drive.google.com/file/d/1tPxeQBzI_ELlAmmoA70j2Xh98MxFFV4H/view?usp=sharing}{model}.

\subsection{WanFly Team's Method}
\begin{figure*}[t!]
  \centering
  \includegraphics[width=0.95\linewidth]{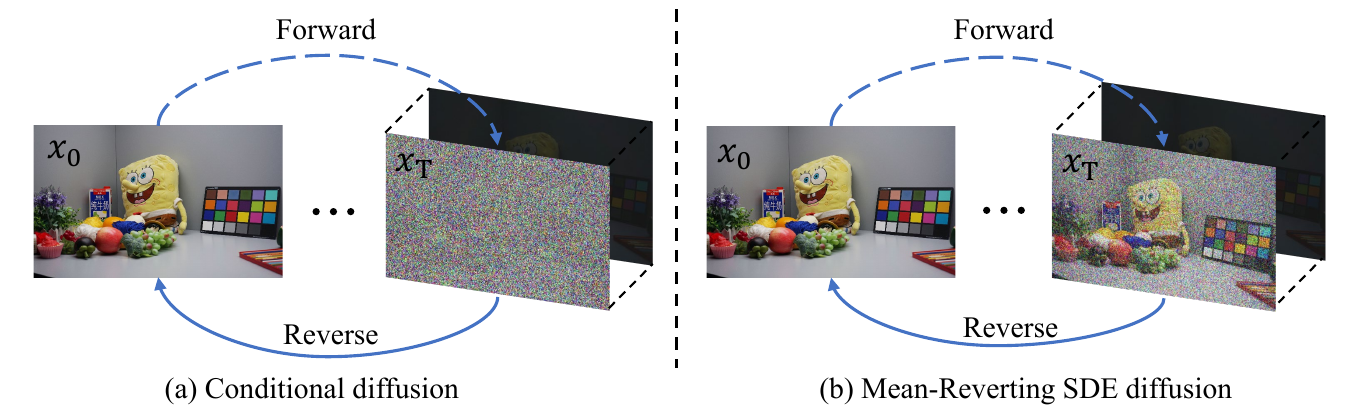}
   \caption{(a) Conditional diffusion; (b) Mean-Reverting SDE diffusion}
   \label{fig:track3_WanFly1}
\end{figure*}

Diffusion models are increasingly applied 
in low-light image enhancement tasks due 
to their exceptional capability to model 
data distributions, but an inherent drawback 
of diffusion models in image restoration tasks 
is that starting the reverse process from pure 
Gaussian noise can lead to artifacts \cite{clediff,docdiff}. 
Therefore, as illustrated in Fig. \ref{fig:track3_WanFly1}, we adopt the 
Mean-Reverting Stochastic Differential Equation (SDE) \cite{irsde} 
as the base diffusion framework, directly implementing the mapping 
from low-quality to high quality images.
\\
\indent
The fundamental idea of diffusion models is to gradually corrupt images by injecting noise, and then learn how to progressively remove this noise to reconstruct the original image. U-Net plays a crucial role in this denoising process. It is trained to predict the noise injected at each step, thereby methodically eliminating the noise and restoring the image. The U-Net used in diffusion models typically consists of residual blocks, upsampling and downsampling operations, and attention mechanisms. While the stacking of multiple residual blocks is beneficial for feature extraction, it increases the computational load, and the extensive convolutional operations are not friendly to low pixel values in low-light images.
\\
\indent
Our motivation is to reduce multiplication operations in U-Net, protect low pixel values, and lighten the computational load. The simplified U-Net designed in this paper, as illustrated in Fig.~\ref{fig:track3_WanFly1}(a), is only constructed from the feature extraction module SimpleGate \cite{simple} and Parameter-free attention \cite{pfam} (SimPF) block, and includes upsampling and downsampling operations, making it suitable for both processing low-light images and reducing the resource consumption of the diffusion model for faster sampling.
\\
\indent
The code is available at \url{https://github.com/MrWan001/SFDiff}.
\subsubsection{Network Architecture}
\begin{figure*}[t!]
  \centering
  \includegraphics[width=0.95\linewidth]{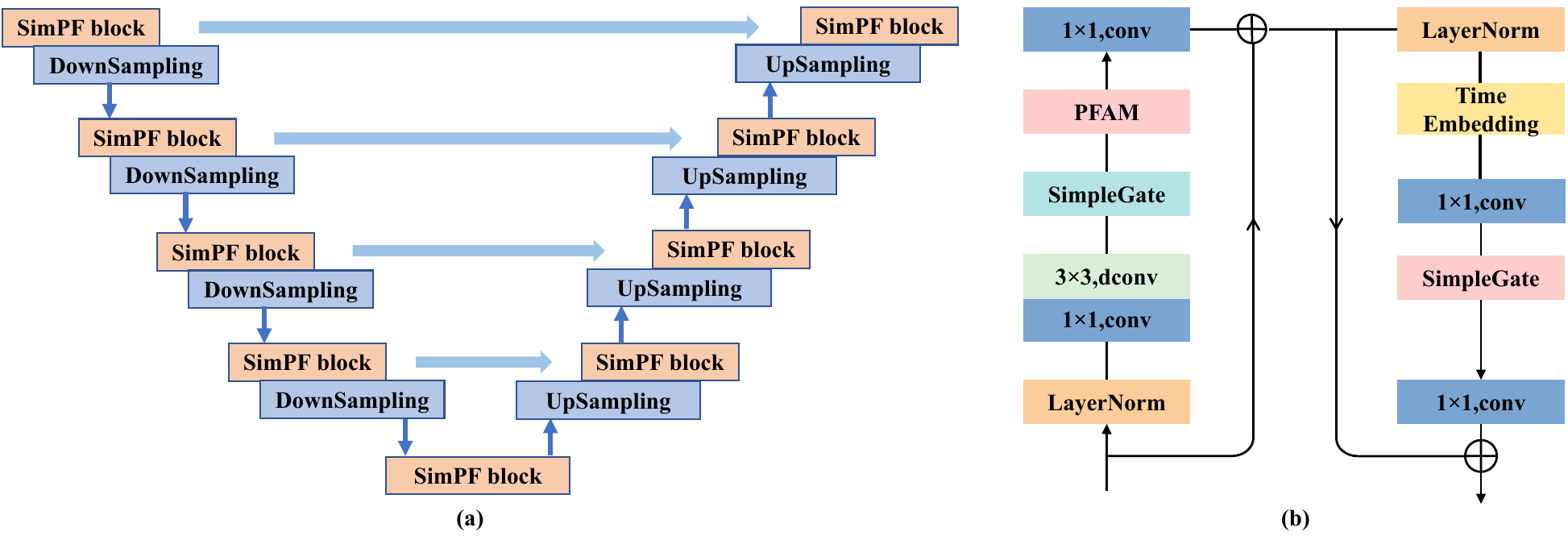}
   \caption{(a) U-Net with SimPF block. It is composed of SimPF blocks, upsampling and downsampling operations, along with skip connections; (b) SimPF block. It retains necessary convolution and normalization layers, incorporates SimpleGate and PFAM to minimize multiplication operations, and utilizes Time
   Embedding to align with diffusion models.}
   \label{fig:track3_WanFly2}
\end{figure*}

As shown in  Fig.~\ref{fig:track3_WanFly2}(b), we designed the SimPF block with the idea of retaining the necessary convolution and normalisation layers and using less computationally intensive components to reduce multiplication operations across feature maps. We use $1\times1$ convolutions and $3\times3$ depth-wise separable convolutions for feature extraction, both convolution types have been applied and proven effective in a variety of image restoration tasks. Specifically, the feature map first undergoes a $1\times1$ convolution to expand the number of channels while preserving spatial information. Subsequently, a $3\times3$ depth-wise separable convolution is employed to encode features from spatially adjacent pixel positions, facilitating the learning of local image structures.
\\
\indent
Since the activation function requires multiple multiplication operations, we use SimpleGate to replace complex nonlinear activation functions. SimpleGate can achieve the effect of nonlinear mapping through a single multiplication operation, which is particularly beneficial for preserving information in low pixel values, as complex functions like the cubic operations required in the GELU activation function can be detrimental to such information. The computation of SimpleGate is illustrated in Equation (\ref{eq8}):
\begin{equation}\label{eq8}
	\operatorname{SimpleGate}(\boldsymbol{X}, \boldsymbol{Y})=\boldsymbol{X} \odot \boldsymbol{Y}
\end{equation}
\indent
$\boldsymbol{X}$ and $\boldsymbol{Y}$ represent the division of a feature map with channels $C$, height $H$, and width $W$ along the channel dimension into two parts of ($\frac{C}{2}$, $H$, $W$). The essence of this multiplication operation is a type of nonlinear mapping that can substitute for an activation function.
\\
\indent
After the feature matrix has been given weights through Parameter-Free Attention Mechanism (PFAM), a $1\times1$ convolution is used to aggregate pixel-level cross-channel context information. The subsequent two $1\times1$ convolutions serve to facilitate interaction and combination among features across different channels, creating more complex and effective feature representations. In order to apply to the diffusion model, we have incorporated a time embedding block, which takes the current diffusion time step $t$ as input and encodes $t$ into the feature matrix, enabling the model to perceive noise at different time steps $t$. Overall, the design of SimPF block, while minimizing multiplication operations, maintains robust feature extraction capabilities. \vspace{-1em}
\subsubsection{Implementation Details}
Our method is implemented using the PyTorch framework. The diffusion time step $T$ is established at $100$. A cosine scheduling scheme is utilized for noise scheduling. The optimization is carried out using the LION optimizer. The batch size is set to $6$. The initial learning rate is set at $4 \times 10^{-5}$, and the Cosine Annealing strategy is employed for learning rate scheduling. The model is trained on a single NVIDIA GeForce RTX 3090 GPU and converged after 300,000 iterations.
\\
\indent
During the training phase, we first attempted to crop or randomly crop the center of the training set to 256 x 256, but did not achieve good results. Finally, we resized the training set to 256 x 256 and achieved good results. During testing, due to the large size of 6720 x 4480, which exceeded the maximum range that the model could handle, we first attempted to crop the image into 2240 x 2240 and merge it, but the effect was not good. Finally, we resized the image to 480 x 320, and after using model enhancement, we resized it to 6720 x 4480, achieving good results. In addition, due to the unknown GT, we used the lpips metric to preliminarily evaluate the enhancement results of the model. We found that our proposed SimPF block performed better on the three images 1162, 496, 735, while the model trained on the original NAF block performed better on the other test images. Therefore, we combined the results of the two models to obtain the final version for submission.

\subsection{Teams and Affiliations}
\noindent\textbf{IMAGCX}

\noindent\textbf{Title:} PBDL-Challenge-IMAGCX for the Low-Light-srgb-enhancement track Technical Report

\noindent\textbf{Members:} 
Xiang Chen (\href{mailto:chenxiang@njust.edu.cn}{chenxiang@njust.edu.cn}), 
Hao Li, 
Jinshan Pan

\noindent\textbf{Affiliations:} 
Nanjing University of Science and Technology
\vspace{1em}

\noindent\textbf{chm}

\noindent\textbf{Title:} PBDL Challenge on Low Light SRGB Image Enhancement

\noindent\textbf{Members:} 
Chuanlong Xie\textsuperscript{1} (\href{mailto:jiechuanlong@stu.sau.edu.cn}{jiechuanlong@stu.sau.\linebreak edu.cn}), 
Hongming Chen\textsuperscript{1}, 
Mingrui Li\textsuperscript{2}, 
Tianchen Deng\textsuperscript{3}, 
Jingwei Huang\textsuperscript{4}, 
Yufeng Li\textsuperscript{1}

\noindent\textbf{Affiliations:} 
\textsuperscript{1}Shenyang Aerospace University, 
\textsuperscript{2}Dalian University of Technology, 
\textsuperscript{3}Shanghai Jiao Tong University, 
\textsuperscript{4}University of Electronic Science and Technology of China

\vspace{1em}

\noindent\textbf{WanFly}

\noindent\textbf{Title:} WanFly for the Low-Light-srgb-enhancement track

\noindent\textbf{Members:} 
Fei Wan\textsuperscript{1,2} (\href{mailto:20221083510904@buu.edu.cn}{20221083510904@buu.edu.cn}), 
Bingxin Xu\textsuperscript{1,2*}, 
Jian Cheng\textsuperscript{1,2}, 
Hongzhe Liu\textsuperscript{1,2}, 
Cheng Xu\textsuperscript{1,2}, 
Yuxiang Zou\textsuperscript{1,2}, 
Weiguo Pan\textsuperscript{1,2}, 
Songyin Dai\textsuperscript{1,2}

\noindent\textbf{Affiliations:} 
\textsuperscript{1*}Beijing Key Laboratory of Information Service Engineering, Beijing Union University, Beijing, 100101, China.
\textsuperscript{2}College of Robotics, Beijing Union University


\section{Extremely Low-light Image Denoising}
Light is crucial for photography. Nighttime and low-light conditions impose significant challenges due to the limited number of photons and unavoidable noise. The typical response is to increase light capture by, for example, enlarging the aperture, lengthening the exposure time, or using a flash. However, each approach has its drawbacks: a larger aperture results in a shallow depth of field and is not feasible for smartphone cameras; extended exposure times can lead to blurriness from scene changes or camera movement; and flash can cause color distortions and is effective only for objects close to the camera.

A practical solution for low-light imaging is burst photography~\cite{mildenhall2018burst,hasinoff2016burst,liu2014fast,liba2019handheld}, which aligns and fuses multiple images to increase the signal-to-noise ratio (SNR). However, burst photography is prone to ghosting effects~\cite{hasinoff2016burst,shen2019human} when capturing dynamic scenes involving vehicles, people, etc. An emerging alternative is using neural networks to automatically learn the mapping from a low-light noisy image to its long-exposure counterpart~\cite{chen2018learning}. This deep learning approach typically requires a large amount of labeled training data resembling real-world low-light photographs. Collecting extensive high-quality training samples from various modern camera devices is extremely labor-intensive and expensive.

To bridge the domain gap between synthetic images and real photos, some works have collected paired real data for both evaluation and training~\cite{abdelhamed2018high,chen2018learning,schwartz2018deepisp,chen2019seeing,jiang2019learning}. Despite promising results, gathering sufficient real data with true labels to prevent overfitting is very costly and time-consuming. Recent works use paired~\cite{lehtinen2018noise2noise} or single noisy images~\cite{krull2019noise2void,zou2023iterative} as training data instead of paired noisy and clean images. However, they do not significantly reduce the labor required to capture a large volume of real-world training data.

Another research direction focuses on enhancing the realism of synthetic training data to avoid the challenges of obtaining real data from cameras. By considering photon arrival statistics ("shot" noise) and sensor readout effects ("read" noise), works like~\cite{mildenhall2018burst,brooks2019unprocessing} use a signal-dependent heteroscedastic Gaussian model~\cite{foi2008practical} to characterize noise in raw sensor data. Recently, Wang \etal~\cite{wang2019enhancing} proposed a noise model that accounts for dynamic stripe noise, color channel heterogeneity, and clipping effects to simulate high-sensitivity noise in real low-light color images. Additionally, a flow-based generative model called NoiseFlow~\cite{abdelhamed2019noise} was proposed to describe the distribution of real noise using latent variables with a density of one.
However, these methods often oversimplify the imaging pipeline of modern sensors, especially the noise sources introduced by camera electronics, which have been extensively studied in the electronic imaging community.~\cite{konnik2014high,healey1994radiometric,gow2007comprehensive,baer2006model,el2005cmos,farrell2008sensor,irie2008model,irie2008technique,boie1992analysis,wach2004noise,costantini2004virtual}

Therefore, we are honored to collaborate with the CVPR 2024 Workshop to launch the Extremely Low-Light Image Denoising Challenge. The primary objective of this challenge is to use deep learning methods for denoising real extremely low-light images, optimizing denoising performance and model robustness. Participants are tasked with enhancing model robustness and denoising effectiveness by modeling noise in real imaging processes and using synthetic datasets for training. This challenge aims to explore realistic low-light noise models and efficient denoising models for extremely low-light images. We aim to rigorously assess their effectiveness and identify key trends in network design. We welcome participants to push the boundaries of innovation and advance the technology of extremely low-light image denoising.

\subsection{Extreme Low-Light Image Denoising challenge}

\subsubsection{The dataset}
\begin{figure}[t]
    \centering
    \includegraphics[width=\linewidth]{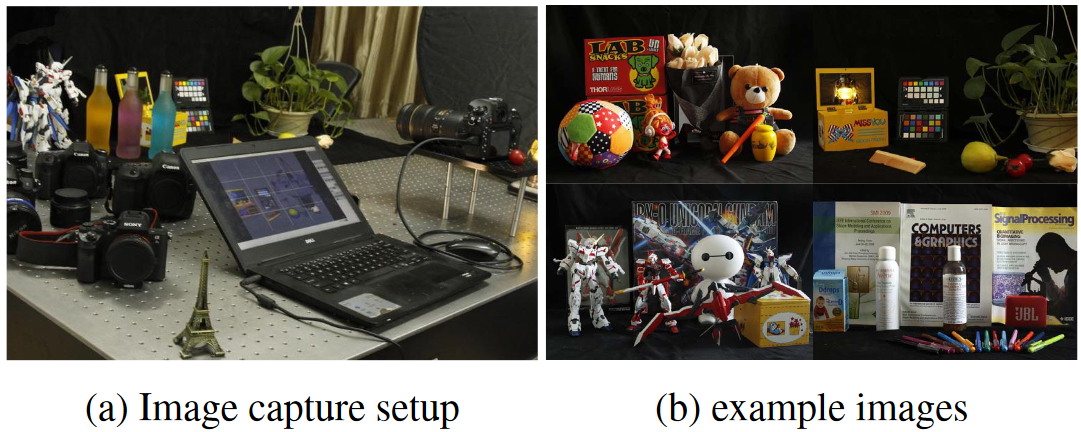}
    \caption{Capture setup and example images from our dataset.}
    \label{track4_fig:ELD_dataset}
\end{figure}

To systematically study the generality of the proposed noise formation model, we collect an extreme low-light denoising (ELD) dataset~\cite{wei2021physics} that covers 10 indoor scenes and 4 camera devices from multiple brands (SonyA7S2, NikonD850, CanonEOS70D, CanonEOS700D). We also record bias and flat field frames for each camera to calibrate our noise model. The data capture setup is shown in Fig.~\ref{track4_fig:ELD_dataset} For each scene and each camera, a reference image at the base ISO was firstly taken, followed by noisy images whose exposure time was deliberately decreased by low light factors f to simulate extreme low light conditions. Another reference image then was taken akin to the first one, to ensure no accidental error (e.g. drastic illumination change or accidental camera/scene motion) occurred. We choose three ISO levels (800, 1600, 3200) and two low light factors (100, 200) for noisy images to capture our dataset, resulting in 240 (3×2×10×4) raw image pairs in total. The hardest example in our dataset resembles the image captured at a “pseudo” ISO up to 640000 (3200×200).

\subsubsection{The evaluation protocols}
The quantitative evaluation metrics include the standard Peak Signal-to-Noise Ratio (PSNR) and Structural Similarity Index (SSIM), which are commonly used to assess image quality. To ensure fairness and accuracy in testing, we use the Codalab platform (https://codalab.lisn.upsaclay.fr/competitions/17787) for result evaluation. Due to the storage limitations of the Codalab platform, we have cropped the original raw images to a size of 1024×1024 and saved the cropped images along with relevant camera parameters in mat files provided to participants. During testing, participants are also required to save the denoised images in mat files for submission. The Codalab platform will then compute the evaluation metrics based on the ground truth. The final competition score is :
\begin{equation}
    Score = log_k(SSIM * k^{PSNR}) = PSNR + log_k(SSIM),
\end{equation}
where k=1.2.

\begin{table}[t]
	\centering
	\caption{Leaderboard of the Extremely Low-light Denoising}
	\label{track4_tab:rank}
	\begin{threeparttable}
		\begin{tabular}{cccccc}
			\toprule
			\textbf{Rank} & \textbf{User} & \textbf{Score} & \textbf{PSNR} & \textbf{SSIM} \\ \hline
			1 & jly724215288 & 43.80 & 43.89 & 0.99 \\ 
			2 & yuxiaoxi & 43.07 & 43.15 & 0.99 \\ \bottomrule
		\end{tabular}
	\end{threeparttable}
\end{table}

As shown in Tab.~\ref{track4_tab:rank}, in the extremely low-light detection track, jly724215288 achieved first place with a score of 43.80, a PSNR of 43.89, and an SSIM of 0.99, demonstrating their exceptional denoising capabilities for extremely low-light images. Yuxiaoxi secured second place with a score of 43.07, a PSNR of 43.15, and an SSIM of 0.99. Both teams performed excellently in low-light instance segmentation, further highlighting the significance of their contributions.

These results highlight the remarkable progress made by the participating teams in addressing the challenges of noise in extremely low-light images. The top-ranked teams showcased their expertise and innovation in developing robust algorithms adapted to low-light conditions, paving the way for future advancements in computer vision research.

\subsection{jly724215288 Team's Method}

\subsubsection{Network Architecture}
\vspace{+0.918mm}
\noindent \textbf{Datasets processing.} The authors note that the specialty of the Bayer pattern raw sensor lies in each individual pixel receiving only one spectral wavelength of light at a time. Given the limited amount of training data available, it becomes imperative to consider the spectral properties of every training pair. There are four channels in the sensor: R, Gr, B, and Gb. The Gr and Gb pixels have slightly different intensity responses even though they both capture the green color wavelength, due to imperfections in the Color Filter Array lens (usually compensated by the Image Signal Processing GbGr balance module). Additionally, the offset in the R and B channels impacts denoising performance.

The authors' method first employs accurate patch-based registration while the images are captured on a tripod.

\begin{figure}[t]
    \includegraphics[width=\linewidth]{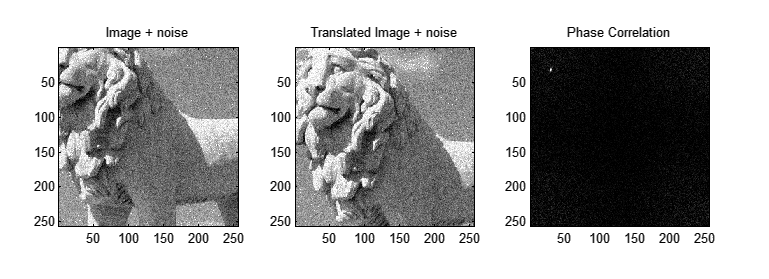}
    \caption{phase correlation. The white point corresponded phase position is the image offset.}
    \label{track4_fig:phase_correlation}
\end{figure}

The registration is performed using phase-correlation Fig.\ref{track4_fig:phase_correlation}, which is robust against strong noise and brightness changes. First, the images are transformed into the frequency domain using Fast Fourier Transform, then the cross-power spectrum is calculated by taking the complex conjugate multiplication with element-wise normalization.
$\mathbf{G}_a = \mathcal{F}\{I_a\}, \; \mathbf{G}_b = \mathcal{F}\{I_b\}$

$$
    R = \frac{ \mathbf{G}_a \circ \mathbf{G}_b^*}{|\mathbf{G}_a \circ \mathbf{G}_b^*|}
$$

Then they find the maximum response phase as the image patch offset $(\Delta x, \Delta y)$ in the inverse Fast-Fourier transform result.

$$
    r = \mathcal{F}^{-1}\{R\}
$$

$$
    (\Delta x, \Delta y) = \arg \max_{(x, y)}\{r\}
$$

The authors acknowledge that due to the settings and exposure time of different brands of sensors, as well as the sensitivity of the expected exposure value (EV) to the final result, they introduce a variable $\lambda$, ranging from 0.1 to 10, to reduce reliance on precise exposure accuracy. The digital gain (ISO) and exposure time in seconds are extracted from EXIF metadata and calculated into the exposure value (EV) for ratio estimation.

$$
    ratio = \lambda \frac{EV_{gt}}{EV_{in}} = \lambda \frac{ISO_{gt} \times TIME_{gt}}{ISO_{in} \times TIME_{in}}
$$

The authors carefully perform data augmentation through random size cropping and rotation while maintaining Bayer pixel alignment. They avoid any scale-like resampling augmentation to preserve the sensor noise properties.

\noindent \textbf{Network Architecture.} The authors use a two-stage training strategy for Bayer raw denoising, employing slightly different networks for each stage. In the first stage, they use a U-Net with residual blocks as the denoising network, utilizing the L1 loss function for faster convergence. In the second stage, they add attention blocks after the residual blocks and freeze the weights of the first-stage network, enhancing the denoising capability by minimizing the mean square error (MSE).

\begin{figure}[h]
    \begin{center}
        \includegraphics[width=\linewidth]{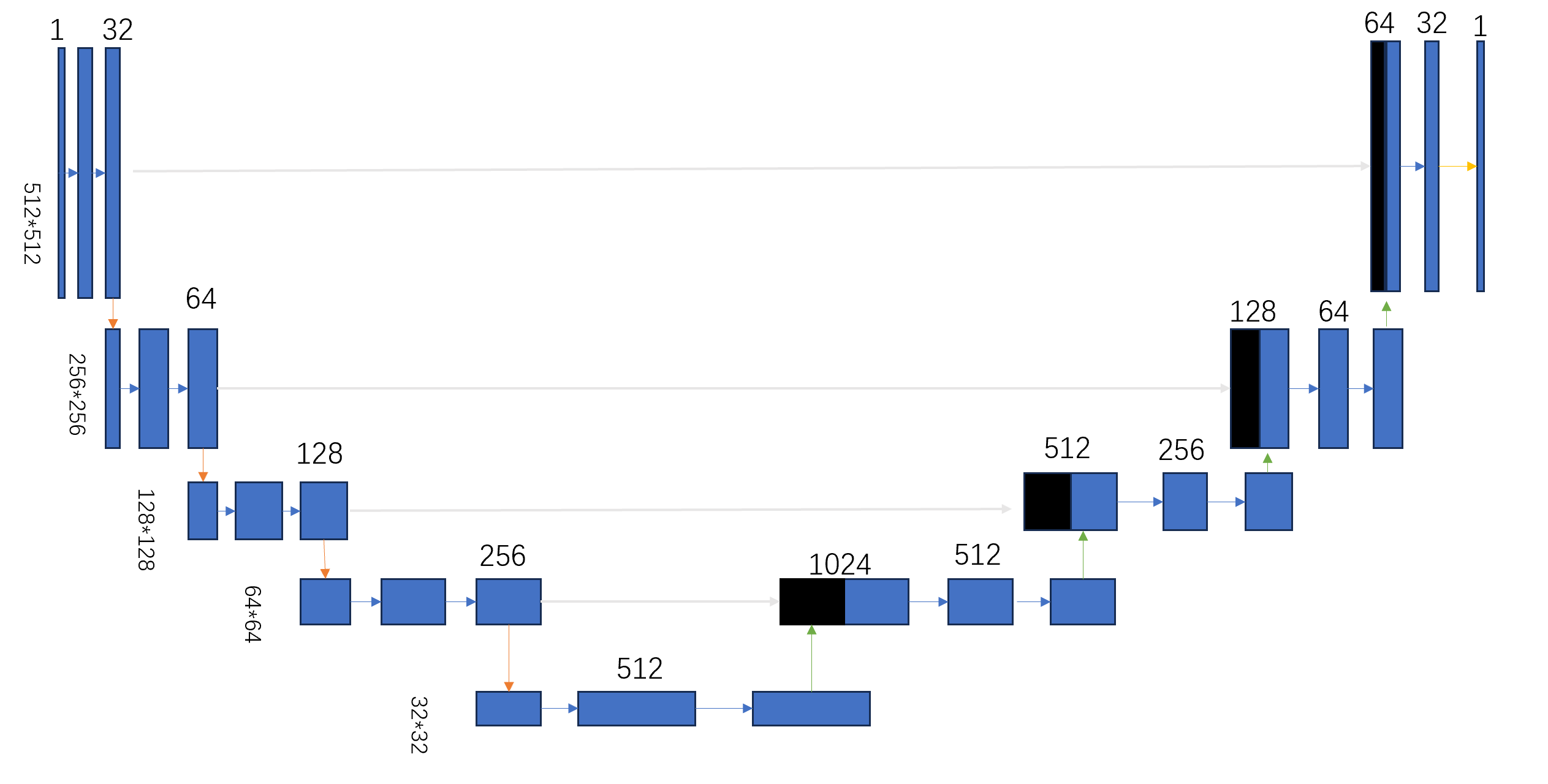}
        \caption{U-Net architecture~\cite{ronneberger2015u}.}
        \label{track4_fig:unet}
    \end{center}
\end{figure}

\subsubsection{Implementation Details}
The authors first train their model on raw files from various camera brands. Each training image patch pair undergoes extensive augmentation, including cropping, phase alignment with the ground truth, rotation, and scaling by the $\lambda$ ratio.

Following the initial pretraining, the authors fine-tune the model using the specific camera brand intended for testing to improve noise model estimation.

In practice, different sensors exhibit unequal black levels and white points per channel. Another contribution by the authors is the development of a method to normalize raw Bayer data values $I_n$ across images taken by different camera brands and exposure settings. When an image is captured, the raw file also records the black levels $L_b$ and white points per channel $L_w$. After multiplying by the EV ratio (which is approximately 10 times larger than the camera EV change) to achieve normal brightness, the noisy input might be clipped by the sensor's maximum bit value (usually 14 bits), leading to signal loss. The authors address this problem by normalizing using a large denominator $L_m$ (above 16 bits) after subtracting the black levels. Replacing $L_w$ with $L_m$ helps preserve image detail. Additionally, the variable $\lambda$ in training results in better denoised images while maintaining GPU-capable precision.

$$
I_n = \frac{I - L_b}{L_m - L_b} 
$$

The following equation shows a normalized Bayer raw multiply exposure ratio, then $f(.)$ donates our network function, finally de-normalize to the expect denoised bright image $I_{nr}$.

$$
I_{nr} = f(I_n * ratio) (L_m - L_b) + L_b
$$

\begin{figure*}[t]
    \centering
    \includegraphics[width=\textwidth]{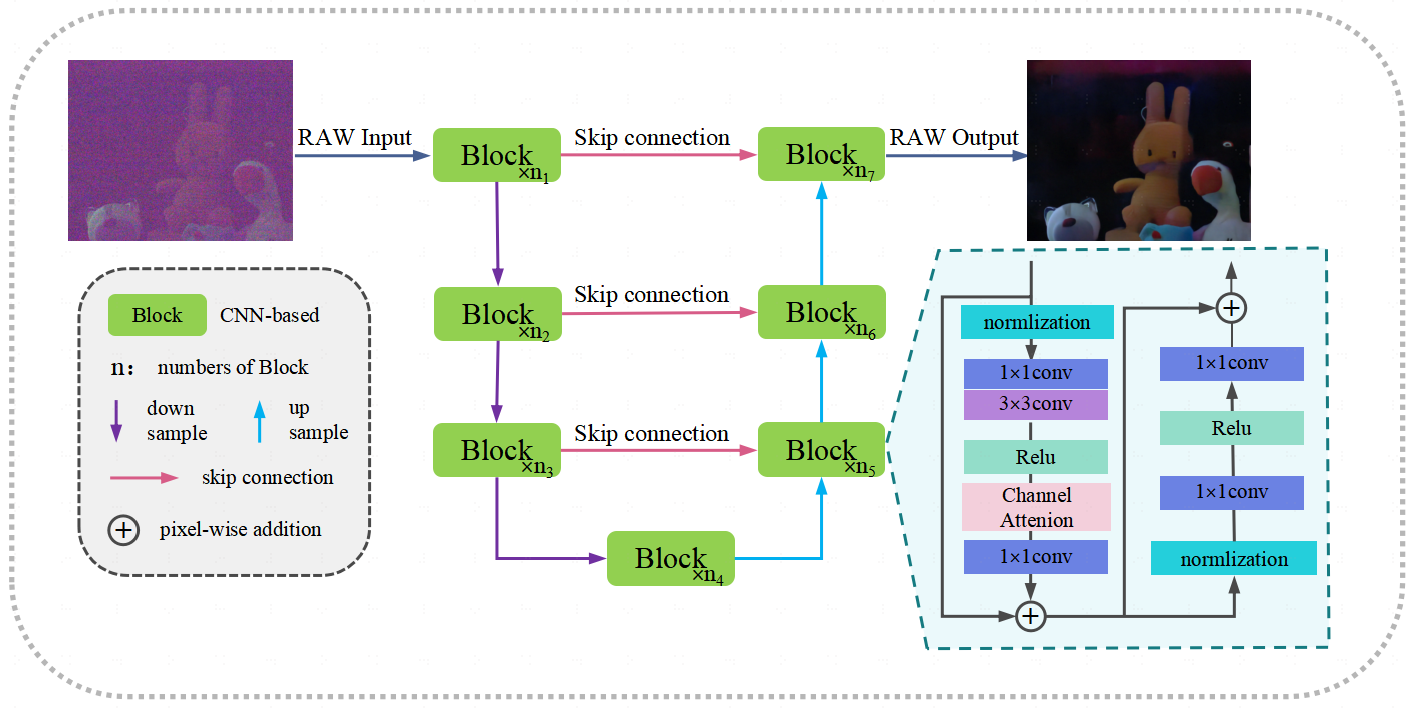}
    \caption{The overall network architecture of method}
    \label{track4_fig:yuxiaoxi_network}
\end{figure*}

\subsection{yuxiaoxi Team's Method}

\subsubsection{Network Architecture}
The authors adopt the classic single-stage U-shaped architecture with skip-connections, as shown in Fig.~\ref{track4_fig:yuxiaoxi_network}, to reduce inter-block complexity. The neural networks are constructed by stacking blocks. They start with a plain block containing the most common components: convolution, ReLU, and shortcut. Additionally, they find that vanilla channel attention meets the requirements for computational efficiency and brings global information to the feature map. Given the proven effectiveness of channel attention in image restoration tasks, the authors incorporate channel attention into the plain block.

The authors use convolution with a kernel size of 2 and a stride of 2 for the downsample layer. For the upsample layer, they double the channel width using a pointwise convolution followed by a pixel shuffle module. There are skip connections from the encoder block to the decoder block, and the authors simply add the encoder and decoder features element-wise for feature fusion. The default width and number of blocks are 64 and 36, respectively, and their network architecture consists of a total of 5 layers. The encoder has 4 layers with block quantities of 2, 2, 4, and 8, respectively. The intermediate connection layer has 12 blocks. The decoder also has 4 layers, with each layer containing 2 blocks.

\subsubsection{Training strategy}
The authors utilize the $L_1$ loss function as the training objective, similar to most denoising methods. They employ the same data preprocessing and optimization strategy as ELD during pre-training. The raw images with long exposure times in the SID train subset are used for noise synthesis. For data preprocessing, they pack the Bayer images into 4 channels, then crop the long exposure data into patches of size 512×512 with a non-overlapping step of 256. The models are trained for 300 epochs using the Adam optimizer with $\beta_1$ = 0.9 and $\beta_2$ = 0.999, without applying weight decay. The initial learning rate is set to $10^{-4}$, halved at the 150th epoch, and further reduced to $10^{-5}$ at the 220th epoch.

The inference code and the pre-trained models are released at \href{https://pan.baidu.com/s/1P0c91nulXhHGQZXBFDiDJQ?pwd=e5xc}{here}.

\subsection{Teams and Affiliations}

\noindent \textbf{jly724215288 }

\noindent \textbf{Title:} Technique Report of Team jly724215288 for CVPR
2024 PBDL Challenge Extremly Low-light Image Denoising

\noindent \textbf{Members:} Linyan Jiang (\href{mailto:724215288@qq.com}{724215288@qq.com}), Bingyi Song, Zhuoyu An, Haibo Lei, Qing Luo, Jie Song

\noindent \textbf{Affiliations:} Tencent \\

\noindent \textbf{yuxiaoxi}

\noindent \textbf{Title:} Technique Report of Team yuxiaoxi for CVPR
2024 PBDL Challenge Extremly Low-light Image Denoising

\noindent \textbf{Members:} Yuan Liu (\href{mailto:970811119@qq.com}{970811119@qq.com}), Qihang Li, Haoyuan Zhang, Lingfeng Wang, Wei Chen, Aling Luo

\noindent \textbf{Affiliations:} Sanechips Technology Co., LTD

\section{Low-light RAW Image Enhancement}

Performing image enhancement under low-light conditions poses several challenges, such as degradation of details, color distortion, and severe noise, which significantly affect the quality of images \cite{wei2020physics, wei2021physics, fu2022gan, zhang2021learning}. Meanwhile, compared to the 8-bit camera's sRGB output, the RAW data has not been processed by the Image Signal Processor (ISP); thus, it can retain the linearity with the scene and more unquantified information \cite{zou2023rawhdr, fu2023raw}. Based on the advantages of RAW data, the CVPR 2024 PBDL Challenge Low-light RAW Image Enhancement aims to assess and enhance algorithms' robustness on images captured in low-light environmental conditions to address the challenge of image quality degradation.

In the low-light RAW image enhancement track (Table \ref{tab: track5_track}), the top two teams demonstrated exceptional performance. Miers achieved total scores of 30.11, 31.13 dB in PSNR, and 0.84 in SSIM. ISS achieved total scores of 25.95, 27.09 dB in PSNR, and 0.82 in SSIM. These results highlight the remarkable advancements made by the participating teams in addressing the challenges of low-light RAW image enhancement.

\begin{table}[t]
	\centering
	\setlength{\tabcolsep}{8pt}
	\caption{Leaderboard of the low-light RAW image enhancement.}
	\label{tab: track5_track}
	\begin{threeparttable}
		\begin{tabular}{cccccc}
			\toprule
			\textbf{Rank} & \textbf{Team} & \textbf{Scores} & \textbf{PSNR} & \textbf{SSIM} \\ \hline
			1 & Miers & 30.11 & 31.13 & 0.84 \\ 
			2 & ISS & 25.95 & 27.09 & 0.82 \\ \bottomrule
		\end{tabular}
	\end{threeparttable}
\end{table}

\begin{figure}[tbp]
  \centering
  \includegraphics[width=1\linewidth]{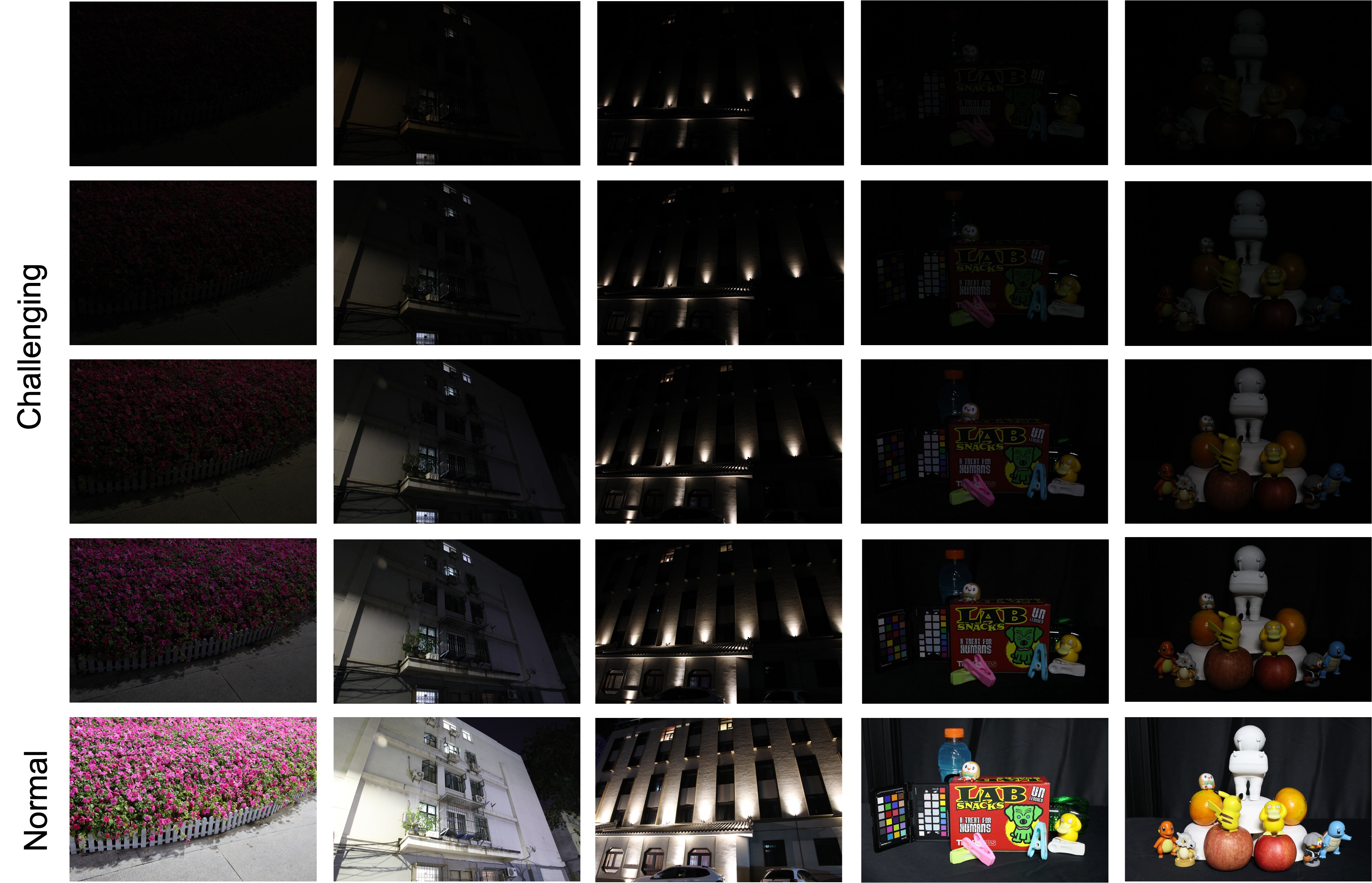}

   \caption{Example scenes in our captured RAW Image dataset. The last row is the reference images, and above it is the low-light image at 4 different ratios.}
   \label{fig:track5_dataset}
\end{figure}

\subsection{Low-light RAW Image Dataset}

To systematically investigate the effectiveness of the proposed method in real-world conditions, a real low-light image dataset for enhancement is necessary and fundamental.

We use Canon EOS 5D Mark IV to capture the data. To capture low/normal-light image pairs, the camera was mounted on a sturdy tripod and controlled remotely via a mobile APP. The camera was not touched between the capture process of normal-light and low-light images to avoid vibration. For each pair, we first take the normal-light image and fix ISO and aperture. Then the low-light images are captured by changing the shutter (exposure time) to simulate low-light conditions. We capture our dataset indoor and outdoor to increase the richness of the scene, where include both natural scenarios and manual builds. The dataset exhibits the following characteristics:
\begin{itemize}
	\item \textbf{Paired samples.} The dataset includes images in RAW format, which consists of a normal-light reference image and four low-light images at different ratios (8,16,32,64).
	\item \textbf{Diverse scenes.} The dataset contains 832 image pairs in 208 scenes. Our dataset stands out with its high resolution of $6720\times4480$, surpassing the common resolutions (below $1920\times1080$) found in other datasets. This higher resolution captures finer details, offering a more comprehensive analysis for low-light enhancement.
\end{itemize}

The dataset includes images captured in indoor and outdoor scenes under varying lighting conditions as shown in Fig.~\ref{fig:track5_dataset}.

\subsection{Miers Team's Method}

\subsubsection{Network Architecture}

\begin{figure}[tbp]
  \centering
  \includegraphics[width=1\linewidth]{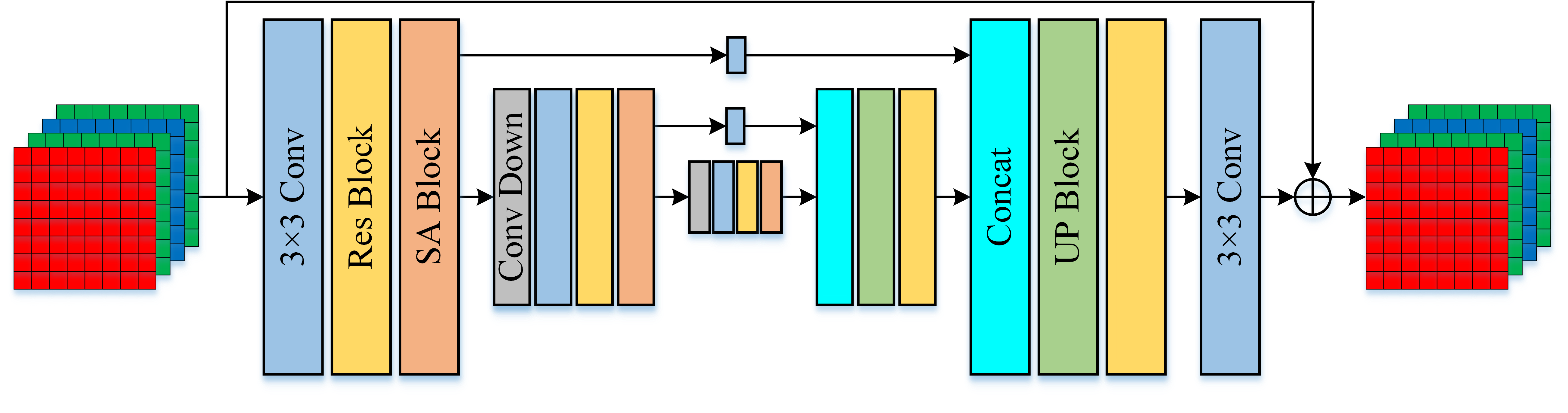}

   \caption{The proposed method.}
   \label{fig:track5_method}
\end{figure}

The Miers proposed a multi-scale, light-weight transformer model for low-light raw image enhancement. Unlike previous Retinex-based methods \cite{cai2023retinexformer, li2018structure, Yi2023ICCV} that generally decompose the input image into illumination components and reflection components, the proposed method adaptively aligns brightness-induced differences by introducing a learnable guidance vector in the self-attention mechanism. The network architecture called SANet is shown in Fig.~\ref{fig:track5_method}. The SANet extracts features at different scales sequentially and performs feature fusion through long connections, which can effectively reduce the calculation. At each scale, the proposed method uses concatenated residual blocks and the SABlock as basic modules to obtain non-local views.

\begin{figure}[tbp]
  \centering
  \includegraphics[width=1\linewidth]{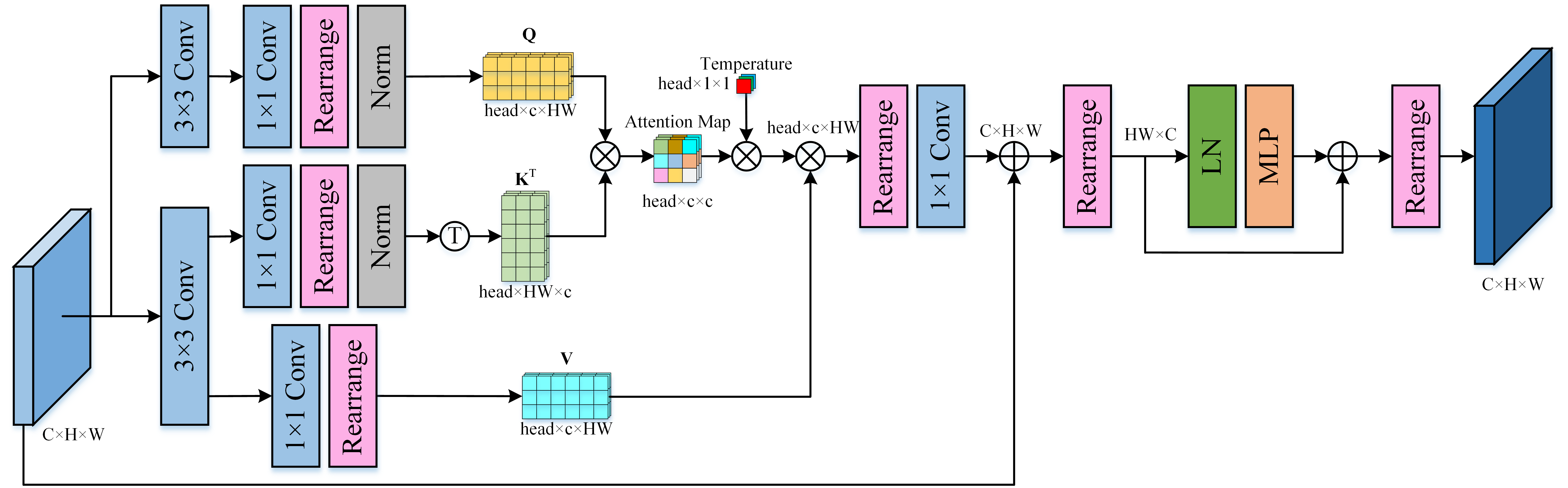}

   \caption{The structure of the SABlock.}
   \label{fig:track5_SABlock}
\end{figure}

The self-attention mechanism in Transformer structure has been proven to have great advantages in low-level image enhancement, and the proposed SABlock (shown in Fig.~\ref{fig:track5_SABlock}) is also based on this. The SABlock captures global dependency information by building key-value pairs on feature blocks. In the low-light enhancement task, due to the large differences in input image distribution caused by illumination, the model is not easy to fit for natural state. This team introduced a learnable adaptive vector in SABlock to control the gap between the input RAW and the target. This allows the model to be effectively fitted to the direction that contributes to the correct output.

It is also worth noting that downsampling in SANet is implemented using $4 \times 4$ convolution with $ stride=2 $ and the UP Block consists of a $3 \times 3$ depthwise separable convolution and pixelshuffle.

\subsubsection{Implementation Details}

The code is based on BasicSR \cite{wang2022basicsr} and EFNet \citep{sun2022event}. During the training stage, only the data provided by the competition is used. First, the input image is cropped to 128x128, rotation and flipping are added as data augmentation. It should be noted that due to the large difference in image brightness under different ISOs, the input RAW image subtracts the black level and divides by the difference between the white level and the black level, and then multiplies by the ratio for normalization. The ratio is calculated by \[ ratio=\frac{1}{\max( \frac{\text{raw\_image - black\_level}}{\text{white\_level - black\_level}} )}  \]

In the training process, the batch size is 4, total iterations is set to 500,000. This team uses L1 loss as the training loss and MultiStepLR for learning rate decay. In addition, the model weight uses exponential moving average (EMA), and the model with the highest PSNR on the validation set is finally selected for testing.

\subsection{ISS Team's Method}

\subsubsection{Network Architecture}

The ISS used the algorithm proposed in the Lighting Every Darkness in Two Pairs: A Calibration-Free Pipeline for RAW Denoising \cite{jin2023lighting}, which can adapt to the target camera without calibrating noise parameters and repeated training, requiring only a small amount of lens pairing data and fine-tuning, eliminating the complicated calibration steps, and achieved good performance. 

\begin{figure}[tbp]
  \centering
  \includegraphics[width=1\linewidth]{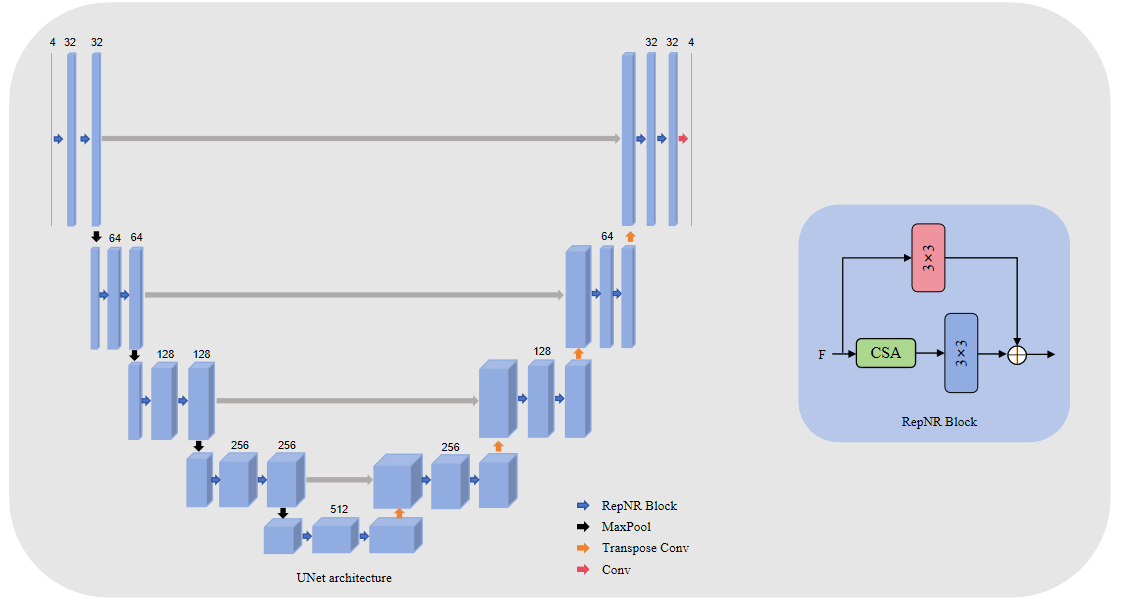}

   \caption{Overview of network architecture.}
   \label{fig:track5_network architecture}
\end{figure}

As shown in Fig.~\ref{fig:track5_network architecture}, the whole network adopts the macro architecture of Unet \cite{ronneberger2015u}, in which the convolution blocks of the Unet network itself is replaced with the reparameterized noise removal (RepNR) block \cite{jin2023lighting}. In the Pre-train stage, In RepNR Block has k branches of Camera-Specific Alignment (CSA) module \cite{jin2023lighting}, Each of these branches is fitted to a class of camera noise, In the Fine-tune phase, By averaging the k CSA module, Equivalent to the model integration of noise from multiple classes of cameras, At this time the RepNR block consists of two branches, Where the upper 3x3 convolution is designed to fit the out-of-model noise, Lower Camera-Specific Alignment (CSA) module, The main role is to adjust the distribution of the input features. 

\subsubsection{Implementation Details}

By utilizing LED methods \cite{jin2023lighting}, we ultimately reduced the number of scenes to just four groups, randomly selecting three distinct images for each scene to pair with their respective ground truth images. These image pairs were rapidly deployed to a new camera using the provided pre-trained weights. 

Subsequently, fine-tuning was conducted using a small amount of real data, with the RepNR block replacing the convolutional layers in the UNet architecture. During the fine-tuning process, we initially iterated the CSA from the pre-trained model for 5000 iterations until convergence. The optimizer used was Adam with a learning rate of 0.0001, and the training strategy employed a cosine annealing approach. An additional branch was then fine-tuned for an additional 3000 iterations, with the optimizer and training strategy consistent with the main branch. The loss function chosen was L1 loss.

\subsection{Teams and Affiliations}
\noindent\textbf{Miers}

\noindent\textbf{Title:} A Light-weight Aligned Attention for Low-light Raw Enhancement

\noindent\textbf{Members:} Cheng Li (\href{mailto:licheng8@xiaomi.com}{licheng8@xiaomi.com}), Jun Cao, Shu Chen, Zifei Dou

\noindent\textbf{Affiliations:} Xiaomi Inc., China

~\\

\noindent\textbf{ISS}

\noindent\textbf{Title:} Low-light Raw Image Enhancement Technical solution

\noindent\textbf{Members:} Xinyu Liu, Jing Zhang (\href{mailto:jingzhang\_work@163.com}{jingzhang\_work @163.com}, Kexin Zhang, Yuting Yang, Licheng Jiao, Shuyuan Yang

\noindent\textbf{Affiliations:} Intelligent Perception and Image Understanding Lab, Xidian University



%


\section{HDR Reconstruction from a Single Raw Image}

The dynamic range of real-world scenes frequently exceeds the capture capabilities of standard consumer camera sensors, often resulting in loss of detail in both overly bright and dark areas. In underexposed regions, noise becomes significant and affects the visual quality~\cite{chen2018learning,yue2020supervised,wei2021physics,zou2022estimating}, while in overexposed regions, information is often clipped~\cite{huang2022exposure,fu2023raw,afifi2021learning}. To address this, the computational imaging community has extensively explored High Dynamic Range (HDR) imaging, which records a broader spectrum of intensity levels and captures more scene information. Unlike conventional Low Dynamic Range (LDR) images, HDR preserves greater detail in both over- and under-exposed areas. This enhancement not only benefits various vision tasks, such as segmentation~\cite{martinez2017image} and object detection~\cite{onzon2021neural,wang2024multi}, but also produces more visually pleasing images—a goal long pursued by computer vision researchers.

To advance HDR reconstruction research, we are launching a challenge focused on reconstructing HDR images from single Raw images. This approach specifically targets single Raw image HDR reconstruction, avoiding potential misalignments that can occur in multi-image fusion. We will utilize a Raw-to-HDR dataset that focus on HDR reconstruction from a single Raw image, as shown in Fig. \ref{fig:real_data}, which contains pairs of Raw and HDR images. The Raw input is captured under challenging lighting conditions, representing the over- and under-exposed regions of a high dynamic range scene. The corresponding ground truth HDR images in the dataset are produced through bracketed exposures of each scene, subsequently merged using basic HDR fusion algorithms~\cite{debevec2008recovering}.

\begin{figure*}[h!] \small
	\centering
		\includegraphics[width=1\linewidth,clip,keepaspectratio]{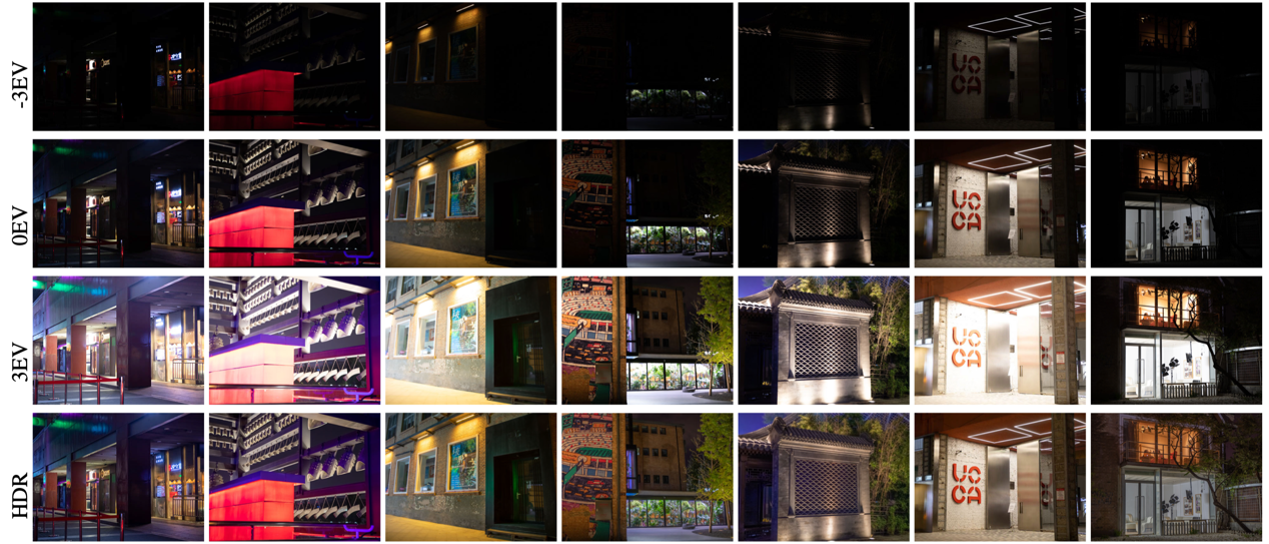}
	\caption{Representative examples for Raw-to-HDR dataset SRHDR (Single Raw HDR). For each scene, a Raw image captured under challenging lighting is served as input image, and the HDR image that is merged by bracket exposures are used as ground truth.}
	\label{fig:real_data}
\end{figure*}

\subsection{Dataset}

We capture and curate a real paired Raw-to-HDR dataset called SRHDR for HDR reconstruction from a single Raw image. The SRHDR dataset extended RawHDR~\cite{zou2023rawhdr} in both quantities and difficulty, and covers a large range of HDR scenarios including modern and ancient buildings, art districts, tourist attractions, street shops and restaurants, abandoned factories, city views, and more. These images are captured at different times of the day, including daytime and nighttime, which further guarantees the diversity of the paired Raw-to-HDR dataset. The data capture process involves several steps. Initially, we carefully select scenes with high dynamic range potential. Then, using a Canon 5D Mark IV camera mounted on a tripod, we employ bracket exposure mode to capture different exposures of the same scene. The Raw images taken in challenging lighting conditions, specifically from -3EV to +3EV, are used as input images. The corresponding ground truth images are created using an HDR merging method, as described by Debevec~\etal~\cite{debevec2008recovering}. The challenge dataset has the following characteristics

\begin{itemize}
	\item \textbf{High resolution.} Our dataset stands out with its high resolution of 6720x4480, surpassing the common resolutions (below 1920x1080) found in other HDR datasets. This higher resolution captures finer details, offering a more comprehensive analysis for HDR reconstruction.
	\item \textbf{High bit-depth ground truth.} The SRHDR dataset features ground truth HDR images with a bit-depth of over 20 bits, utilizing a linear HDR format. This high bit-depth ensures a richer and more precise representation of color and light intensities.
	\item \textbf{Real paired samples.} Each image pair in the dataset is meticulously captured through multi-exposure fusion. The input comprises actual images shot with a DSLR under challenging lighting conditions. The corresponding ground truth HDR images are generated using a widely accepted HDR merging algorithm, ensuring authenticity and relevance.
	\item \textbf{Raw images as input.} The use of unprocessed Raw sensor data as the input format leverages the higher bit-depth and superior intensity tolerance of Raw data, effectively addressing the common issue of insufficient scene information in HDR image processing.
\end{itemize}

\begin{table}[t]
	\centering
	\footnotesize
	\setlength{\tabcolsep}{5pt}
	\caption{Results and rankings of top-2 competitors.}
	\begin{threeparttable}
		\begin{tabular}{ccccccc}
			\toprule
			\textbf{Rank} & \textbf{Team} & \textbf{Score} & \textbf{PSNR} & \textbf{SSIM} & \textbf{PSNR-$\mu$} & \textbf{MS-SSIM} \\ \hline
			1 & Alanosu & 66.90 & 34.02 & 0.94 & 33.43 & 0.97 \\ 
			2 & USTCX & 65.82 & 32.33 & 0.95 & 33.89 & 0.98 \\ 
			\bottomrule
		\end{tabular}
	\end{threeparttable}
\end{table}

\subsection{Alanosu Team's Method}
\subsubsection{Network Architecture}

\begin{figure*}[tb]
	\centering
	\includegraphics[width=0.8\linewidth]{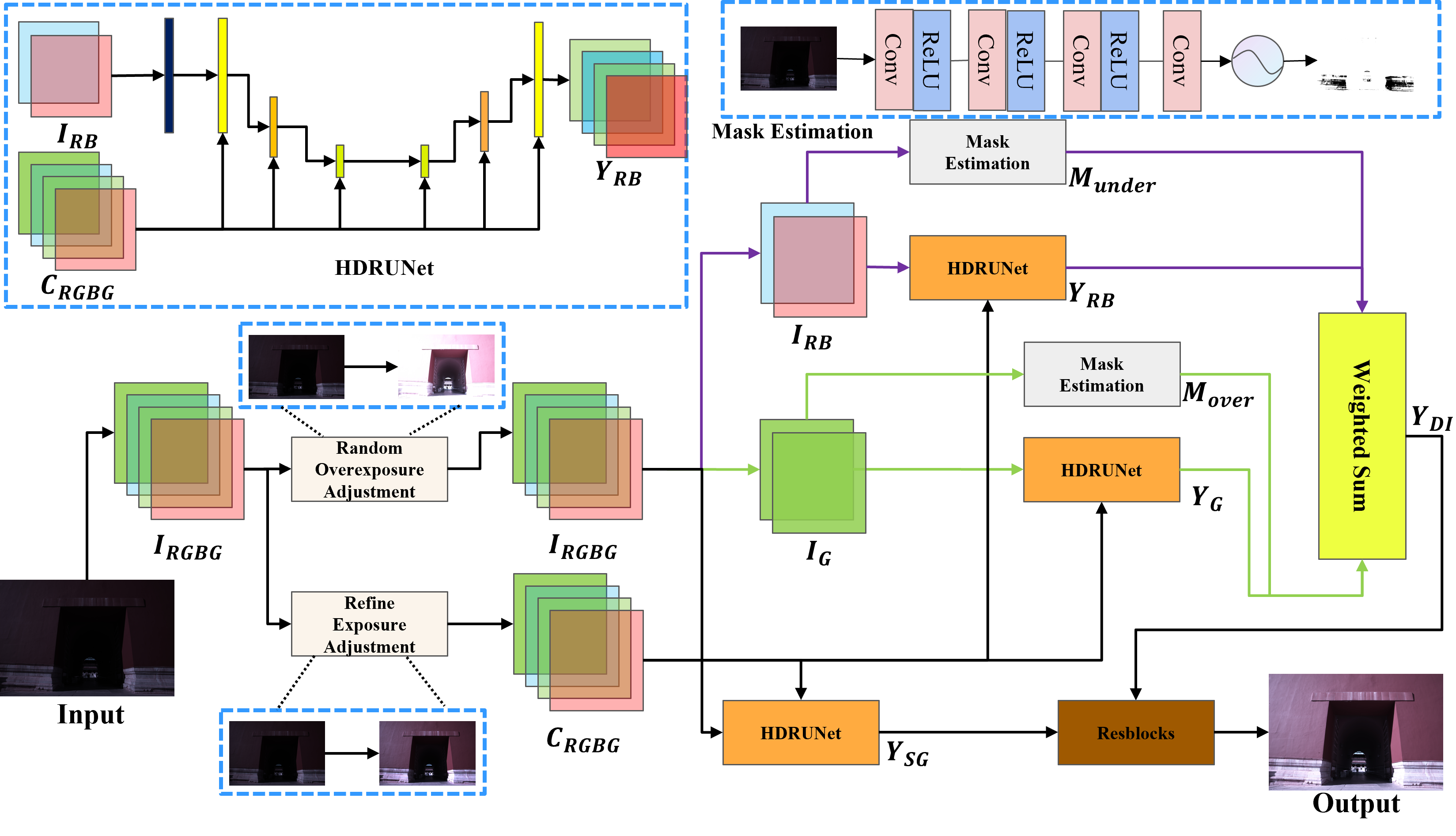}
	\caption{The Network Architecture of our proposed MHDRUNet.}
	\label{fig:1}
\end{figure*}

We introduces MHDRUNet, a model that utilizes the emphasis on different channel information in Raw images for exposure guidance, subsequently used for HDR reconstruction of single-frame Raw images. The model framework is shown in Fig.~\ref{fig:1}. Inspired by RawHDR\cite{zou2023rawhdr}, Raw images have higher intensity values in green channels compared to red and blue. Therefore, we split $I_{RGBG}$ into $I_{RB}$ and $I_{G}$, using the RB channels for exposure estimation to derive an underexposure mask $M_{under}$, and then reconstruct the underexposed areas based on the G channel to get $Y_{G}$. Similarly, we use the G channel for exposure estimation to obtain an overexposure mask $M_{over}$, and then reconstruct the overexposed regions using the RB channels to get $Y_{RB}$. We combine $Y_{G}$ and $Y_{RB}$ through a weighted sum. To ensure smoothness in HDR reconstruction across the global range, we utilize the original Raw data for global exposure-guided reconstruction. The exposure reconstruction network is comprised of the complete HDRUNet\cite{chen2021hdrunet}, with inputs including the Raw image and a condition image, which by default matches the Raw image. The exposure estimation mask module consists of a CNN with residual connections, and the exposure reconstruction is carried out by the complete HDRUNet network.
Secondly, we propose the method of Refine Exposure Adjustment. By analyzing the distribution of values in the input Raw image, we can estimate the areas of exposure and underexposure. For images where the area of the underexposure mask is greater than that of the overexposure mask, we consider it to be underexposed; conversely, if larger for the overexposure mask, it is considered overexposed. Based on this, we make appropriate exposure adjustments on the original Raw data, bringing underexposed images to a slightly underexposed state and overexposed images to a slightly overexposed state. These are then used as the condition images inputted into the HDRUNet network, thereby achieving better reconstruction results.

\subsubsection{Implementation Details}
We use the dataset proposed by HDR Reconstruction from a Single Raw Image challenge. Before training, we pre-process the data by cropping images into $768 \times 768$. During training, the mini-batch size is set to 1 and Adam \cite{kingma2014adam} optimizer and Kaiming-initialization \cite{he2015delving} are adopted for training. The initial learning rate is set to $1e-4$ and all models are built on the PyTorch framework and trained with NVIDIA 3090 GPU. It's noteworthy that we find the HDR reconstruction task for overexposed images to be more challenging compared to that for underexposed images. Therefore, we propose the training strategy of Random Overexposure Adjustment. Specifically, during training, we randomly apply varying degrees of overexposure adjustments to the input underexposed images to generate pseudo-overexposed images for data augmentation, thereby enhancing the model's robustness. During training, we use tanh L2 loss\cite{chen2021hdrunet} and SSIM loss~\cite{wang2004image} to achieve better training effects. Additionally, we employ a constraint loss $L_{mask}$ \cite{zou2023rawhdr}, which guides the learning of the mask.

\subsection{USTCX Team's Method}
\subsubsection{Network Architecture}

\begin{figure*}[t]
	\centering
	\includegraphics[width=\linewidth]{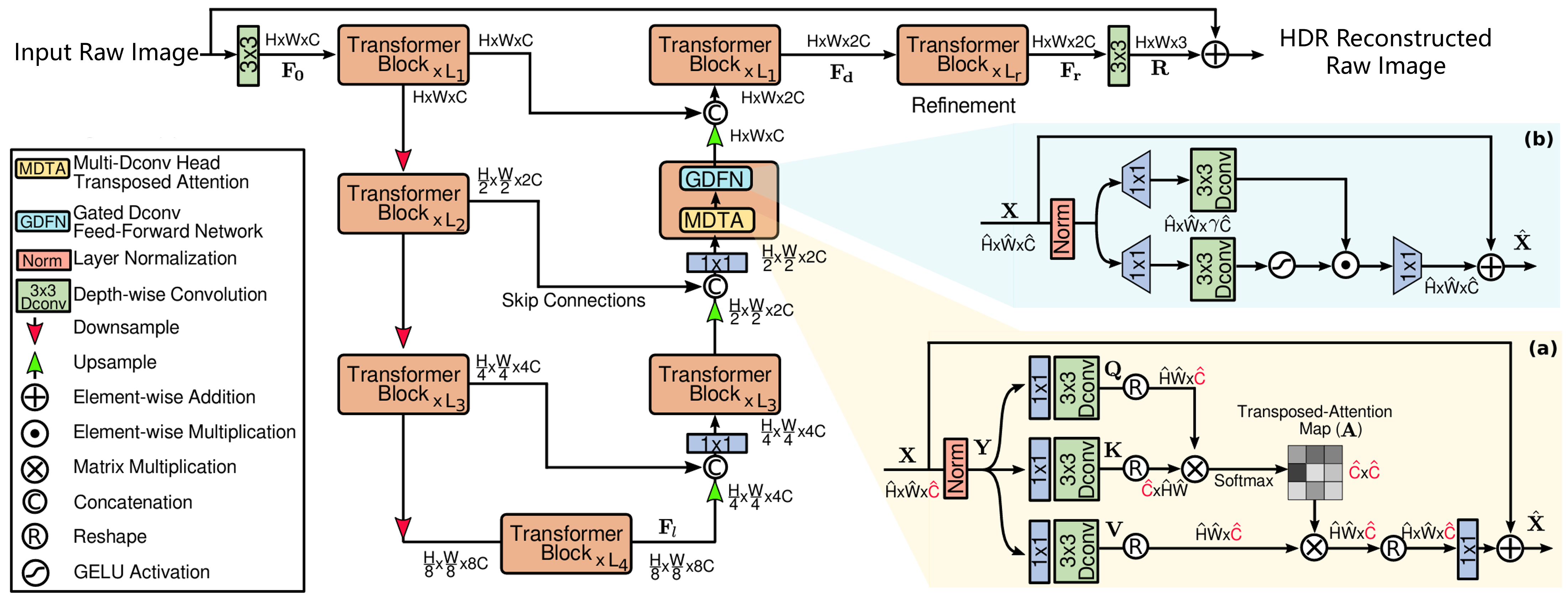}
	\caption{The Restormer Framework~\cite{zamir2022restormer} for HDR Reconstruction.}
	\label{fig:my_diagram}
\end{figure*}

\textbf{Overall Pipeline.} Inspired by Restormer \cite{zamir2022restormer}, our network architecture is shown in Fig.~\ref{fig:my_diagram}. It overcomes the limitations of traditional Convolutional Neural Networks~(CNNs) by utilizing Transformers' ability to capture long-range pixel interactions, which is crucial for High Dynamic Range~(HDR) image reconstruction. We introduce the two modules of the Transformer block: \textbf{(a)} multi-Dconv head transposed attention~(MDTA) and \textbf{(b)} gated-Dconv feed-forward network~(GDFN).
\textbf{Multi-Dconv Head Transposed Attention} This module replaces the standard multi-head self-attention mechanism. It operates across feature dimensions instead of spatial dimensions, reducing complexity. It uses $1\times1$ convolutions for pixel-wise cross-channel context aggregation and $3\times3$ depth-wise convolutions for channel-wise spatial context encoding.

\vspace{0.8mm}
\noindent\textbf{Gated-Dconv Feed-Forward Network} This module includes a gating mechanism and depth-wise convolutions to control information flow and encode local image structures. The gating mechanism is an element-wise product of two linear transformation paths, one activated with GELU non-linearity. 

\vspace{0.8mm}
\noindent\textbf{Loss Functions}  In our work, we use the Charbonnier loss~\cite{9025464} to optimize our network. This loss function is particularly effective for handling outliers and robust to noise. Its formulation is as follows:

\begin{equation}
	\mathcal{L}_{\mathrm{content}} = \sqrt{\left\|\hat{I}_{\mathrm{}}-I_{\mathrm{}}\right\|_{2}+\epsilon^{2}}
\end{equation}

where $\hat{I}$ is the predicted HDR Raw image, $I$ is the ground truth, and $\epsilon$ is set to 0.0001 as default. 

In addition to the content loss, we leverage frequency domain information to introduce auxiliary loss to our network, which is defined as follows:

\begin{equation}\mathcal{L}_{\text{frequency}}=\left\|\mathcal{F}\left(\hat{I}_{\text{}}\right)-\mathcal{F}\left(I_{\text{}}\right)\right\|_{1}\end{equation}

where $\mathcal{F}\left(\cdot\right)$ indicates the Fast Fourier Transform~(FFT). Finally, the total loss could be defined as:

\begin{equation}\mathcal{L}_{\mathrm{total}} =\mathcal{L}_{\mathrm{content}} + \lambda\mathcal{L}_{\mathrm{frequency}}\end{equation}

where $\lambda$ denotes the balanced weight, and we empirically set $\lambda$ to 0.5 as default.

\subsection{Implementation details}

We utilized PyTorch 1.8 within an NVIDIA 3090 GPU environment, equipped with 24GB of memory, to train our model on official datasets with a batch size of 4. The input images were standardized to an $80 \times 80$ resolution. The training spanned approximately 23 hours, with a learning rate that started at $3 \times 10^{-4}$, reduced to $1 \times 10^{-7}$ over 75,000 iterations using a Cosine Annealing schedule. This was followed by a second phase with a learning rate of $6 \times 10^{-5}$, also reduced to $1 \times 10^{-7}$ over an additional 60,000 iterations. Notably, no special efficiency optimization strategies were applied during this process.

\subsection{Teams}

\noindent\textbf{Alanosu}

\noindent\textbf{Title:} MHDRUNet

\noindent\textbf{Members:} Liwen Zhang, Zhe Xu (\href{mailto:xuzhenwpu@126.com}{xuzhenwpu@126.com}), Dingyong Gou, Cong Li

\noindent\textbf{Affiliations:} ZTE Corporation.

\vspace{1em}

\noindent\textbf{USTCX}

\noindent\textbf{Title:} Restormer

\noindent\textbf{Members:} Senyan Xu (\href{mailto:syxu@mail.ustc.edu.cn}{syxu@mail.ustc.edu.cn}), Yunkang Zhang, Siyuan Jiang

\noindent\textbf{Affiliations:} University of Science and Technology of China.



%


\section{Highspeed HDR Video Reconstruction from Events}
Event cameras, differing from conventional cameras that capture scene intensities at a fixed frame rate, use a unique approach by detecting pixel-wise intensity changes asynchronously. This is triggered whenever a pixel's intensity change surpasses a certain contrast threshold. Unlike traditional frame-based cameras, event cameras have several advantages: low latency, low power consumption, high temporal resolution, and high dynamic range (HDR). These qualities make them particularly useful for a range of vision tasks, including real-time object tracking~\cite{saner2014high,ramesh2018long,zhang2021object}, high-speed motion estimation~\cite{lee2014real}, optical flow estimation \cite{gallego2018unifying}, ego motion analysis~\cite{ye2020unsupervised}, and so on.

However, the distinct triggering mechanism of event cameras presents a challenge. The event data they capture, which lacks absolute intensity values and is represented as 4-tuples, is incompatible with standard frame-based vision algorithms. This discrepancy necessitates specialized processing pipelines, different from traditional image processing methods. Consequently, there is a growing interest in transforming event data into intensity images to leverage the high-speed and HDR capabilities of event cameras in practical applications~\cite{wang2020event, yang2023learning}.

To this end, we are launching a challenge focused on reconstructing high-speed HDR videos from event streams. We will utilize the high-quality Event-to-HDR dataset, captured by a co-axis system and developed by \cite{zou2021learning}. This dataset includes aligned pairs of event streams and HDR videos in both spatial and temporal dimensions.

In the challenge evaluation, three evaluation metrics are used for assessment: Peak signal-to-noise ratio (PSNR), tone-mapped PSNR (PSNR-$\mu$), Structural
Similarity (SSIM) and multi-scale SSIM (MS-SSIM).
The training dataset consists of 300 paired LDR/HDR images.  The input images of validation and testing sets are provided, while the GT are not available to participants. The final leaderboard of top-3 participants are shown in Table \ref{tab:track7_results}.

\begin{table}[t]
	\centering
	\setlength{\tabcolsep}{6pt}
	\caption{Results and rankings of methods.}
	\label{tab:track7_results}
	\begin{threeparttable}
		\begin{tabular}{cccccc}
		\toprule
		\textbf{Rank} & \textbf{Team} & \textbf{Score} & \textbf{PSNR} & \textbf{SSIM} & \textbf{RMSE} \\ \hline
		1 & IVISLAB & 16.56 & 18.52 & 0.73 & 0.13 \\ 
		2 & Jackzou & 16.21 & 18.50 & 0.70 & 0.13 \\ 
		3 & apolloUI & 16.21 & 18.39 & 0.70 & 0.13 \\ \bottomrule
	\end{tabular}
	\end{threeparttable}
\end{table}

\subsection{Dataset}
The dataset for this challenge is shown in Fig.~\ref{fig:real_data}. This dataset provides real paired Event-to-HDR data for the purpose of high-speed HDR video reconstruction from event streams. The collection process involves an integrated system designed to simultaneously capture high-speed HDR videos and corresponding event streams. This is achieved by utilizing an event camera to record the event streams, alongside two high-speed cameras that capture synchronized Low Dynamic Range (LDR) frames. These LDR frames are later fused to create High Dynamic Range (HDR) frames. The careful alignment of these cameras within the system ensures the accurate synchronization of the high-speed HDR videos with the event streams, offering a robust dataset for the challenge participants. The challenge dataset has the following characteristics

\begin{figure*}[h!] \small
	\centering
	\setlength{\tabcolsep}{1pt}
	\begin{tabular}{cccccccccccc}
		& \multicolumn{3}{c}{Scene 1} & & \multicolumn{3}{c}{Scene 2} & & \multicolumn{3}{c}{Scene 3}\\
		\rotatebox{90}{Low bits} &
		\includegraphics[width=.1\linewidth,clip,keepaspectratio]{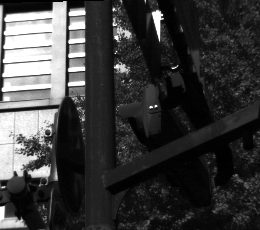} & \includegraphics[width=.1\linewidth,clip,keepaspectratio]{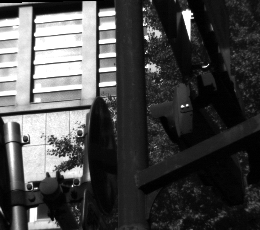} & \includegraphics[width=.1\linewidth,clip,keepaspectratio]{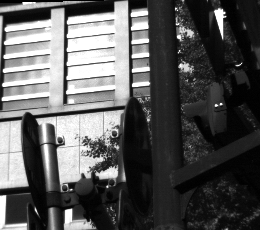} && \includegraphics[width=.1\linewidth,clip,keepaspectratio]{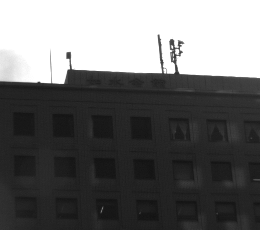} & \includegraphics[width=.1\linewidth,clip,keepaspectratio]{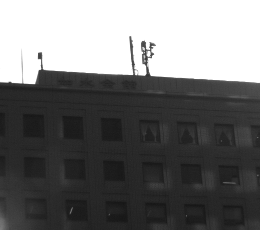} & \includegraphics[width=.1\linewidth,clip,keepaspectratio]{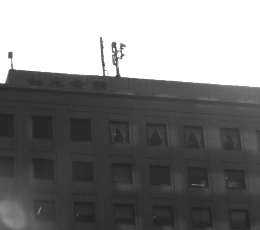} && \includegraphics[width=.1\linewidth,clip,keepaspectratio]{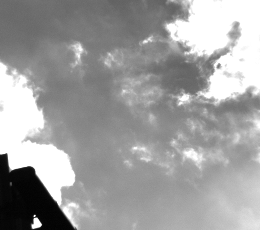} & \includegraphics[width=.1\linewidth,clip,keepaspectratio]{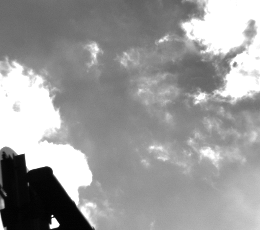} & \includegraphics[width=.1\linewidth,clip,keepaspectratio]{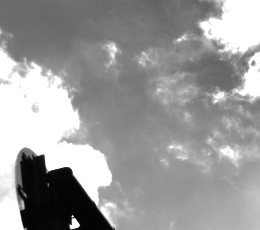}\\
		\rotatebox{90}{High bits} &
		\includegraphics[width=.1\linewidth,clip,keepaspectratio]{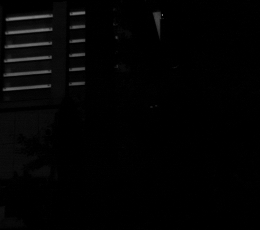} & \includegraphics[width=.1\linewidth,clip,keepaspectratio]{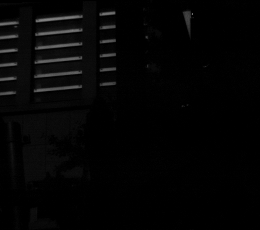} & \includegraphics[width=.1\linewidth,clip,keepaspectratio]{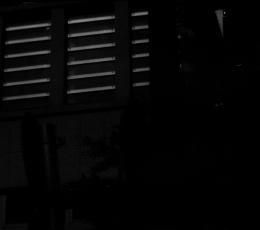} && \includegraphics[width=.1\linewidth,clip,keepaspectratio]{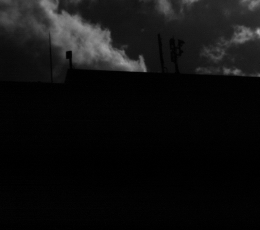} & \includegraphics[width=.1\linewidth,clip,keepaspectratio]{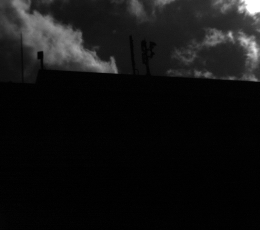} & \includegraphics[width=.1\linewidth,clip,keepaspectratio]{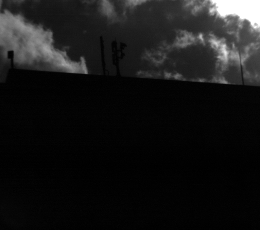} && \includegraphics[width=.1\linewidth,clip,keepaspectratio]{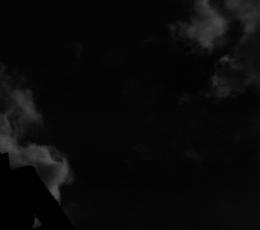} & \includegraphics[width=.1\linewidth,clip,keepaspectratio]{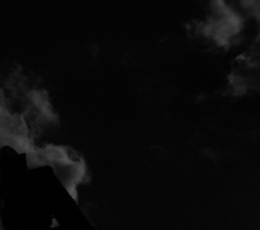} & \includegraphics[width=.1\linewidth,clip,keepaspectratio]{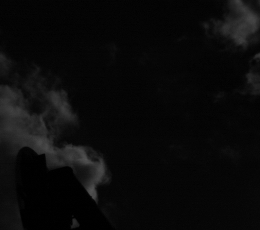}\\
		\rotatebox{90}{ \; HDR} &
		\includegraphics[width=.1\linewidth,clip,keepaspectratio]{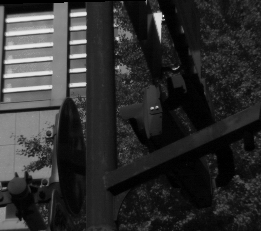} & \includegraphics[width=.1\linewidth,clip,keepaspectratio]{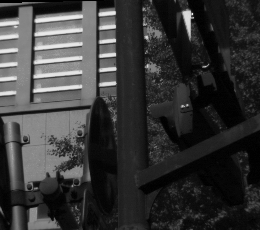} & \includegraphics[width=.1\linewidth,clip,keepaspectratio]{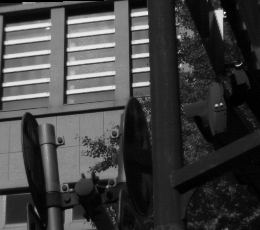} && \includegraphics[width=.1\linewidth,clip,keepaspectratio]{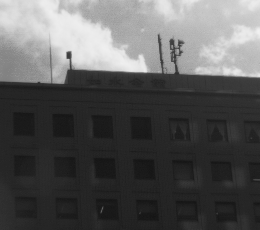} & \includegraphics[width=.1\linewidth,clip,keepaspectratio]{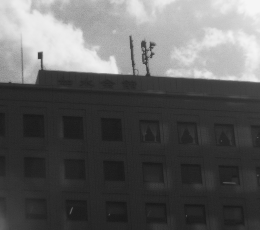} & \includegraphics[width=.1\linewidth,clip,keepaspectratio]{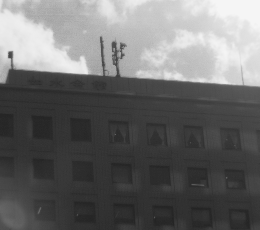} && \includegraphics[width=.1\linewidth,clip,keepaspectratio]{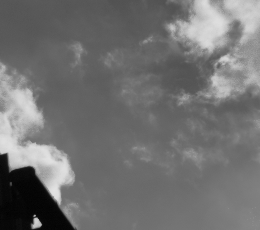} & \includegraphics[width=.1\linewidth,clip,keepaspectratio]{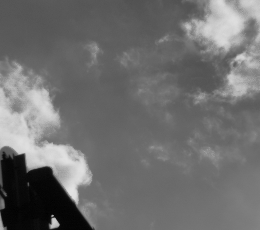} & \includegraphics[width=.1\linewidth,clip,keepaspectratio]{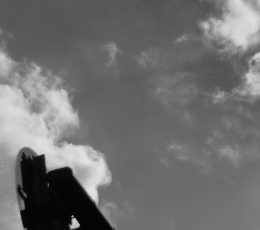}\\
		& Frame 1 & Frame 2 & Frame 3 & & Frame 1 & Frame 2 & Frame 3 & & Frame 1 & Frame 2 & Frame 3\\
	\end{tabular}
	\caption{An overview of the challange dataset: Event-to-HDR video dataset.}
	\label{fig:real_data}
\end{figure*}

\begin{itemize}
	\item \textbf{Real high-bit HDR.} Unlike existing methods that primarily leverage the HDR feature of event data, our dataset includes real high-bit HDR data. This data is created by fusing two images with different exposures using an HDR fusion strategy. This inclusion is crucial as most current methods do not use real high-bit depth HDR data for training, limiting their ability to generate such HDR formats.
	\item \textbf{Paired Event-to-HDR dataset.} While existing datasets often contain only paired testing data created by simulating a virtual camera's trajectory, this dataset provides real paired training data. This approach overcomes the domain gap that synthetic training data typically has with real-world testing scenarios. This dataset captures genuine paired training data, offering a more realistic and applicable training environment.
	\item \textbf{Highspeed.} In alignment with the high-speed nature of event streams, our videos are captured with a high-speed camera at a frame rate of 500fps. This speed significantly exceeds that of APS or any other event-to-HDR dataset, making our dataset uniquely suited for applications requiring high temporal resolution.
\end{itemize}

\subsection{IVISLAB Team's Method}
\begin{figure}[t]
	\centering
	\includegraphics[width=1.\linewidth]{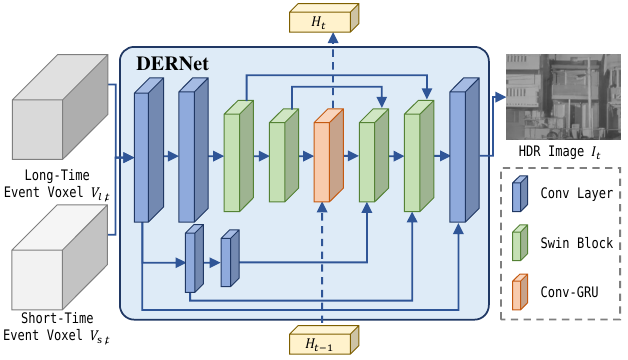}
	\caption{Architecture of DERNet}
	\label{fig:dernet}
\end{figure}

To achieve high-speed HDR video reconstruction from events, our team introduces the Dual Event-stream  Reconstruction Network (DERNet). As depicted in Figure~\ref{fig:dernet}, DERNet uses long-time and short-time event voxels to reconstruct the low-frequency brightness and high-frequency texture of HDR video. Furthermore, DERNet integrates Swin Transformer and Conv-GRU blocks to capture spatial and temporal contexts, thereby enhancing reconstruction accuracy.

\subsubsection{Network Architecture}
DERNet adopts an encoder-decoder network with a recursive design to process dual-stream event voxels to estimate high-speed HDR videos. Specifically, when reconstructing the $t$-th frame of the HDR video, considering that the long-time event stream around frame $t$ can help reconstruct the low-frequency brightness, DERNet voxelizes the event data from frame $t-T_l$ to frame $t+T_l$ into a $b$-bins event voxel $V_{l,t}$. Simultaneously, considering that the short-time event stream around frame $t$ can help reconstruct high-frequency texture, DERNet voxelizes the event data from frame $t-T_s$ to frame $t+T_s$  into a $b$-bins event voxel $V_{s,t}$. Subsequently, the event voxels $V_{l,t}$ and $V_{s,t}$ are concatenated and input into the network. To fuse the features of the two event voxels, DERNet utilizes convolutional layers to generate fused features from the event voxels. The network then adopts a two-branch encoder. This structure includes a complex branch that extracts high-level semantic information from the fused features, leveraging Swin Transformer~\cite{liu2021swin} blocks to capture spatial context and Conv-GRU blocks to capture temporal context by integrating historical states. It also includes a simple branch that utilizes convolutional layers to capture detailed information from the fused features. Next, the decoder of DERNet adopts multiple Swin Transformer blocks to fuse and upsample the features extracted by the two-branch encoder, finally using convolutional networks to predict the $t$-th frame HDR image $I_{t}$.

\subsubsection{Implementation Details}
To train DERNet, a reconstruction loss $\mathcal{L}_r$ is designed for the estimated HDR image $\mathcal{L}_{t}$ as
\begin{equation}
	\begin{aligned}
		L_r &= \lambda_1 {\rm L_1}(I_{t}, I_{t}^{gt}) +\lambda_2 {\rm L_1}(M(I_{t}), M(I_{t}^{gt}))  \\
		&\quad + \lambda_3 {\rm L_2}(I_{t}, I_{t}^{gt}) + \lambda_4 {\rm L_2}({\rm M}(I_{t}), M(I_{t}^{gt}))
	\end{aligned}
\end{equation}
where $\lambda_1$, $\lambda_2$, $\lambda_3$, and $\lambda_4$ coeﬃcients balancing the loss terms, ${\rm L_1}(\cdot,\cdot)$ is the absolute loss function, ${\rm L_2}(\cdot,\cdot)$ is the mean squared error loss function, $I_{t}^{gt}$ is the ground truth $t$-th frame HDR image, and ${\rm M}(\cdot)$ is the HDR to SDR function defined as ${\rm M}(x) = \frac{\log(1 + 5000x)}{\log(5001)}$.

DERNet is implemented using PyTorch. During training, a batch size of 2 is utilized, with a video sequence length of 10 and a data size of $224 \times 224$. An AdamW optimizer~\cite{loshchilov2018decoupled} is adopted with a learning rate of $4 \times 10^{-5}$ and weight decay of $10^{-6}$ to optimize the network weights for 60 epochs. A cosine annealing scheduler is adopted to decay the learning rate. To prevent overfitting, random flipping, rotation, and cropping are applied to the event voxels for data augmentation. The coefficients are defined as $b = 6$, $T_l = 16$, $T_s = 5$, $\lambda_1 = 1$, $\lambda_2 = 0.1$,$\lambda_3 = 500$,  and $\lambda_4 = 10$.

\subsection{Jackzou Team's Method}

\begin{figure*}[ht]
	\centering
	\includegraphics[width=\linewidth, clip, keepaspectratio]{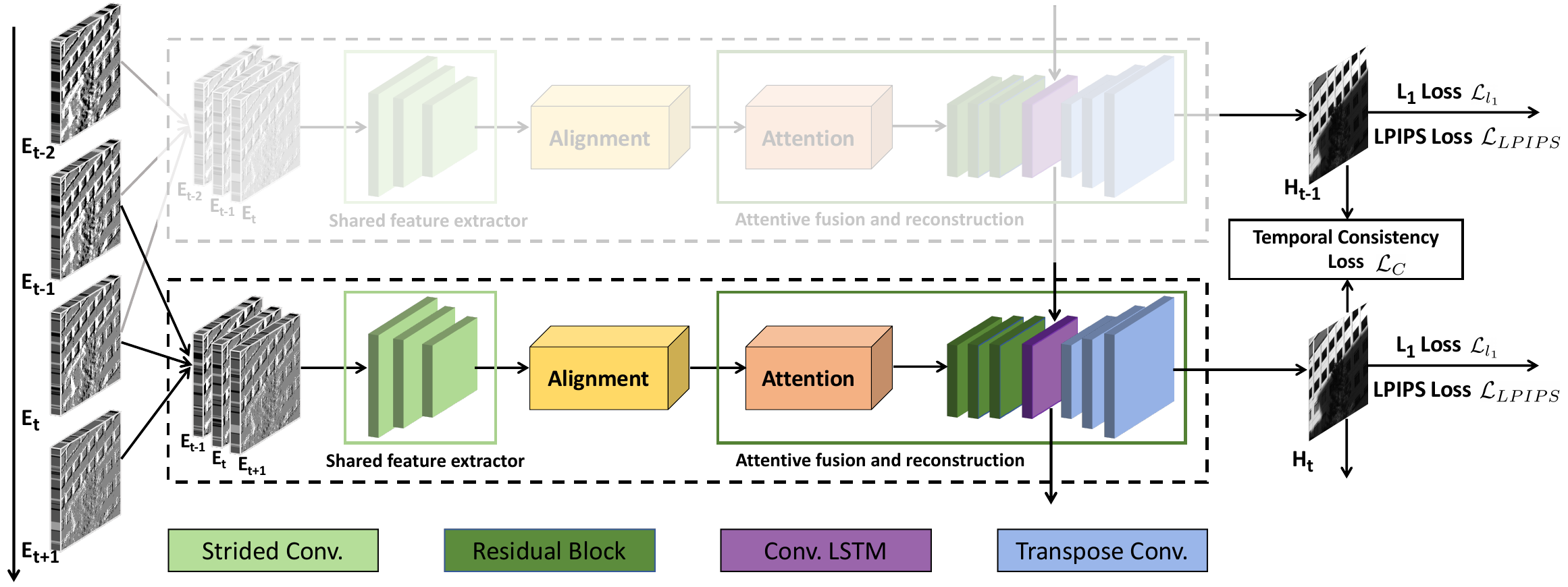}
	\caption{The overview of  recurrent convolutional neural network for HDR video reconstruction from events.}
	\label{fig:overview}
\end{figure*}

\subsubsection{Network Architecture}
Our method employs a convolutional recurrent neural network designed to reconstruct HDR videos from event streams~\cite{zou2021learning}. As shown in Figure~\ref{fig:overview}, the network processes $T=2N+1$ consecutive event voxel grids $\{\mathbf E_{t-N}, \ldots, \mathbf E_{t+N}\}$ to generate the HDR frame $\mathbf H_t$ at timestamp $t$. The architecture includes several key modules.

Firstly, the shared feature extractor downsamples event frames to a low spatial resolution feature space using strided convolution layers, producing $2N+1$ output feature maps, $\{\mathbf F_{t-N}, \ldots, \mathbf F_{t+N}\}$. This shared encoding facilitates subsequent alignment by transforming the input data into a consistent feature space.

We then employ a deformable convolution-based alignment module~\cite{wang2019edvr}, which uses pyramidal deformable convolutions to align features of different event frames with the central frame feature $\mathbf F_t$. This approach predicts offsets for the convolution kernels through a pyramidal processing structure, allowing the network to handle larger movements and align features accurately, thereby avoiding the pitfalls of inaccurate optical flow estimation.

The aligned features are combined in the attentive fusion and reconstruction module. Here, the features are stacked and processed by attention mechanisms that independently focus on height, width, and temporal/channel correlations. The fused features are passed through a recurrent residual network and a ConvLSTM~\cite{shi2015convolutional} module, which help maintain temporal continuity by remembering information from successive sequences.

To enhance temporal consistency, we introduce a novel temporal consistency loss based on the integral relationship between consecutive frames and events, modeled using a pre-trained UNet-like~\cite{unet2015} network. This loss ensures smooth transitions between frames, mitigating issues related to temporal discontinuity.

\subsubsection{Implementation Details}
The training strategy involves a combination of losses to optimize HDR video reconstruction and maintain temporal consistency. Given the reconstructed video sequence $\mathbf H_i$ and the corresponding ground truth frames $\hat{\mathbf H}_i$, we employ several loss functions.

We use the $l_1$ loss to measure the pixel-wise difference between the reconstructed and ground truth frames:
\begin{equation}
	\mathcal L_{l_1} = \sum_{i=1}^{T} \Vert \mathbf H_i - \hat{\mathbf H}_i \Vert.
\end{equation}
To enhance perceptual quality, we introduce the Learned Perceptual Image Patch Similarity (LPIPS) loss~\cite{zhang2018perceptual}, which focuses on high-level and structural similarity. Additionally, the temporal consistency loss $\mathcal L_{C}$, derived from the pre-trained network, is defined as:
\begin{equation}
	\mathcal L_{C} = \sum_{i=1}^{T} \Vert \mathbf E_t - \mathcal C(\mathbf H_{t-1}, \mathbf H_t) \Vert_2^2.
\end{equation}
This loss ensures that the reconstructed frames maintain smooth transitions.

The overall loss function used to train our model is:
\begin{equation}
	\mathcal L = \mathcal L_{l_1} + \tau_1 \mathcal L_{LPIPS} + \tau_2 \mathcal L_{C},
\end{equation}
where $\tau_1$ and $\tau_2$ are empirically set to $2$ and $0.2$, respectively.

The network is initialized using Kaiming initialization~\cite{he2015delving}, and trained with the Adam optimizer (momentum set to 0.9). The initial learning rate is $10^{-4}$, reduced by a factor of 10 every 50 epochs. We set the batch size to 4, and train the model for 100 epochs. The implementation uses the PyTorch framework, and training is performed on NVIDIA TITAN V GPUs.

In summary, our network architecture and training strategy effectively reconstruct high-quality HDR videos from event data, ensuring both spatial and temporal coherence.

\subsection{ApolloUI Team's Method}

\subsubsection{Network Architecture}

\begin{figure}
	\centering
	\includegraphics[width=0.95\linewidth]{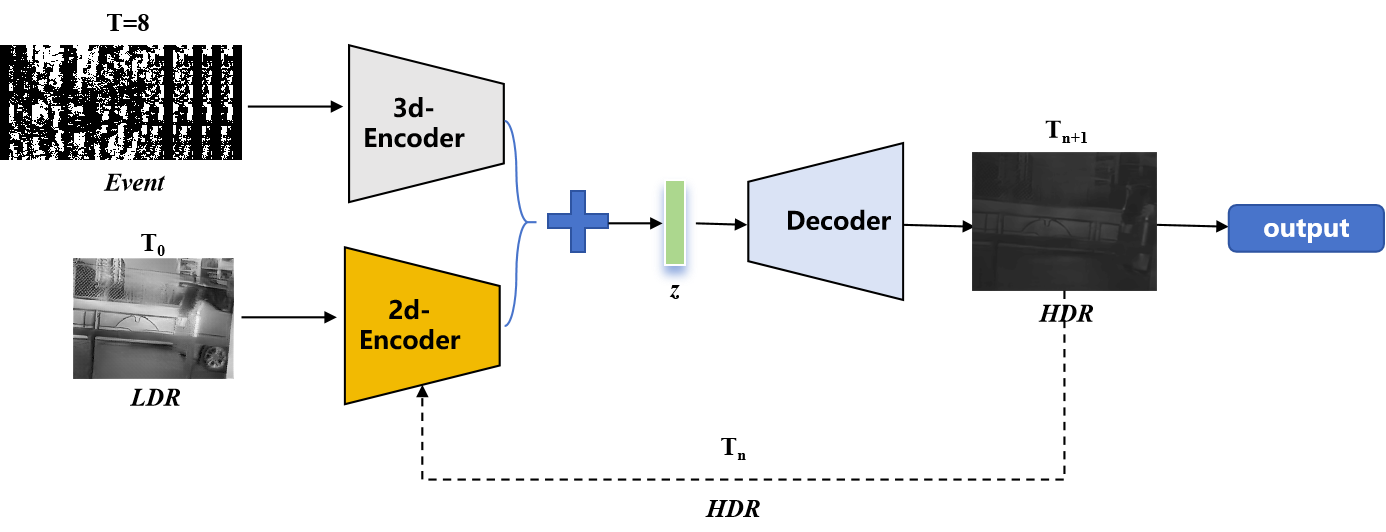}
	\caption{\label{fig:network}The network architecture.}
\end{figure}

As shown in the figure \ref{fig:network}, we introduce three types of inputs at the input end. The first part is data generated from the voxel structure of raw event camera data, with the time step of 8. This part of the data represents the most primitive detail information of the image. The second part is 2D image data generated by E2VID~\cite{rebecq2019events}, which serves as the reference frame for LDR 2D images. The third part is HDR image data generated by the decoder, representing the HDR data of historical frames, which can help smooth the entire video and ultimately generate HDR image data for the current frame.

\subsubsection{Implementation Details}
According to the aforementioned network architecture, this method is trained using the following approach. Firstly, the training data is converted into voxels based on the original timestamps and stored in the \textit{npy} format. For the purpose of facilitating temporal training, each \textit{npy} file is stored with a length of T=16. Subsequently, e2vid is employed to generate reference frames corresponding to each \textit{npy} file. It is noteworthy that the reference frames generated by E2VID are often affected by the actual data distribution density, leading to occurrences of blank spaces or excessive noise.

Regarding the network itself, data augmentation techniques such as random horizontal flipping and the addition of Gaussian noise with a standard deviation of 0.001 are applied to the input data. Additionally, to better align with the evaluation metrics of this challenge, four types of noise are introduced, including KL divergence noise (to ensure alignment between HDR ground truth images and generated images), L1 noise, L2 noise, and SSIM noise. Due to time constraints, rigorous ablation experiments were not conducted. However, from a holistic analysis of the results, KL divergence noise yielded relatively favorable gains.

Other training parameters include the Adam optimizer with a learning rate of 0.001, cosine annealing for learning rate scheduling, a batch size of 48, and iterative training conducted using four NVIDIA 4090 GPUs. Training is performed for 1000 epochs, with the overall training and testing dataset split in an 8:2 ratio. To address challenging samples, this method manually removes data with poor distributions (some data lack original event information due to bandwidth congestion, rendering them unsuitable for training). 
Finally, inference can be performed, followed by truncating and normalizing the image with a maximum value of 65535, resulting in the HDR image of the current moment in the corresponding distribution domain.
\subsection{Teams}
\noindent\textbf{IVISLAB}

\noindent\textbf{Title:}  Dual Event-stream Reconstruction Network

\noindent\textbf{Members:} Qinglin Liu\textsuperscript{1} (\href{mailto:qlliu@hit.edu.cn}{qlliu@hit.edu.cn}), Wei Yu\textsuperscript{1}, Xiaoqian Lv\textsuperscript{1}, Jianing Li\textsuperscript{2}, Shengping Zhang\textsuperscript{1}, Xiangyang Ji\textsuperscript{3}

\noindent\textbf{Affiliations:} 
\textsuperscript{1}Harbin Institute of Technology, 
\textsuperscript{2}Peking University,
\textsuperscript{3}Tsinghua University.

\vspace{1em}

\noindent\textbf{Jackzou}

\noindent\textbf{Title:} Learning to Reconstruct High Speed and High Dynamic Range Videosfrom Events

\noindent\textbf{Members:} Yunhao Zou (\href{mailto:zouyunhao@bit.edu.cn}{zouyunhao@bit.edu.cn}), Ying Fu

\noindent\textbf{Affiliations:}  Beijing Institute of Technology.

\vspace{1em}

\noindent\textbf{apolloUI}

\noindent\textbf{Title:} Generating High Dynamic Range Image Sequences with Event Cameras Based on Multi-Head Encoding Networks

\noindent\textbf{Members:} Yuanpei Chen (\href{mailto:ypchen\_code@outlook.com}{ypchen\_code@outlook.com}), Yuhan Zhang, Weihang Peng

\noindent\textbf{Affiliations:} Intelligent Science \& Technology Academy of CASIC.

\section{Overexposure Image Correction}
Over-exposure is a prevalent issue in digital camera sensor systems, caused by automatic exposure errors during image processing. This problem particularly arises in dynamic scenes with fluctuating brightness levels, \ie, a car exiting a tunnel or the sudden illumination of a dark environment.
Exposure correction aims to correct the brightness errors that occur during the image capture process~\cite{afifi2021learning,nsamp2018learning}.
However, most CCD or CMOS cameras can only capture a limited illumination range and will produce clipped or over-exposed pixels when sensor elements are saturated due to improper settings or physical constraints in sensors.
This largely degrades the essential details in bright areas of photographs as well as the image quality~\cite{assefa2014correction}.
Therefore, correcting the brightness and texture details of the over-exposed images becomes a crucial task to improve the visual aesthetics of captured images and the performance of downstream image processing applications.

On the overexposure correction track (Table~\ref{table:track8_final_results}), the top three teams have shown outstanding performance. Gxj ranked first with a comprehensive score of 21.58. Specifically, gxj achieved PSNR of 21.58 and SSIM of 0.95. CVCV achieved PSNR of 20.56 and SSIM of 0.94. LiGoxin achieved PSNR of 19.45 and SSIM of 0.92.

These results highlight the remarkable advancements made by the participating teams in addressing the challenges of over-exposure image correction. The top-ranking teams have showcased their expertise and innovation in developing robust algorithms that excel in over-lighted conditions, paving the way for future advance ments in computer vision research.
\begin{table}[t]
	\centering
	\setlength{\tabcolsep}{6pt}
	\caption{Leaderboard of the Over-exposure correction.}
	\label{table:track8_final_results}
	\begin{threeparttable}
		\small 
		\setlength{\tabcolsep}{10pt} 
		\begin{tabular}{cccccc}
			\toprule
			\textbf{Rank} & \textbf{Team} & \textbf{PSNR} & \textbf{SSIM} & \textbf{Score} \\ \hline
			1 & gxj & 21.85 & 0.95 & 21.58 \\ 
			2 & CVCV & 20.90 & 0.94 & 20.56 \\ 
			3 & LiGoxin & 19.45 & 0.92 & 18.95 \\ \bottomrule
		\end{tabular}
	\end{threeparttable}

\end{table}

\subsection{RAW based Over-Exposure Correction dataset}
\begin{figure}[t]
	\centering
	\includegraphics[width=0.43\textwidth]{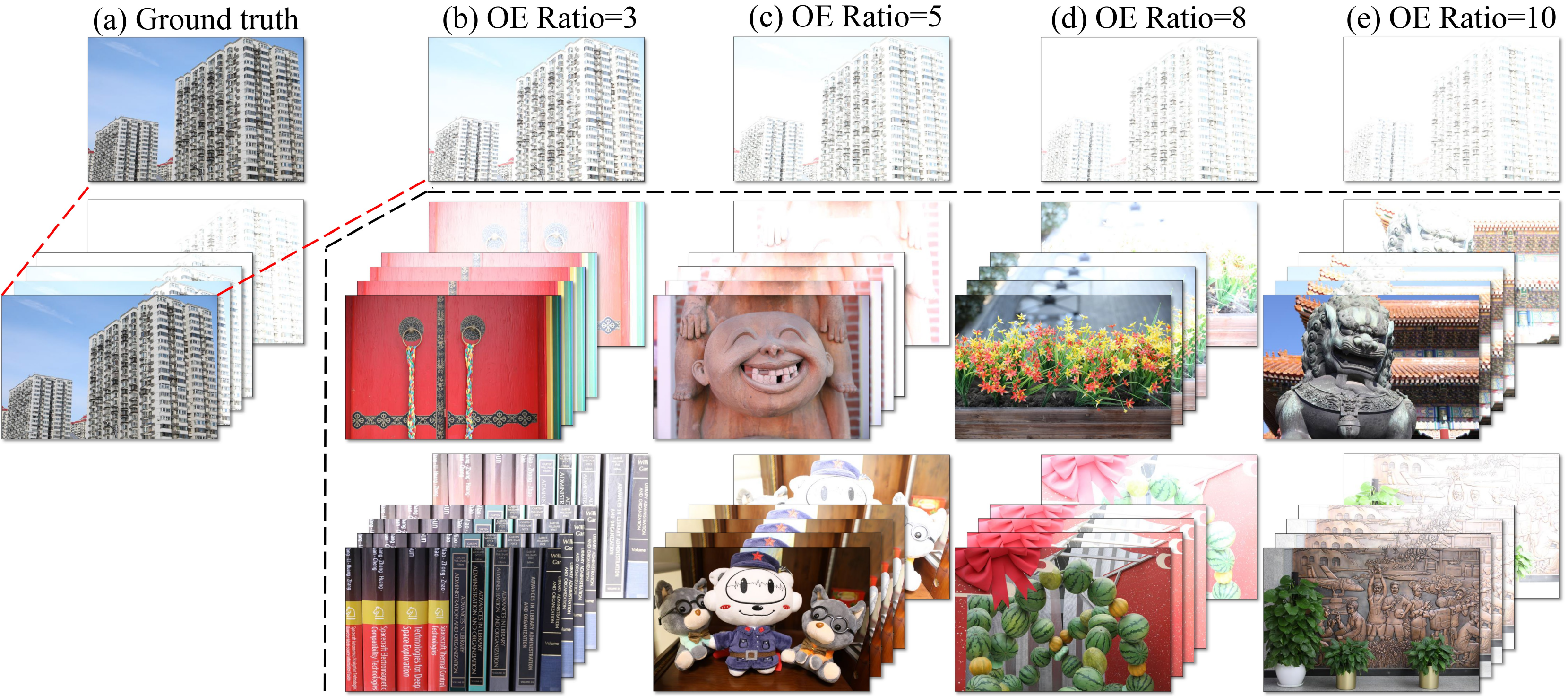}
	\centering
	\caption{\textbf{Examples in our Real-world Paired Over-exposure (RPO)~\cite{CGNet} dataset} include outdoor (the second row) and indoor (the third row) scenes.
	For each scene, we capture four different over-exposure ratios of 3, 5, 8, and 10 instances, both in RAW and sRGB formats.
	The most front image in the bottom two rows is the properly-exposed reference image.
	The behind images are correspondingly over-exposed images. 
	``OE" indicates Over-Exposure.}
	\label{fig:track8_dataset}
\end{figure}

To propel research in this field forward, it is essential to assess proposed methods in real-world scenarios. Consequently, we will utilize the RAW image-based Real-world Paired Over-exposure (RPO) dataset, introduced by Prof. Fu’s team in \cite{CGNet}, captured using a Canon EOS 5D Mark IV camera. The RPO dataset comprises paired images collected across various scenes. Each short-exposure (normal-exposure) image is paired with long-exposure (over-exposure) images with 4 ratios (x3, x5, x8, x10). Some representative examples of RPO dataset are shown in Fig.~\ref{fig:track8_dataset}.

The RPO dataset exhibits the following characteristics:
\begin{itemize}
\item \textbf{Short Exposure Images (Normal, GT)}: Captured in each scene using a tripod-mounted camera. The camera was set to automatic mode to find optimal aperture and exposure time settings, then switched to manual mode to lock these settings. Images were taken using a remote mobile app to control the shutter, minimizing lens vibration.
\end{itemize}
\begin{itemize}
\item \textbf{Long Exposure Images (Over-exposure, OE)}: Following the capture of short exposure GT images, only the "exposure time" setting was adjusted using the mobile app to simulate real over-exposure caused by incorrect settings. Four predetermined over-exposure ratios were used (×3, ×5, ×8, ×10). It was ensured that the camera was not touched during both long and short exposure captures to prevent any misalignment due to lens vibration.
\end{itemize}

\subsection{Gxj Team's Method}
\subsubsection{Network Architecture}
\begin{figure}[t]
    \centering
    \includegraphics[width=1\linewidth]{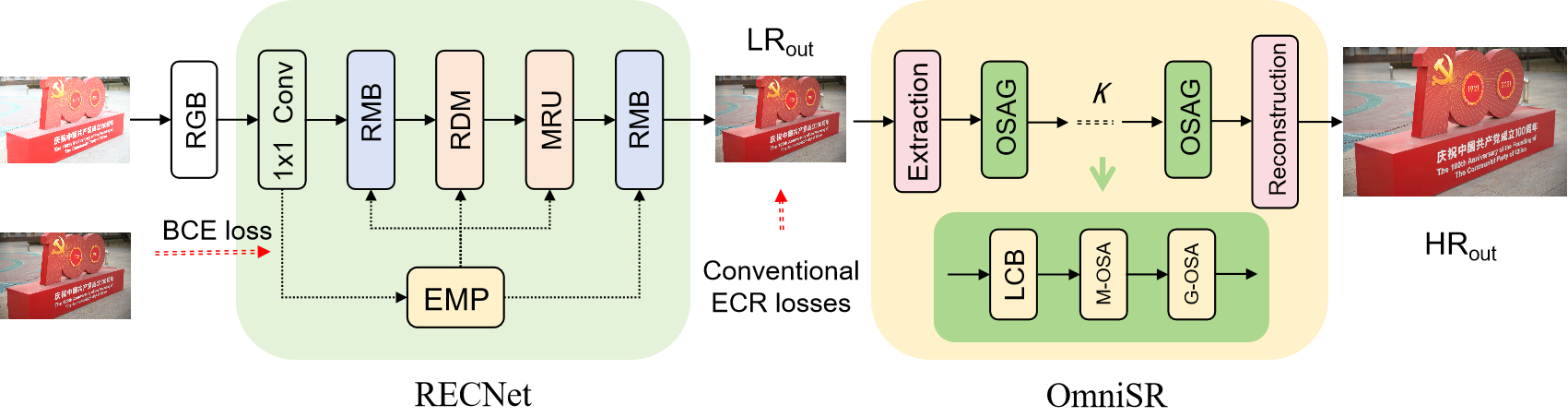}
    \centering
    \caption{Framework diagram of the model.}
    \label{fig:track_8_model1}
\end{figure}

Our solution of the whole task as shown in Fig.~\ref{fig:track_8_model1}, for the original excessive exposure image, first data pretreatment to RGB format, and then after area perception exposure correction network RECNet~\cite{REC} back to normal light state, the image light level is normal, but there are differences in resolution, after super resolution model OmniSR~\cite{Omni} will double the resolution to get the final result.

The key of the task is for the color correction of exposure area, but because the test file format directly with existing models, so we first convert the existing data set format, then choose compatible with low exposure and excessive exposure area perception exposure correction network (RECNet), through adaptive learning and bridging different area exposure representation to handle the mixed exposure, then the super resolution model (OmniSR) for the final result, finally achieve high excessive exposure image quality recovery.

\paragraph{\textbf{Data preprocessing}
}
Since the final test stage of this task provides the processed mat format files, in order to make the model better correct the test image, we further converted the image in mat format based on the existing RAW format data set and then converted the JPG format file which is more acceptable to the model. In general, the data set used for training is transformed into an image distribution file similar to the input of the test image. In this process, because the original image size is too large, we modified all the images to unify the size of the test set as the data for the training model.

\paragraph{\textbf{Exposure correction}}

Correction for image exposure has been studied for a long time. Traditional methods will rely mainly on manual adjustment of models, such as histogram equilibria and gamma correction. Although existing methods achieve commendable results in exposure correction, many of them rely on complex manual designs or struggle with excessive limitations that ultimately lead to suboptimal results.

After investigating the existing model and analyzing the data set, and also being inspired by the RECNet~\cite{REC} model, the exposure correction model used in this task was finally selected. When processing single images with mixed exposures, the network is difficult to stably converge, due to the large difference in over-and under-exposed regions, resulting in unbalanced performance for different exposures. To this end, the model takes into account the locality of different exposures to reduce the adverse effects of inconsistent optimization.To achieve this, the model adopt the idea of the divide and conquer strategy, and design aregion-aware exposure correction framework consisting of two well-designed modules concatenated in a chain of consecutive RMBs.

The model mainly contains a series of Blocks (RMB) with Region-aware De-exposure Module (RDM) and Mixed-scale Restoration Unit (MRU). The RDM maps exposure features Fin to a three-branched exposure-invariant feature Fn, while the MRU integrates the features Fs and Fc by the spatial-wise and channel-wise restoration, respectively. The exposure mask predictor (EMP) assists in generating the underexposure feature Fu and overexposure feature.It optimize the model with Exposure Contrastive Regularization (ECR).

\paragraph{\textbf{Image super resolution}
}
To match the results after exposure correction with the size of the resulting images required for the task, we used the Omni-SR~\cite{Omni} model to achieve a 2x super-resolution of the resulting images. Specifically, the model proposes a Omni Self-Attention (OSA) block based on the principle of dense interaction, which can model the pixel interaction from both the dimensions of space and channel, and mine the potential correlation between the global axis (i. e., space and channel). Combined with mainstream window partition strategies, OSA can achieve superior performance with a compelling computational budget. Second, a multi-scale interaction scheme is proposed to alleviate suboptimal ERFs in the shallow model, promoting local propagation and meso global scale interactions to form full-scale aggregate blocks.

\subsubsection{Implementation Details}

\begin{figure}[t]
	\centering
	\includegraphics[width=0.95\linewidth]{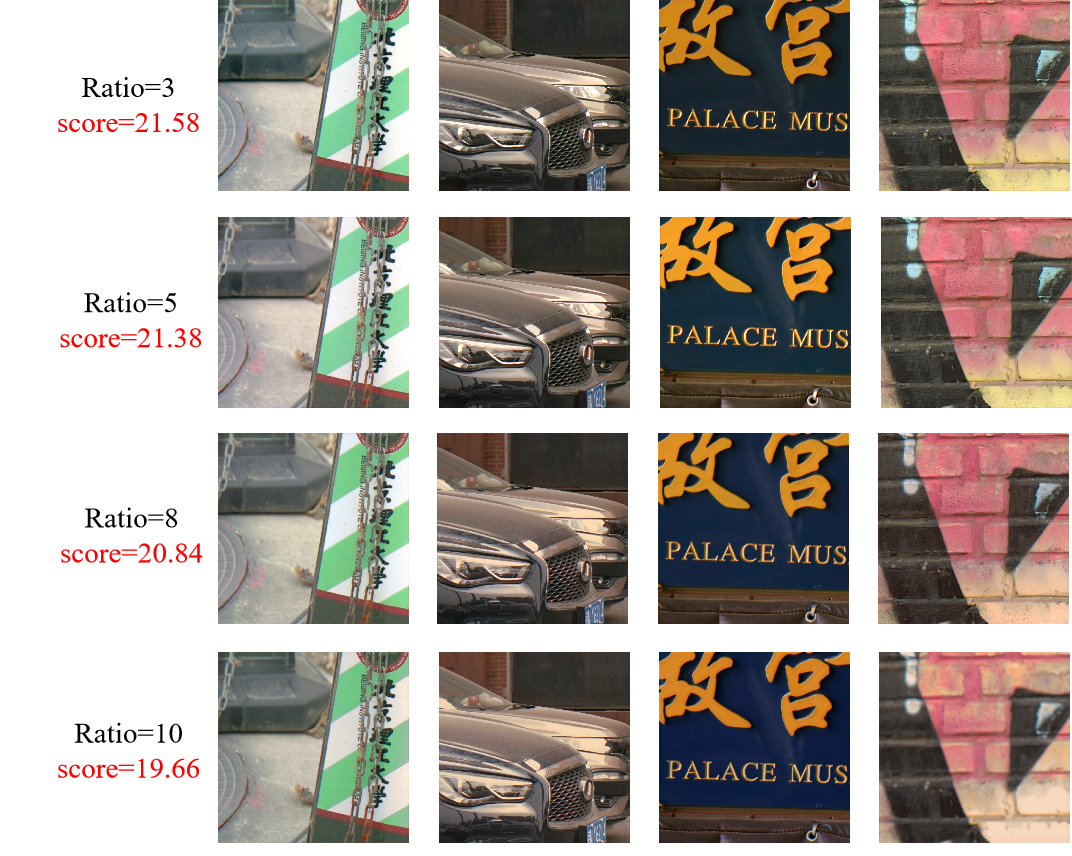}
	\centering\caption{Different ratio result image examples and scores.}
	\label{fig:track8_result1}
\end{figure}

We conducted experiments using two different models: RECNet and Omni-SR. For RECNet \cite{REC}, we processed and merged existing datasets, yielding a total of 1200 images, including 1120 images as the training set and 80 images as the validation set. No pre-training model was loaded, and training was done from scratch. The training parameters were a batch size of 8, a learning rate of 1e-4, and 300000 iterations, using a single NVIDIA RTX 4090 GPU. For Omni-SR\cite{Omni}, the pre-trained model of epoch885 with OmniSR on the DF2K dataset was used to treat the exposure-corrected images for 2x super-resolution. The experimental results are shown in Fig.~\ref{fig:track8_result1}, including the results and scores obtained by the original data after the recovery of the process structure.

The title is dedicated to the correction task of overexposed images. In this report, we detail our team's data processing methods and the use of models in this task. For the recovered images, the image quality is improved again through the super-resolution model, yielding better results. The experiments proved that our strategy to solve this task is reasonable and effective, ultimately achieving a score of 21.58 in the Ratio = 3 track of this task dataset.

\subsection{CVCV Team's Method}
\subsubsection{Network Architecture}
\begin{figure}[t]
    \centering
    \includegraphics[width=0.85\linewidth]{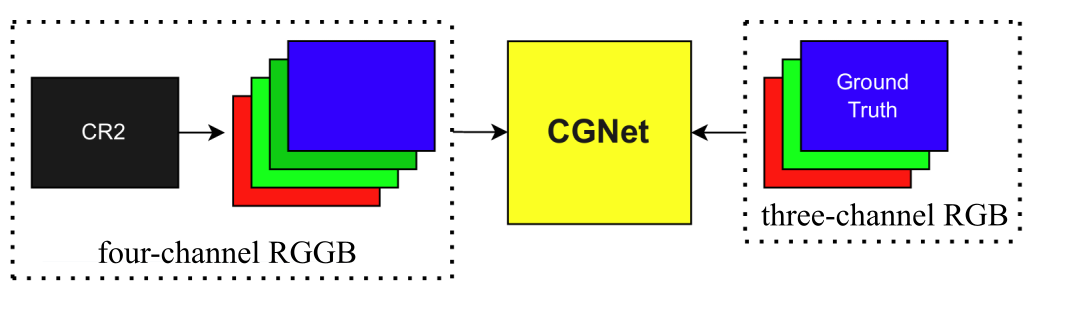}
    \caption{Training process diagram}
    \label{fig:track8_training2}
\end{figure}

Our training process for the entire task is shown in Fig.~\ref{fig:track8_training2}. For the original training dataset, it is first converted from CR2 to RGGB four-channel PNG. Then the data is input into the CGNet~\cite{CGNet} model and supervised learning is carried out by the GroundTruth of the RGB three-channel. Finally, the overexposed image can be corrected to make the overexposed image return to normal.

\begin{figure}[t]
    \centering
    \includegraphics[width=0.85\linewidth]{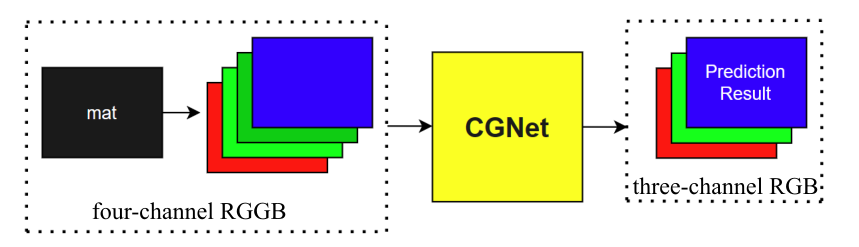}
    \caption{Test process diagram}
    \label{fig:track8_test2}
\end{figure}

Our test process for the entire task is shown in Fig.~\ref{fig:track8_test2}. For the original validation and test data set, it is first converted from mat format to RGGB four-channel PNG. Then the data is input into the CGNet model, the trained weights are loaded for model inference, and the inference results of RGB three-channel are obtained.

Most existing methods of overexposure in image correction have been developed based on sRGB images, which can lead to complex and non-linear degradation due to the image signal processing pipeline. Compared to sRGB-based technologies, RAW images are characterized by a near-linear correlation with scene brightness and exhibit superior performance due to the rich information content due to higher bit depth. Traditional digital camera sensors are designed to have a higher response ratio and relative spectral sensitivity to green channels. Therefore, in RAW images captured by most digital camera systems, the green channel is usually more likely to be overexposed in bright scenes than the red or blue channel. The red and blue channels of RAW images show more appropriate brightness and richer texture details than the green channels. This indicates that the green channel in the RGGB RAW image is more saturated than the red or blue channel and requires stronger correction.

Channel-Guidance Network(CGNet), which takes advantage of RAW images for overexposure correction. CGNet estimates correctly exposed sRGB images directly from overexposed RAW images in an end-to-end manner. Specifically, they introduce a RAW based channel guide branch into the U-Net-based backbone, which utilizes color channel intensity priors of RAW images to achieve superior overexposure correction performance.

\iftrue
\vspace{+0.918mm}
\noindent{\bf Data preprocessing.}
Our team chose CGNet model for overexposure image correction, and the model default input format is RGGB four-channel. In order to maintain the performance of the model, we decided to convert the original images in the dataset into RGGB format for training. The original training image is stored in CR2 format, and the rawpy library is directly called to batch convert CR2 files to RGB three-channel format. Then copy the green channel and convert it to RGGB four-channel format. Then, the overexposed images of four ratios (3,5,8,10) were input into CGNet together and divided into the training set and validation set according to the ratio of 8:2.

The storage format of the original test image is mat. After reading the mat file, it is found that the pixel value is between 0 and 1, and the four-channel is RGBG. Therefore, the pixels are multiplied by 255, and the four channels are converted to RGGB, and the processed PNG image is obtained.

\begin{figure*}[t]
    \centering
    \includegraphics[width=1.0\linewidth]{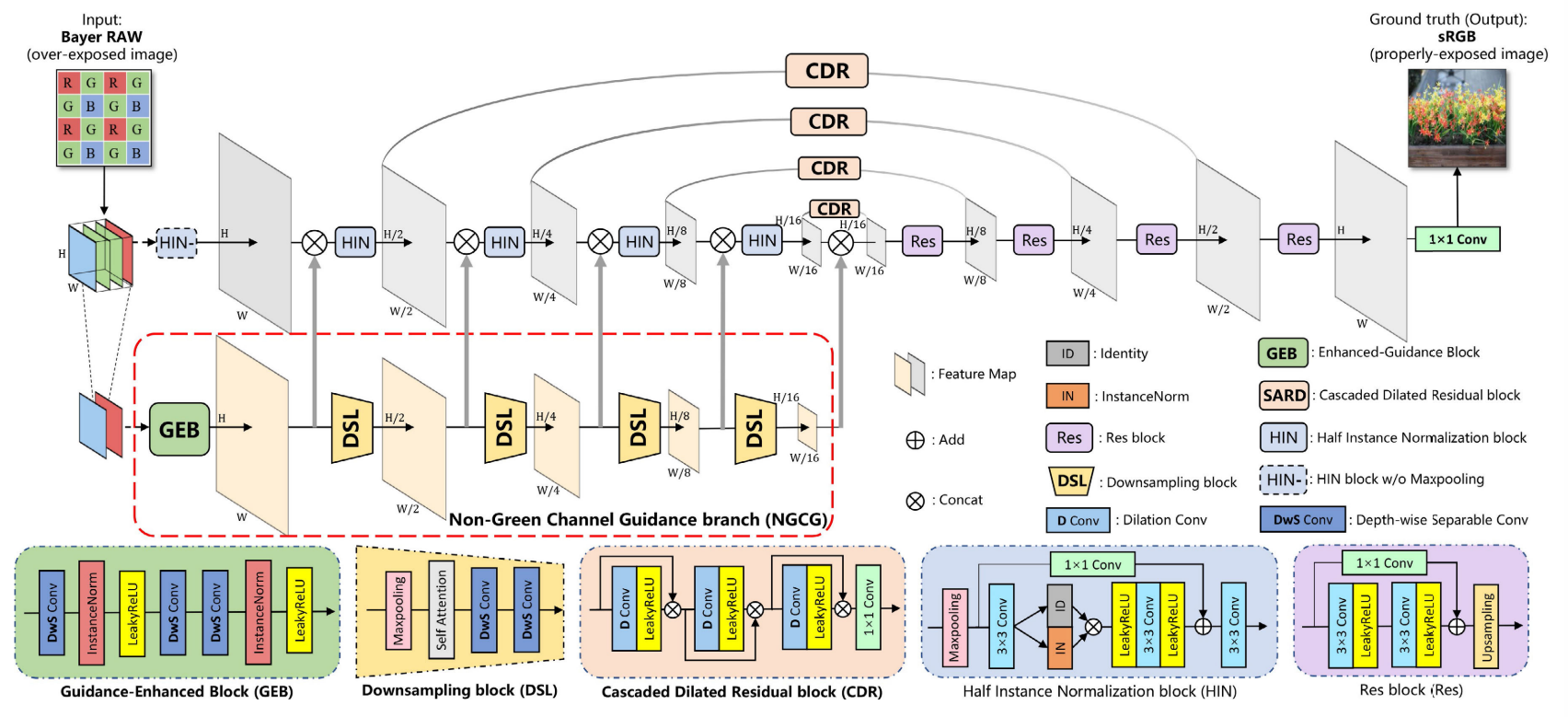}
    \caption{\textbf{Architecture of our Channel-Guidance Network (CGNet~\cite{CGNet}) for image over-exposure correction.} Given an over-exposed RAW image, they pad it into a four-channel RGGB image, and then feed it into their CGNet. Their CGNet is based on a U-Net backbone. The encoder consists of "Half-Instance Normalization", while the decoders are Residual blocks. They replace the original skip connection with Cascaded Dilated Residual (CDR) blocks. The red and blue channels are input to a Non-Green Channel Guidance (NGCG) branch for texture detail reconstruction. Their CGNet is pre-trained on their synthetic RAW image-based dataset, and fine-tuned on their Real-world Paired Over-exposure dataset.}
    \label{fig:track8_CGNET}
\end{figure*}

\iftrue
\vspace{+0.918mm}
\noindent{\bf CGNet.}
As shown in Fig.~\ref{fig:track8_CGNET}, the main branch is based on a basic U-net~\cite{ronneberger2015u} with four encoder (downsample) and decoder (upsample) stages. Specifically, they first extract the initial features from the four-channel RAW images using a standard 3×3 convolution. In the encoder section, the HIN blocks are utilized to broaden the receptive field and enhance the robustness of features at various scales. During the downsampling operation, they double the number of channels in the feature maps. Moving on to the decoder part, residual blocks are employed to capture high-level features more effectively. For the skip connection, they introduce a novel Cascaded Dilated Residual (CDR) block to extract multi-scale features, which are then merged with the encoder's features to mitigate the loss of detail information resulting from downsampling. The proposed NGCG branch integrates the prior knowledge of blue and red channels into each scale of the main branch encoder, aiding the main branch in recovering over-exposed areas more effectively.

\iftrue
\vspace{+0.918mm}
\noindent{\bf Non-Green Channel Guidance.} Firstly, the pixels pertaining to the red (or blue) channel are extracted from their respective positions within each 2×2 block of a Bayer image. Subsequently, these red and blue channels are input into the NGCG branch, which then generates an initial prediction of the corresponding components in the output sRGB image. The NGCG branch is structured with a Guidance-Enhanced Block (GEB) and four downsampling blocks, serving to guide the five corresponding encoder blocks. Within the GEB, there are two 3×3 convolutional layers, with an Instance Normalization~\cite{IN16} and LeakyReLU following each layer, as well as a 3×3 convolutional operation in between. It is worth noting that depth-wise separable convolution is utilized instead of the traditional 3×3 convolution to efficiently capture local information. The downsampling blocks comprise a maxpooling layer, a modified self-attention structure, and two depth-wise separable convolution layers. This arrangement allows the NGCG branch to aid the primary backbone network in restoring over-exposed RAW images in a multi-scale manner.

\iftrue
\vspace{+0.918mm}
\noindent{\bf Cascaded Dilated Residual Block.} In detail, each CDR block incorporates three residual connections that feature dilated convolution and the LeakyReLU activation function. Subsequently, a 1×1 convolutional layer follows, which enables the CDR block to effectively utilize the features extracted from each stage of the encoder and adequately explore local texture information. Additionally, it is worth noting that the dilated convolution employed in this configuration effectively enhances the receptive field of the CDR block, thereby facilitating multi-scale contextual feature extraction.

\subsubsection{Implementation Details}

We chose the CGNet~\cite{CGNet} model for training. Since the training dataset contains overexposed images with four ratios (3, 5, 8, 10), we input the overexposed images with these four ratios into CGNet and divided them into the training set and the verification set according to a ratio of 8:2.

The training parameters include a batch size of 24, a learning rate of 0.0001, 4 undersampling layers, 9 residual layers, and the Adam optimizer. The model was trained for a total of 500 epochs, with a learning rate reduction strategy starting from 100 epochs. Training was performed using a single NVIDIA V100 GPU without loading pre-trained weights.

The experimental results are presented in Table~\ref{table:track8_results2}. After loading the weights obtained from the training, we predicted the processed test set. The pixels of the predicted results were multiplied by a suitable multiplier to improve the score. Through experiments, we found that the highest scores for overexposed images with ratios of 3, 5, 8, and 10 were 20.56, 20.49, 20.54, and 20.03, respectively.

\begin{table}[h!]
	\centering
	\setlength{\tabcolsep}{20pt}
	\caption{Experiment Results}
	\begin{tabular}{ccc}
		\toprule
		Ratio & Argument & Score \\
		\hline
		ratio=3 & 2.20 & 20.56 \\
		ratio=5 & 1.65 & 20.49 \\
		ratio=8 & 1.27 & 20.54 \\
		ratio=10 & 1.10 & 20.03 \\
		\bottomrule
	\end{tabular}
	\label{table:track8_results2}
\end{table}

The main task of this competition is to correct overexposed images. This report details our team's approach to data processing and the details of model training and predictions. The experiment verifies that the data converted from CR2 or mat format to four-channel RGGB will not be affected by performance degradation when using the CGNet model for training. Additionally, by multiplying the pixels of the predicted results, we improved the quality of the images and achieved higher scores. The experiment verifies that our strategy to solve this problem is reasonable and effective. In the end, we scored 20.56 on the RPO dataset, placing us in second place.

\subsection{LiGoxin Team's Method}

\subsubsection{Network Architecture}

We use CGNet \cite{CGNet} as a solution to the problem.CGNet contains two branches, namely a main branch based on U-net and a non-green channel guided (NGCG) branch, as shown in Fig.~\ref{fig:track8_CGNet3}.

The main branch is based on a basic U-net~\cite{ronneberger2015u} with four encoder (downsampling) and decoder (upsampling) stages. Specifically, the model first extracts initial features from a four-channel RAW image through a standard 3 × 3 convolution. In the encoder part, the model uses a HIN block to expand the receptive field and improve the robustness of the features at each scale. During the downsampling operation, the number of channels in the feature map is doubled. In the decoder part, the model uses a residual block to better extract high-level features. For skip connections, the model uses a novel cascaded dilated residual (CDR) block to extract multi-scale features and fuse them with features from the encoder part to compensate for the loss of detail information caused by downsampling. Specifically, each CDR block contains three residual connections with dilated convolutions and LeakyReLU activation functions, followed by a 1 × 1 convolution layer. This allows the CDR block to make good use of features from each stage of the encoder and fully explore local texture information. In addition, the dilated convolution used here can effectively expand the receptive field of the CDR block for multi-scale context feature extraction. 

For the NGCG branch, the pixels belonging to the corresponding position of the red (or blue) channel are first extracted in each 2 × 2 block of the Bayer image. Then, the red and blue channels are input into the NGCG branch to produce an initial estimate of the corresponding elements in the output sRGB image. The NGCG branch consists of a Guidance-Enhanced Block (GEB) and four downsampling blocks, guiding 5 corresponding encoder blocks. GEB contains 2 3 × 3 convolutional layers (followed by instance normalization and LeakyReLU) and 3 × 3 convolution operations between them. It is worth noting that the model uses depthwise separable convolution instead of standard 3 × 3 convolution to fully extract local information. The downsampling block consists of a maximum pooling layer, an improved self-attention structure, and two depthwise separable convolutional layers.

\begin{figure}[t]
    \centering
    \includegraphics[width=1\linewidth]{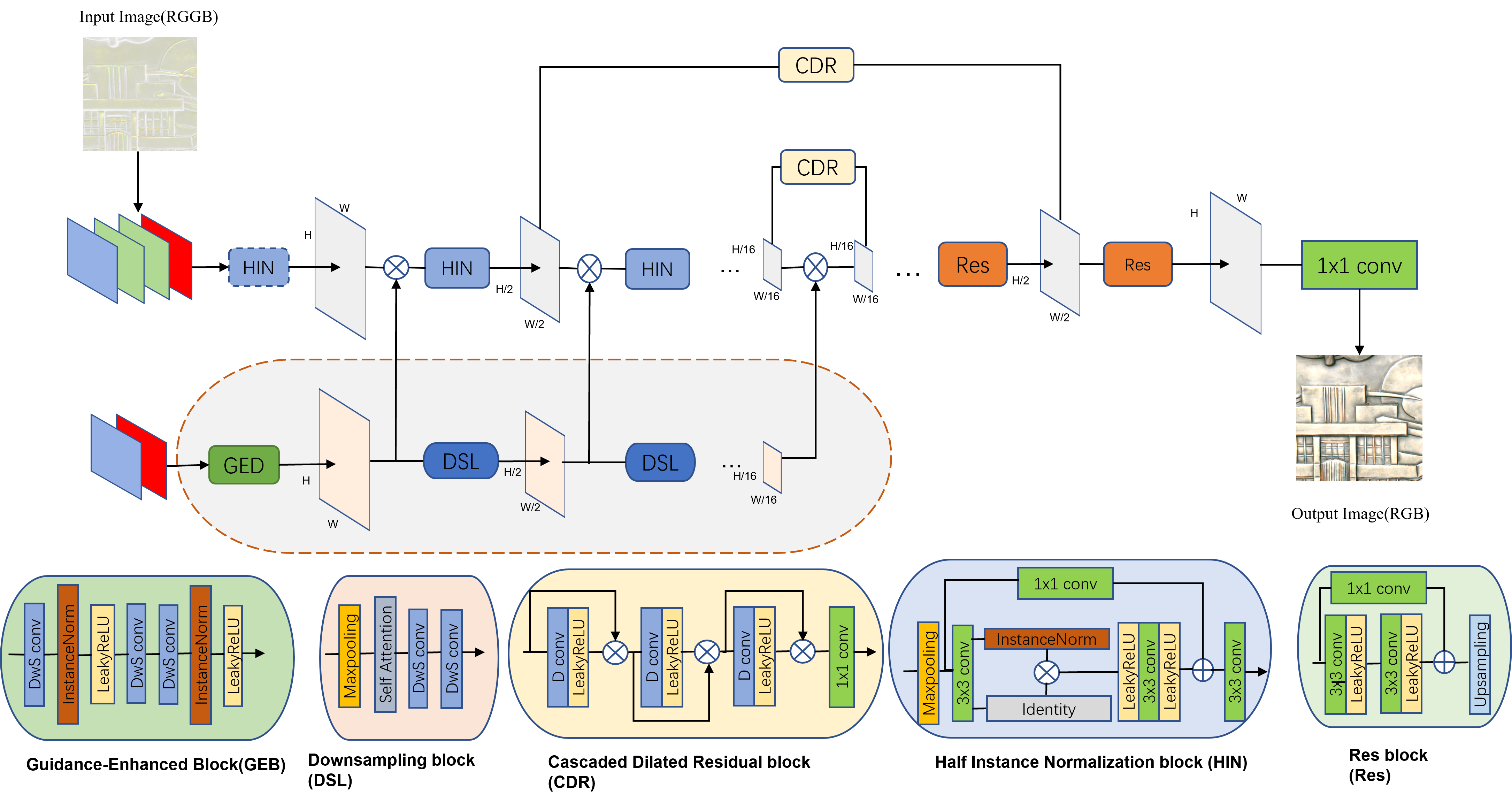}
    \caption{Architecture of CGNet~\cite{CGNet} for image over-exposure correction}
    \label{fig:track8_CGNet3}
\end{figure}

\subsubsection{Implementation Details}

The training dataset consists of 300 ground truth (gt) and corresponding overexposed images (ratios = 3, 5, 8, 10). The resolutions of RAW images and corresponding sRGB images are both 6744 x 4502.

The validation set processes RAW images into four-channel (RGGB) images, crops them, and saves them as .mat files. Unlike the training set, the validation set only includes input files and does not have ground truth.

For the test data, we converted the .mat file into a .png file and adjusted the channel order of the image. We process the exposure images using a pre-trained model that is pre-trained on the SOF dataset~\cite{CGNet} and fine-tuned on the RPO dataset. After processing, we adjust the images at different ratios and upscale the corrected images. The experimental results are shown in Fig.~\ref{fig:track8_result3}.

\begin{figure}[t]
	\centering
	\includegraphics[width=0.95\linewidth]{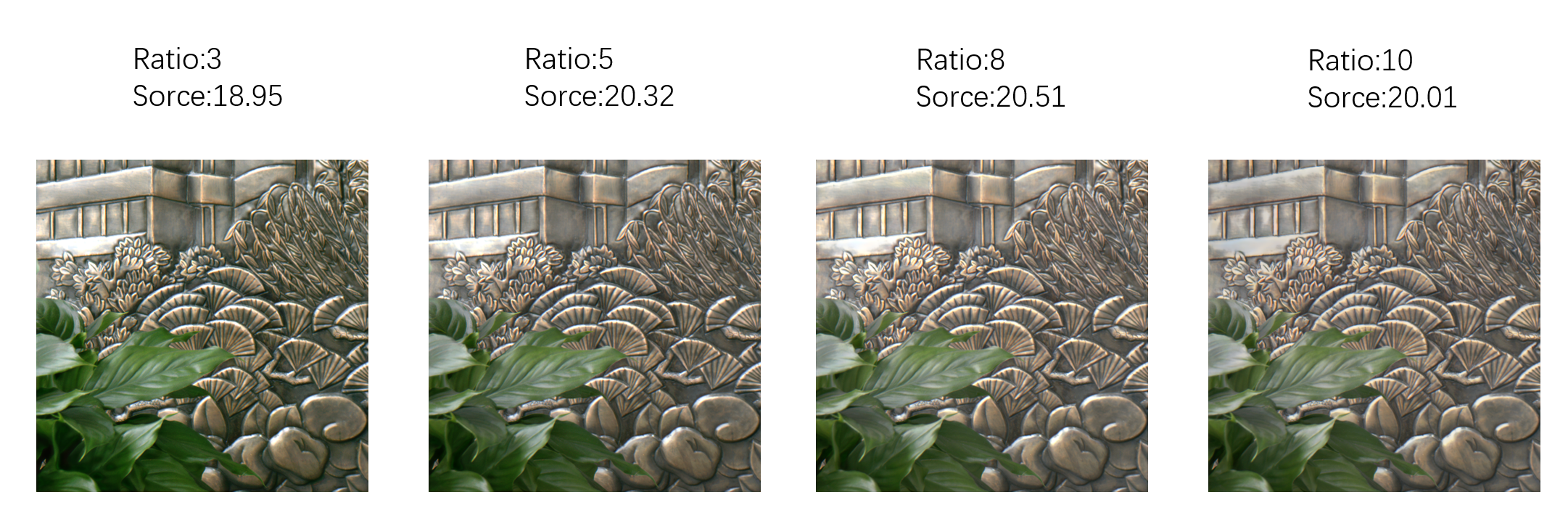}
	\caption{Different ratio examples and scores.}
	\label{fig:track8_result3}
\end{figure}

This report details our data processing methods and model usage in this task. Experiments have proven that our strategy for solving this task is reasonable and effective, and we ultimately achieved a score of 18.95 on the Ratio = 3 track of this task dataset, ranking third.

\subsection{Teams and Affiliations}
\noindent\textbf{gxj}\\
\noindent\textbf{Title:} 1st Solution Places for PBDL2024  Raw Image Based Over-Exposure Correction Challenge\\
\noindent\textbf{Members}: Xuejian Gou (\href{mailto:2497548055@qq.com}{\textcolor{magenta}{2497548055@qq.com}}), Qinliang Wang, Yang Liu, Fang Liu, Lingling Li, Wenping Ma\\
\noindent\textbf{Affiliations:} School of Artificial Intelligence, Xidian University \\

\noindent\textbf{CVCV}\\
\noindent\textbf{Title:} Raw Image Based Over-Exposure Correction\\
\noindent\textbf{Members}:Shizhan Zhao (\href{mailto:23171214628@stu.xidian.edu.cn}{\textcolor{magenta}{23171214628@stu.xidian.edu.cn}}), Yanzhao Zhang, Libo Yan, Xiaoqiang Lu, Licheng Jiao, Yuwei Guo\\
\noindent\textbf{Affiliations:} Intelligent Perception and Image Understanding Lab, Xidian University\\

\noindent\textbf{LiGoxin}\\
\noindent\textbf{Title:} Raw Image Based Over-Exposure Correction Challenge\\
\noindent\textbf{Members}:Guoxin Li (\href{mailto:lguoxin49@gmail.com}{\textcolor{magenta}{lguoxin49@gmail.com}}), Qiong Gao,  Chenyue Che, Long Sun, Xu Liu, Shuyuan Yang\\
\noindent\textbf{Affiliations:} Intelligent Perception and Image Understanding Lab, Xidian University\\

\section{Conclusion}
The three-month-long competition attracted over 300 participants, with more than 500 submissions from both industry and academic institutions. This high level of participation underscores the growing interest and investment in the field of computer vision, particularly in the integration of physics-based approaches with deep learning.

Looking forward, we anticipate continued advancements and breakthroughs in this interdisciplinary area. The success of this challenge has set a strong foundation for future research and development, encouraging more collaboration between academia and industry to solve complex vision problems. We are excited to see the future innovations and practical applications that will emerge from these efforts.


{
	\bibliographystyle{ieeenat_fullname}
	\bibliography{main.bib}
}
\end{document}